\documentclass[final]{cvpr}

\usepackage{times}
\usepackage{epsfig}
\usepackage{graphicx}
\usepackage{amsmath}
\usepackage{amssymb}
\usepackage{booktabs}
\usepackage{multirow}
\usepackage{lipsum} 
\usepackage{caption}
\usepackage{subcaption}
\usepackage{array}
\usepackage{wrapfig}
\usepackage[pagebackref=true,breaklinks=true,letterpaper=true,colorlinks,bookmarks=false,citecolor=blue]{hyperref}

\newcommand{\PreserveBackslash}[1]{\let\temp=\\#1\let\\=\temp}
\newcolumntype{C}[1]{>{\PreserveBackslash\centering}p{#1}}
\newcolumntype{R}[1]{>{\PreserveBackslash\raggedleft}p{#1}}
\newcolumntype{L}[1]{>{\PreserveBackslash\raggedright}p{#1}}

\usepackage[pagebackref=true,breaklinks=true,colorlinks,bookmarks=false]{hyperref}

\def\cvprPaperID{8732} 
\def\confYear{CVPR 2021}

\begin{document}

\title{CutPaste: Self-Supervised Learning for Anomaly Detection and Localization}

\author{Chun-Liang Li\thanks{Equal contributions.}, Kihyuk Sohn\textsuperscript{\textasteriskcentered}, Jinsung Yoon, Tomas Pfister\\
Google Cloud AI Research\\
{\tt\small \{chunliang,kihyuks,jinsungyoon,tpfister\}@google.com}
}

\maketitle

\begin{abstract}
We aim at constructing a high performance model for defect detection that detects unknown anomalous patterns of an image without anomalous data. 
To this end, we propose a two-stage framework for building anomaly detectors using normal training data only. We first learn self-supervised deep representations and then build a generative one-class classifier on learned representations. We learn representations by classifying normal data from the CutPaste, a simple data augmentation strategy that cuts an image patch and pastes at a random location of a large image. 
Our empirical study on MVTec anomaly detection dataset demonstrates the proposed algorithm is general to be able to detect various types of real-world defects. We bring the
improvement upon previous arts by 3.1 AUCs when learning representations from scratch. By transfer learning on pretrained representations on ImageNet, we achieve a new state-of-the-art \textbf{96.6} AUC.
Lastly, we extend the framework to learn and extract representations from patches to allow localizing defective areas without annotations during training.

\end{abstract}
\begin{figure*}[t]
    \centering
    \includegraphics[width=0.86\textwidth]{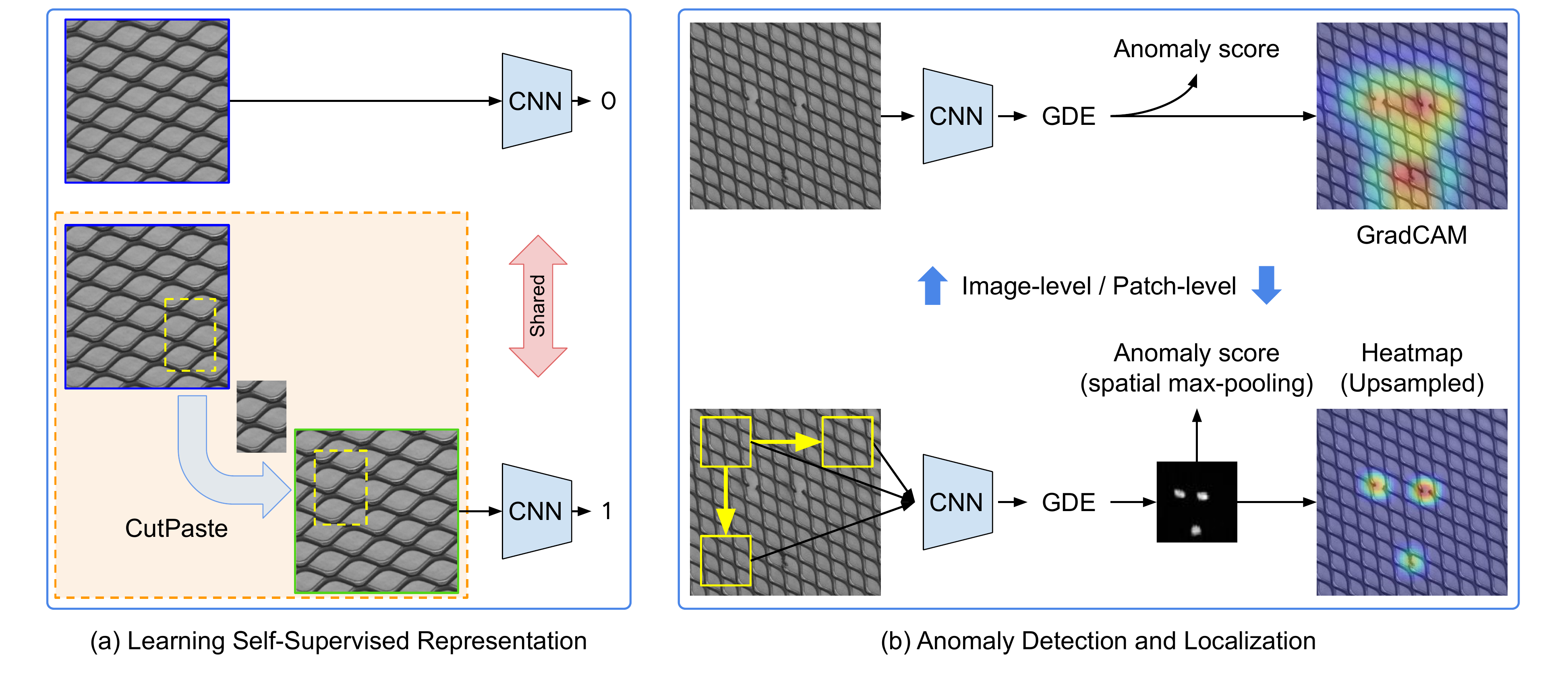}
    \vspace{-0.1in}
    \caption{An overview of our method for anomaly detection and localization. (a) A deep network (CNN) is trained to distinguish images from normal (blue) and augmented (green) data distributions by \emph{CutPaste} (orange dotted box), which cuts a small rectangular region (yellow dotted box) from normal data and pastes it at random location. Representations are trained either from the whole image or local patches. (b, top) An image-level representation makes a holistic decision for anomaly detection and is used to localize defect via GradCAM~\cite{selvaraju2017grad}. (b, bottom) A patch-level representation extracts dense features from local patches to produce anomaly score map, which is then max-pooled for detection or upsampled for localization~\cite{liznerski2020explainable}.}
    \label{fig:method_overview}
    \vspace{-0.2in}
\end{figure*}

\vspace{-0.1in}
\section{Introduction}
\label{sec:intro}
\vspace{-0.05in}

Anomaly detection aims to detect an instance containing anomalous and defective patterns that are different from those seen in normal instances. Many problems from different vision applications are anomaly detection, including manufacturing defect detection~\cite{carrera2015detecting,bergmann2019mvtec}, medical image analysis~\cite{seebock2016identifying,schlegl2017unsupervised}, and video surveillance~\cite{adam2008robust,liu2018classifier,sultani2018real}. Unlike a typical supervised classification problem, anomaly detection faces unique challenges. First, due to the nature of the problem, it is difficult to obtain a large amount of anomalous data, either labeled or unlabeled. Second, the difference between normal and anomalous patterns are often \emph{fine-grained} as defective areas might be small and subtle in high-resolution images.

Due to limited access to anomalous data, constructing an anomaly detector is often conducted under semi-supervised or one-class classification settings using normal data only. Since the distribution of anomaly patterns is unknown in advance, we train models to learn patterns of normal instances and determine anomaly if the test example is not represented well by these models. For example, an autoencoder that is trained to reconstruct normal data is used to declare anomalies when the data reconstruction error is high. Generative models
declare anomalies when the probability density is below a certain threshold. However, the anomaly score defined as an aggregation of pixel-wise reconstruction error or probability densities lacks to capture a high-level semantic information~\cite{ren2019likelihood,nalisnick2019detecting}.

Alternative methods using high-level learned representations have shown more effective for anomaly detection. For example, deep one-class classifier~\cite{ruff2018deep} demonstrates an effective  end-to-end trained one-class classifiers parameterized by deep neural networks. It outperforms its shallow counterparts, such as one-class SVMs~\cite{scholkopf2000support} and reconstruction-based approaches such as autoencoders~\cite{masci2011stacked}. In self-supervised representation learning, predicting geometric transformations of an image~\cite{golan2018deep,hendrycks2019using,bergman2020classification}, such as rotation or translation, and contrastive learning~\cite{tack2020csi,sohn2020learning} have shown to be successful in distinguishing normal data from outliers. However, most existing works focus on detecting semantic outliers (e.g., visual objects from different classes) from object-centric natural images. In Section~\ref{sec:exp_main}, we show these methods do not generalize well in detecting fine-grained anomalous patterns as in defect detection.

\begin{figure*}[t]
    \centering
    \includegraphics[width=0.86\linewidth]{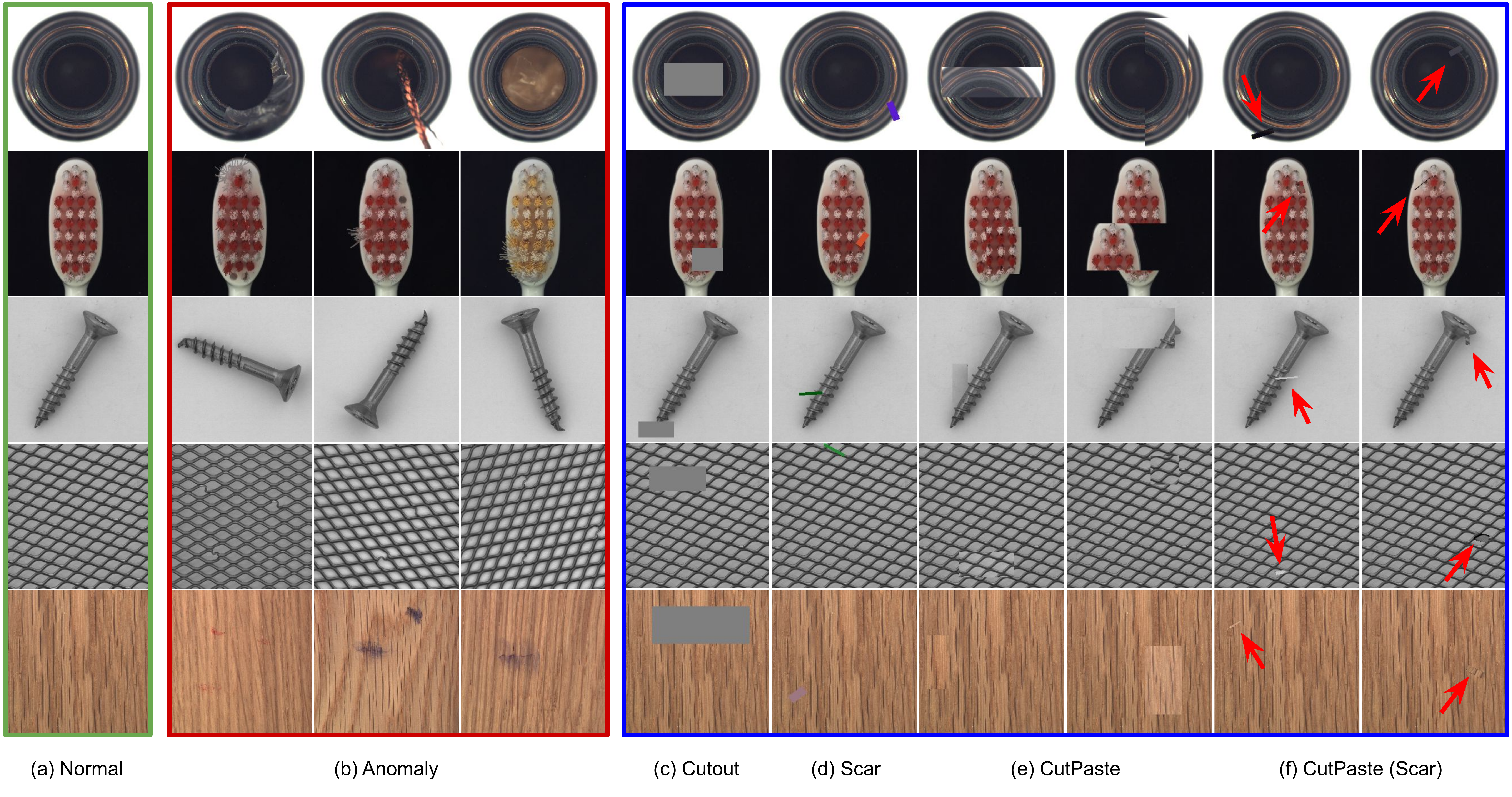}
    \vspace{-0.1in}
    \caption{Visualization of (a, green) normal, (b, red) anomaly, and (c--h, blue) augmented normal samples from bottle, toothbrush, screw, grid, and wood classes of MVTec anomaly detection dataset~\cite{bergmann2019mvtec}. Augmented normal samples are generated by baseline augmentations including (c) Cutout and (d) Scar, and our proposed (e) CutPaste and (f) CutPaste (Scar). We use red arrows in (f) to highlight the pasted patch of scar shape, a thin rectangle with rotation.}
    \label{fig:mvtec_defect_visualize}
    \vspace{-0.15in}
\end{figure*}

In this work, we tackle a one-class defect detection problem, a special case of image anomaly detection, where various forms of unknown anomalous patterns present locally in the high-resolution images.
We follow the two-stage framework~\cite{sohn2020learning}, where we first learn self-supervised representations by solving a proxy task, then build a generative one-class classifier on learned representations to distinguish data with anomalous patterns from normal ones. Our innovation is at designing a novel proxy task for self-supervised learning of representations. Specifically, we formulate a proxy classification task between normal training data and the ones augmented by the \emph{CutPaste}, the proposed data augmentation strategy that cuts an image patch and pastes at a random location of an image. CutPaste augmentation is motivated to produce a spatial \emph{irregularity} to serve as a coarse approximation of real defects, which we have no access at training. 
Rectangular patches of different sizes, aspect ratios, and rotation angles are pasted to generate diverse augmentations. Although CutPaste augmented samples (Figure~\ref{fig:mvtec_defect_visualize}(e)) are easily distinguishable from real defects and thus might be a crude approximation of a real anomaly distribution, we show that representations learned by detecting irregularity introduced by CutPaste augmentations generalize well on detecting real defects. 

We evaluate our methods on MVTec anomaly detection dataset~\cite{bergmann2019mvtec}, a real-world industrial visual inspection benchmark. By learning deep representations from scratch, we achieve \textbf{95.2} AUC on image-level anomaly detection, which outperforms existing works~\cite{huang2019inverse,yi2020patch} by at least 3.1 AUC. Moreover, we report state-of-the-art \textbf{96.6 image-level AUC} by transfer learning from an ImageNet pretrained model. 
Moreover, we explain how learned representations could be used to localize the defective areas in high-resolution images. Without using any anomaly data, a simple patch model  extension can achieve \textbf{96.0 pixel-level localization AUC}, which improves upon previous state-of-the-art~\cite{yi2020patch} ($95.7$ AUC). 
We conduct an extensive study using different types of augmentation and proxy tasks to show the effectiveness of CutPaste augmentations for self-supervised representation learning on unknown defect detection. 

\section{A Framework for Anomaly Detection}
\label{sec:method}
In this section, we present our anomaly detection framework for high-resolution image with  defects in local regions. Following~\cite{tack2020csi}, we adopt a two-stage framework for building an anomaly detector, where in the first stage we learn deep representations from normal data and then construct an one-class classifier using learned representations.
Subsequently, in Section~\ref{sec:method_cutpaste}, we present a novel method for learning self-supervised representations by predicting CutPaste augmentation, and extend to learning and extracting representations from local patches in Section~\ref{sec:method_patch}.

\subsection{Self-Supervised Learning with CutPaste}
\label{sec:method_cutpaste}
Defining good pretext tasks is essential for self-supervised representation learning.
While popular methods including rotation prediction~\cite{gidaris2018unsupervised} and contrastive learning~\cite{ye2019unsupervised,chen2020simple} have been studied in the context of semantic one-class classification~\cite{golan2018deep,hendrycks2019using,bergman2020classification,tack2020csi,sohn2020learning}, our study in Section~\ref{sec:exp_main} shows that naively applying existing methods, such as rotation prediction or contrastive learning, is sub-optimal for detecting local defects as we will show in Section~\ref{sec:exp_main}.

We conjecture that geometric transformations~\cite{golan2018deep,hendrycks2019using,bergman2020classification}, such as rotations and translations, are effective in learning representation of semantic concepts (e.g., objectness), but less of regularity (e.g., continuity, repetition). As shown in  Figure~\ref{fig:mvtec_defect_visualize}(b), anomalous patterns of defect detection typically include irregularities such as cracks (bottle, wood) or twists (toothbrush, grid). Our aim is to design an augmentation strategy creating local irregular patterns. Then we train the model to identify these local irregularity with the hope that it can generalize to unseen real defects at test time. 
%

A popular augmentation method that could create a local irregularity in image is Cutout~\cite{devries2017improved} (Figure~\ref{fig:mvtec_defect_visualize}(c)), which wipes out a randomly selected small rectangular area of an image. Cutout is found to be a useful data augmentation that enforces \emph{invariance}, leading to improved accuracy on multi-class classification tasks. In contrast, we start by \emph{discriminating} Cutout images from the normal ones. 
At first glance, the task seems easy to solve by well-crafted low level image filters. Surprisingly, as we will show in Section~\ref{sec:exp}, without the hindsight of knowing this, a deep convolution network does not learn these shortcuts. Using Cutout in the algorithm design for defect detection can also be found in~\cite{liznerski2020explainable,tayeh2020distance}.  We can make the task harder by randomly choosing colors and the scale as shown in Figure~\ref{fig:mvtec_defect_visualize}(d) to avoid naive shortcut solutions.

To further prevent learning naive decision rules for discriminating augmented images and encouraging the model to learn to detect irregularity, we propose the \emph{CutPaste} augmentation as follows:
\begin{enumerate}
    \setlength\itemsep{0em}
    \item Cut a small rectangular area of variable sizes and aspect ratios from a normal training image.
    \item Optionally, we rotate or jitter pixel values in the patch.
    \item Paste a patch back to an image at a random location.
\end{enumerate}
We show the CutPaste augmentation process in the orange dotted box of Figure~\ref{fig:method_overview} and more examples in Figure~\ref{fig:mvtec_defect_visualize}(e). Following the idea of rotation prediction~\cite{gidaris2018unsupervised}, we define the training objective of the proposed self-supervised representation learning as follows:
\begin{equation}
    \mathcal{L}_{\mathrm{CP}}\,{=}\,\mathbb{E}_{x{\in}\mathcal{X}}\big\{\mathbb{CE}(g(x), 0) + \mathbb{CE}(g(\mathrm{CP}(x)), 1)\big\}\label{eq:cutpaste_prediction}
\end{equation}
where $\mathcal{X}$ is the set of normal data, $\mathrm{CP}(\cdot)$ is a CutPaste augmentation and $g$ is a binary classifier parameterized by deep networks. $\mathbb{CE}(\cdot,\cdot)$ refers to a cross-entropy loss. In practice, data augmentations, such as translation or color jitter, are applied before feeding $x$ into $g$ or $\mathrm{CP}$.

\begin{figure*}[t]
    \centering
    \begin{subfigure}{.17\textwidth}
    \centering
    \includegraphics[width=\textwidth]{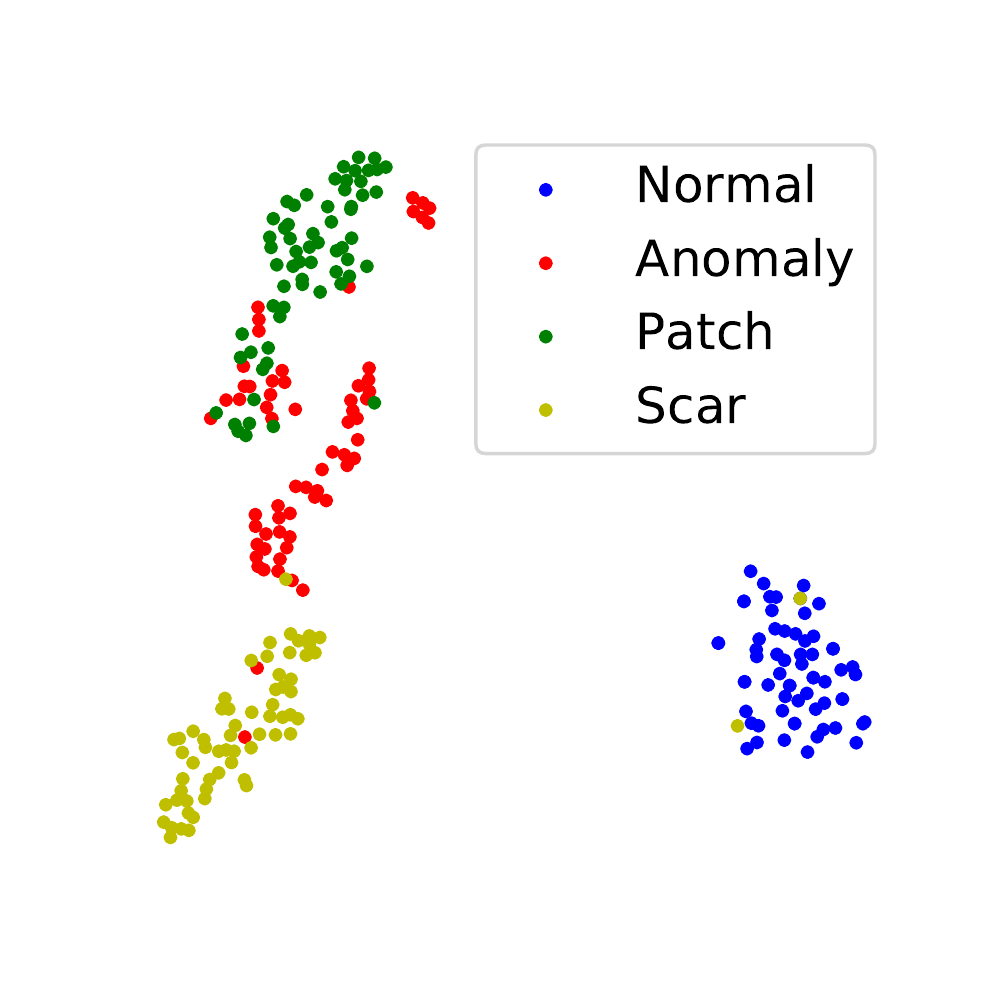}
    \caption{bottle}
    \label{fig:tsne_bottle}
    \end{subfigure}
    \begin{subfigure}{.17\textwidth}
    \centering
    \includegraphics[width=\textwidth]{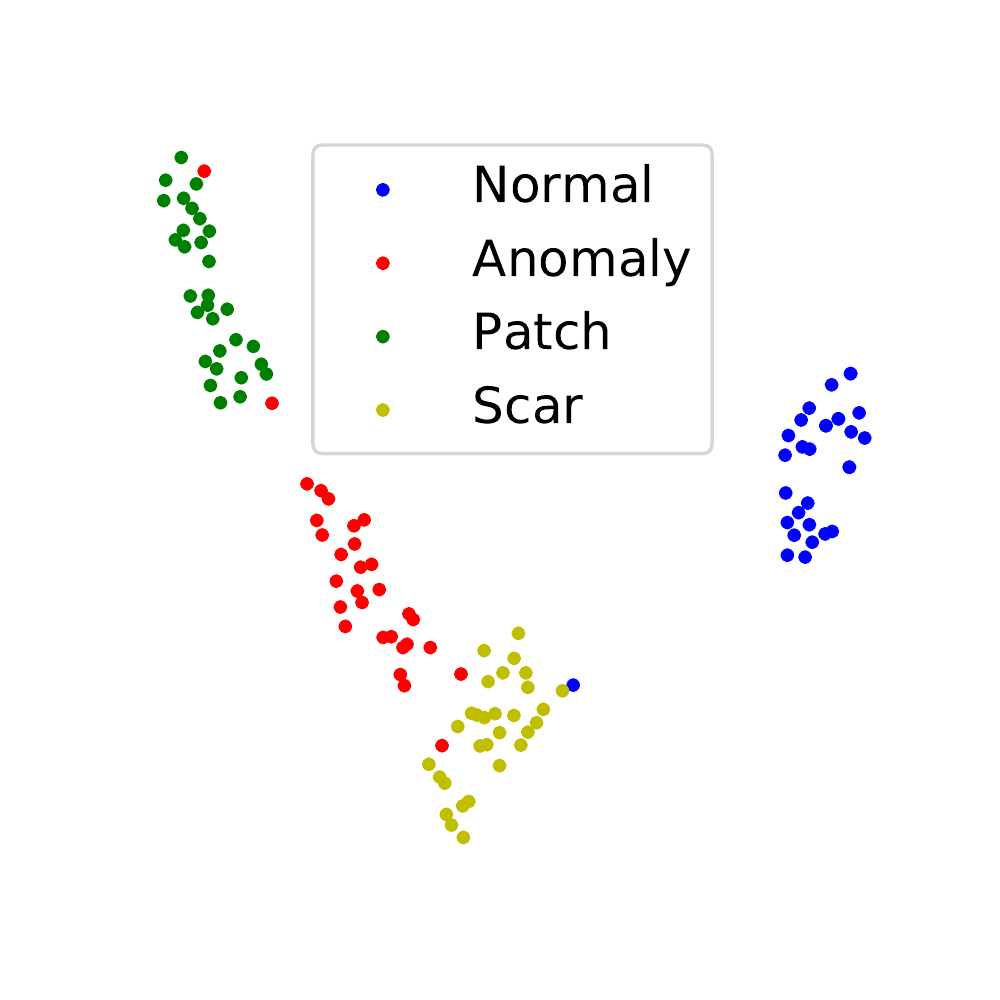}
    \caption{toothbrush}
    \label{fig:tsne_toothbrush}
    \end{subfigure}
    \begin{subfigure}{.17\textwidth}
    \centering
    \includegraphics[width=\textwidth]{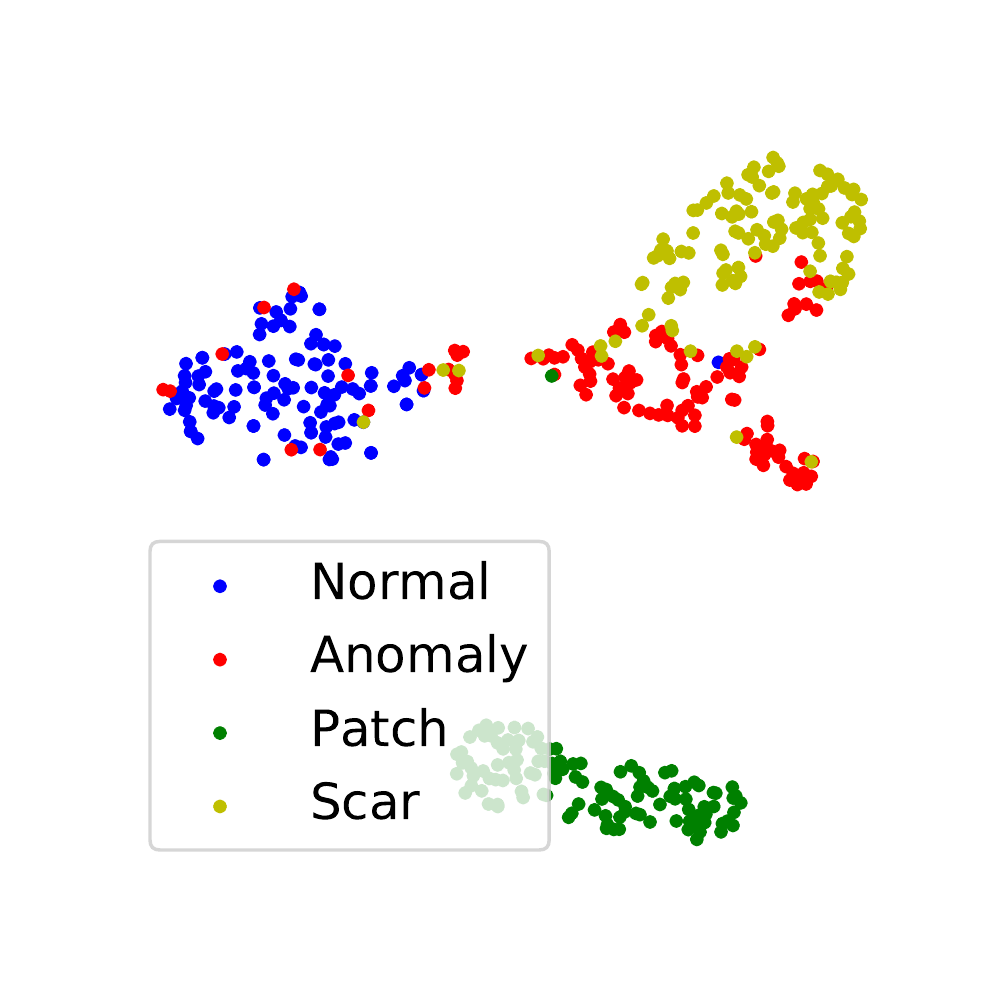}
    \caption{screw}
    \label{fig:tsne_screw}
    \end{subfigure}
    \begin{subfigure}{.17\textwidth}
    \centering
    \includegraphics[width=\textwidth]{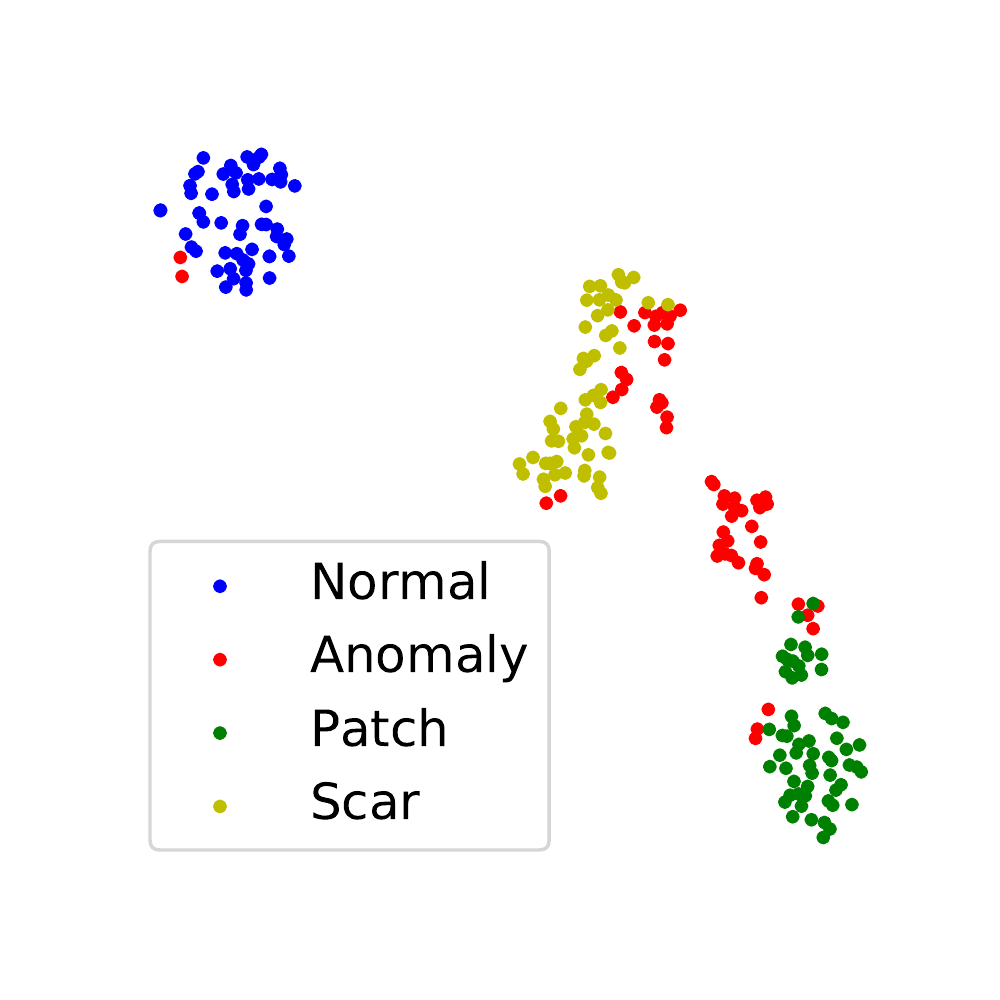}
    \caption{grid}
    \label{fig:tsne_grid}
    \end{subfigure}
    \begin{subfigure}{.17\textwidth}
    \centering
    \includegraphics[width=\textwidth]{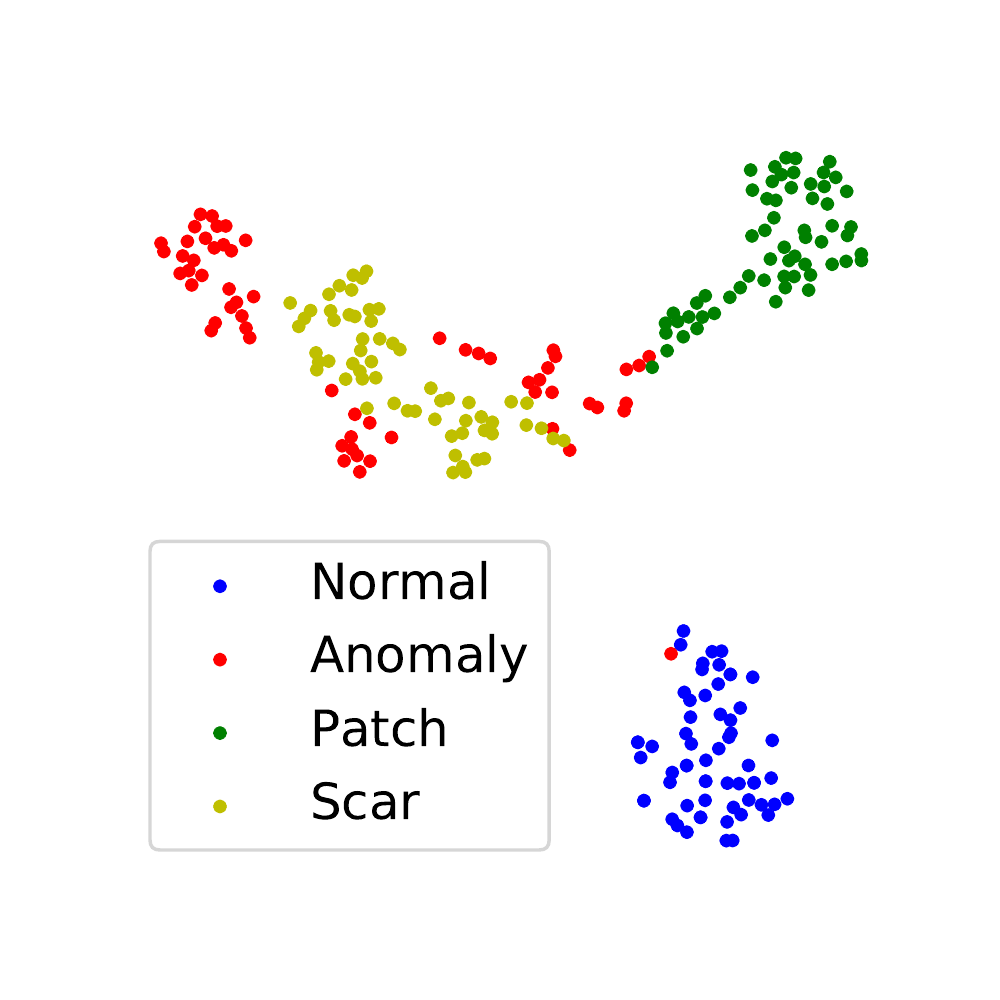}
    \caption{wood}
    \label{fig:tsne_wood}
    \end{subfigure}
    \vspace{-0.1in}
    \caption{t-SNE visualization of representations of models trained with 3-way CutPaste prediction task. We plot embeddings of normal (blue), anomaly (red), and augmented normal by CutPaste (``Patch'', green) and CutPaste-scar (``Scar'', yellow). }
    \label{fig:tsne}
    \vspace{-0.15in}
\end{figure*}

\subsection{CutPaste Variants}
\label{sec:method_cutpaste_multiclass}

\paragraph{CutPaste-Scar.}
A special case of Cutout called ``scar'' using a long-thin rectangular box of random color, as in Figure~\ref{fig:mvtec_defect_visualize}(d), is proposed in \cite{scar2020} for defect detection. 
Similarly, in addition to original CutPaste using a large rectangular patch, we propose a \emph{CutPaste-Scar} using a scar-like (long-thin) rectangular box filled with an image patch (Figure~\ref{fig:mvtec_defect_visualize}(f)).

\paragraph{Multi-Class Classification.} 
While CutPaste (large patch) and CutPaste-Scar share a similarity, the shapes of an image patch of two augmentations are very different. 
Empirically, they have their own advantages on different types of defects. To leverage the strength of both scales in the training,
we formulate a finer-grained 3-way classification task among normal, CutPaste and CutPaste-Scar by treating CutPaste variants as two separate classes.  Detailed study will be presented in Section~\ref{sec:abl_2way_vs_3way}.  

\paragraph{Similarity between CutPaste and real defects.}
The success of CutPaste may be understood from  outlier exposure~\cite{hendrycks2018deep}, where we generate the pseudo anomalies (CutPaste) during the training. Apart from using natural images as in~\cite{hendrycks2018deep}, CutPaste creates examples preserving more local structures of the normal examples (i.e., the pasted patch is from the same domain), which is more challenging for the model to 
learn to find this \emph{irregularity}. 

On the other hand, CutPaste does look similar to some real defects. A natural question is if the success of CutPaste is from a good mimic of real defects. In Figure~\ref{fig:tsne}, we show the t-SNE plots of the representations from the trained model. Clearly, the CutPaste examples are almost not overlapped with real defect examples (anomaly), but the learned representation is able to distinguish between normal example, different CutPaste augmented samples and real defects. It suggests (1) CutPaste is still not a perfect simulation of real defects and (2) learning on it to find irregularity generalizes well on unseen anomalies.

\subsection{Computing Anomaly Score}
\label{sec:method_score}
There exist various ways to compute anomaly scores via one-class classifiers. In this work, we build generative classifiers like kernel density estimator~\cite{sohn2020learning} or Gaussian density estimator~\cite{rippel2020modeling}, on representations $f$. Below, we explain how to compute anomaly scores and the trade-offs.

Although nonparametric KDE is free from distribution assumptions, it requires many examples for accurate estimation~\cite{tsybakov2008introduction} and could be computationally expensive. With limited normal training examples for defect detection, we consider a simple parametric Gaussian density estimator (GDE) whose log-density is computed as follows:
\begin{equation}
    \log p_{\mathrm{gde}}(x)\,{\propto}\,\big\{{-}\frac{1}{2}(f(x)\,{-}\,\mu)^{\top}\Sigma^{-1}(f(x)\,{-}\,\mu)\big\}\label{eq:gde_logdensity}
\end{equation}
where $\mu$ and $\Sigma$ are learned from normal training data.\footnote{We note that a mixture of Gaussian, which is a middle ground between KDE and GDE, can also be used for more expressive density modeling. We do not observe significant performance gain empirically.}

\subsection{Localization with Patch Representation}
\label{sec:method_patch}

While we present a method for learning a holistic representation of an image, learning a representation of an image patch would be preferred if we want to localize defective regions~\cite{napoletano2018anomaly,bergmann2020uninformed,yi2020patch} in addition to image-level detection. By learning and extracting representations from an image patch, we can build an anomaly detector that is able to compute the score of an image patch, which then can be used to localize the defective area.

CutPaste prediction is readily applicable to learn a patch representation -- all we need to do at training is to crop a patch before applying CutPaste augmentation. Similar to Equation~\eqref{eq:cutpaste_prediction}, the training objective can be written as:
\begin{equation}
    \mathbb{E}_{x{\in}\mathcal{X}}\big\{\mathbb{CE}(g(c(x)), 0) + \mathbb{CE}(g(\mathrm{CP}(c(x))), 1)\big\}\label{eq:cutpaste_prediction_patch}
\end{equation}
where $c(x)$ crops a patch at random location of $x$. At test time, 
we extract embeddings from all patches with a given stride.  
For each patch, we evaluate its anomaly score and use a Gaussian smoothing to propagate the score to every pixel~\cite{liznerski2020explainable}. 
In Section~\ref{sec:exp_localization}, we visualize a heatmap using patch-level detector for defect localization, along with that of an image-level detector using visual explanation techniques such as GradCAM~\cite{selvaraju2017grad}.

\section{Related Work}
\label{sec:related}
Anomaly detection under one-class classification setting, where we assume only the normal data is given during the training, has been widely studied~\cite{scholkopf2000support,tax2004support,ruff2018deep,zhai2016deep,zong2018deep,choi2018waic,ren2019likelihood,morningstar2020density,kirichenko2020normalizing}. 
Recent success of self-supervised learning in computer vision~\cite{noroozi2016unsupervised,gidaris2018unsupervised,caron2018deep,ye2019unsupervised,oord2018representation,chen2020simple,he2020momentum} has also been demonstrated effective for one-class classification and anomaly detection. One major family is by predicting geometric transformations~\cite{golan2018deep,hendrycks2019using,bergman2020classification}, such as rotation, translation or flips. The other family includes variants of contrastive learning with geometric augmentations~\cite{tack2020csi,sohn2020learning}. 
However, the success has been limited to semantic anomaly detection benchmarks, such as CIFAR-10~\cite{krizhevsky2009learning} or ImageNet~\cite{deng2009imagenet}, and as we show in Section~\ref{sec:exp_main}, methods relying on geometric transformations perform poorly on defect detection benchmarks.

Because of practical applications, such as industrial inspection or medical diagnosis, defect detection~\cite{carrera2015detecting,bergmann2019mvtec} has received lots of attention. 
The initial steps have been taken with methods including autoencoding~\cite{carrera2015detecting,bergmann2018improving,huang2019inverse,venkataramanan2019attention}, generative adversarial networks~\cite{schlegl2017unsupervised,akcay2018ganomaly}, using pretrained models on ImageNet~\cite{napoletano2018anomaly,ruff2020unifying,bergmann2020uninformed,cohen2020sub,rippel2020modeling,rudolph2020same}, and self-supervised learning by solving different proxy tasks with augmentations~\cite{yi2020patch,salehi2020puzzle,tayeh2020distance,collin2020improved}.
The proposed CutPaste prediction task is not only shown to have strong performance on defect detection, but also amenable to combine with existing methods, such as transfer learning from pretrained models for better performance or patch-based models for more accurate localization, which we demonstrate in Section~\ref{sec:exp}.

\subsection{Relation to Other Augmentations}
\label{sec:method_relation}
Although Cutout~\cite{devries2017improved} and RandomErasing~\cite{zhong2020random} are similar to CutPaste, 
they create irregularities by a small rectangular region filled with either zero or uniformly sampled pixel values instead of a structural image patch as CutPaste. 
Moreover, unlike typical use of augmentations for learning invariant representations, we learn a representation that is \emph{discriminative} to these augmentations.

Scar augmentation~\cite{scar2020} (Figure~\ref{fig:mvtec_defect_visualize}(d)) is a special case of Cutout, which uses a long-thin rectangle with random colors. While it demonstrates  strong performance, we show that CutPaste with the same scale (Figure~\ref{fig:mvtec_defect_visualize}(f)), which fills a long-thin rectangle by a patch from the same image, improves upon representations trained by predicting Cutout.

CutMix~\cite{yun2019cutmix}, which extracts a rectangular image patch from an image and pastes at random location of another image, is related to CutPaste in terms of pasting operations. One main difference is CutMix leverages existing image labels with MixUp~\cite{zhang2017mixup} in the objective while CutPaste prediction is a self-supervised learning without the need of image labels. The other difference is CutMix studies standard supervised tasks, while we aim for one-class classification.

\cite{chen2019self} presents a denoising autoencoder with patch-swap augmentation as noise process. \cite{jenni2020steering} proposes to learn representations by predicting local augmentations using GAN. Our method is simpler (e.g., no need to train decoder or GAN) while highly performant, thus more practical.

\begin{table*}[t]
    \centering
    \caption{Anomaly detection performance on MVTec AD dataset~\cite{bergmann2019mvtec}. We report AUCs of representations trained to classify CutPaste, CutPaste (scar), both (3-way), and baseline augmentations such as rotation, Cutout, or scar. 
    For comparison, we report those of deep one-class classifier~\cite{ruff2020unifying}, uninformed student~\cite{bergmann2020uninformed} and patch-SVDD~\cite{yi2020patch}. We report mean and standard error tested with 5 random seeds. Lastly, we report the AUC using ensemble of 5 CutPaste (3-way) models. The best performing model and those within standard error are bold-faced.}
    \label{tab:mvtec_main}
    \vspace{-0.05in}
    \resizebox{0.92\textwidth}{!}{
    \begin{tabular}{l|l|c|c|c||c|c|c|c|c|c|c}
        \toprule
        \multicolumn{2}{c|}{\multirow{2}{*}{Category}}  &   DOCC    &   U-Student  &   P-SVDD  &   \multirow{2}{*}{Rotation}           &   \multirow{2}{*}{Cutout}         &   \multirow{2}{*}{Scar}           &   \multirow{2}{*}{CutPaste}           &   CutPaste            &   CutPaste            &   \multirow{2}{*}{Ensemble}   \\
        \multicolumn{2}{c|}{}   &   \cite{ruff2020unifying} &   \cite{bergmann2020uninformed}   &   \cite{yi2020patch}  &               &               &               &               &   (scar)          &   (3-way)         &       \\ \midrule
\multirow{6}{*}{texture}    &   carpet  &   90.6    &   \textbf{95.3}    &   92.9    &   29.7\scriptsize{$\pm1.4$} &    35.3\scriptsize{$\pm2.3$} &    92.7\scriptsize{$\pm0.4$} &    67.9\scriptsize{$\pm1.8$} &    94.6\scriptsize{$\pm0.6$} &    93.1\scriptsize{$\pm1.1$} &    93.9    \\
    &   grid    &   52.4    &   98.7    &   94.6    &   60.5\scriptsize{$\pm7.0$} &    57.5\scriptsize{$\pm3.0$} &    74.4\scriptsize{$\pm2.5$} &    \textbf{99.9}\scriptsize{$\pm0.1$} &    95.5\scriptsize{$\pm0.3$} &    \textbf{99.9}\scriptsize{$\pm0.1$} &    \textbf{\color{red}100.0}   \\
    &   leather &   78.3    &   93.4    &   90.9    &   55.2\scriptsize{$\pm1.4$} &    67.7\scriptsize{$\pm1.5$} &    \textbf{99.9}\scriptsize{$\pm0.1$} &    99.7\scriptsize{$\pm0.1$} &    \textbf{100.0}   \scriptsize{$\pm    0.0$} &    \textbf{100.0}   \scriptsize{$\pm    0.0$} &    \textbf{\color{red}100.0}   \\
    &   tile    &   96.5    &   95.8    &   97.8    &   70.1\scriptsize{$\pm1.9$} &    71.8\scriptsize{$\pm4.0$} &    \textbf{96.7}\scriptsize{$\pm0.9$} &    \textbf{95.9}\scriptsize{$\pm1.0$} &    89.4\scriptsize{$\pm2.8$} &    93.4\scriptsize{$\pm1.0$} &    94.6    \\
    &   wood    &   91.6    &   95.5    &   96.5    &   95.8\scriptsize{$\pm1.1$} &    92.0\scriptsize{$\pm0.8$} &    \textbf{98.9}\scriptsize{$\pm0.2$} &    94.9\scriptsize{$\pm0.5$} &    \textbf{98.7}\scriptsize{$\pm0.3$} &    \textbf{98.6}\scriptsize{$\pm0.5$} &    \textbf{\color{red}99.1}    \\ \cmidrule{2-12}
    &   average &   81.9    &   95.7    &   94.5    &   62.3\scriptsize{$\pm2.6$} &    64.9\scriptsize{$\pm2.3$} &    92.5\scriptsize{$\pm0.8$} &    91.7\scriptsize{$\pm0.7$} &    95.7\scriptsize{$\pm0.8$} &    \textbf{97.0}\scriptsize{$\pm0.5$} &    \textbf{\color{red}97.5}    \\ \midrule
\multirow{11}{*}{object}    &   bottle  &   \textbf{99.6}    &   96.7    &   98.6    &   95.0\scriptsize{$\pm0.7$} &    88.7\scriptsize{$\pm0.8$} &    98.5\scriptsize{$\pm0.2$} &    99.2\scriptsize{$\pm0.2$} &    98.0\scriptsize{$\pm0.5$} &    98.3\scriptsize{$\pm0.5$} &    98.2    \\
    &   cable   &   \textbf{90.9}    &   82.3    &   90.3    &   85.3\scriptsize{$\pm0.8$} &    80.2\scriptsize{$\pm1.4$} &    78.3\scriptsize{$\pm1.7$} &    87.1\scriptsize{$\pm0.8$} &    78.8\scriptsize{$\pm2.9$} &    80.6\scriptsize{$\pm0.5$} &    81.2    \\
    &   capsule &   91.0    &   92.8    &   76.7    &   71.8\scriptsize{$\pm1.4$} &    69.5\scriptsize{$\pm1.1$} &    82.9\scriptsize{$\pm0.7$} &    87.9\scriptsize{$\pm0.7$} &    \textbf{95.3}\scriptsize{$\pm0.8$} &    \textbf{96.2}\scriptsize{$\pm0.5$} &    \textbf{\color{red}98.2}    \\
    &   hazelnut    &   95.0    &   91.4    &   92.0    &   83.6\scriptsize{$\pm0.8$} &    69.7\scriptsize{$\pm1.3$} &    \textbf{98.9}\scriptsize{$\pm0.2$} &    91.3\scriptsize{$\pm0.6$} &    96.7\scriptsize{$\pm0.4$} &    97.3\scriptsize{$\pm0.3$} &    98.3    \\
    &   metal nut   &   85.2    &   94.0    &   94.0    &   72.7\scriptsize{$\pm0.5$} &    84.6\scriptsize{$\pm0.7$} &    86.9\scriptsize{$\pm1.5$} &    96.8\scriptsize{$\pm0.5$} &    97.9\scriptsize{$\pm0.2$} &    \textbf{99.3}\scriptsize{$\pm0.2$} &    \textbf{\color{red}99.9}    \\
    &   pill    &   80.4    &   86.7    &   86.1    &   79.2\scriptsize{$\pm1.4$} &    78.7\scriptsize{$\pm0.7$} &    82.2\scriptsize{$\pm1.4$} &    \textbf{93.4}\scriptsize{$\pm0.9$} &    85.8\scriptsize{$\pm1.3$} &    \textbf{92.4}\scriptsize{$\pm1.3$} &    \textbf{\color{red}94.9}    \\
    &   screw   &   86.9    &   \textbf{87.4}    &   81.3    &   35.8\scriptsize{$\pm2.9$} &    17.6\scriptsize{$\pm4.4$} &    11.3\scriptsize{$\pm2.2$} &    54.4\scriptsize{$\pm1.7$} &    83.7\scriptsize{$\pm0.7$} &    86.3\scriptsize{$\pm1.0$} &    \textbf{\color{red}88.7}    \\
    &   toothbrush  &   96.4    &   98.6    &   \textbf{100.0}   &   99.1\scriptsize{$\pm0.2$} &    98.1\scriptsize{$\pm0.6$} &    94.8\scriptsize{$\pm1.0$} &    99.2\scriptsize{$\pm0.2$} &    96.7\scriptsize{$\pm0.4$} &    98.3\scriptsize{$\pm0.9$} &    99.4    \\
    &   transistor  &   90.8    &   83.6    &   91.5    &   88.9\scriptsize{$\pm0.4$} &    82.5\scriptsize{$\pm1.2$} &    92.0\scriptsize{$\pm0.7$} &    \textbf{96.4}\scriptsize{$\pm0.7$} &    91.1\scriptsize{$\pm0.6$} &    \textbf{95.5}\scriptsize{$\pm0.5$} &    \textbf{\color{red}96.1}    \\
    &   zipper  &   92.4    &   95.8    &   97.9    &   74.3\scriptsize{$\pm1.6$} &    75.7\scriptsize{$\pm1.0$} &    86.8\scriptsize{$\pm0.9$} &    \textbf{99.4}\scriptsize{$\pm0.1$} &    \textbf{99.5}\scriptsize{$\pm0.1$} &    \textbf{99.4}\scriptsize{$\pm0.2$} &    \textbf{\color{red}99.9}    \\ \cmidrule{2-12}
    &   average &   90.9    &   90.9    &   90.8    &   78.6\scriptsize{$\pm1.1$} &    74.5\scriptsize{$\pm1.3$} &    81.3\scriptsize{$\pm1.1$} &    90.5\scriptsize{$\pm0.6$} &    92.4\scriptsize{$\pm0.8$} &    \textbf{94.3}\scriptsize{$\pm0.6$} &    \textbf{\color{red}95.5}    \\ \midrule
        \multicolumn{2}{c|}{average}    &   87.9    &   92.5    &   92.1    &   73.1\scriptsize{$\pm1.6$} &    71.3\scriptsize{$\pm1.6$} &    85.0\scriptsize{$\pm1.0$} &    90.9\scriptsize{$\pm0.7$} &    93.5\scriptsize{$\pm0.8$} &    \textbf{95.2}\scriptsize{$\pm0.6$} &    \textbf{\color{red}96.1}    \\ \bottomrule
    \end{tabular}
    }
    \vspace{-0.1in}
\end{table*}

\begin{figure*}[t]
    \centering
    \includegraphics[width=0.95\textwidth]{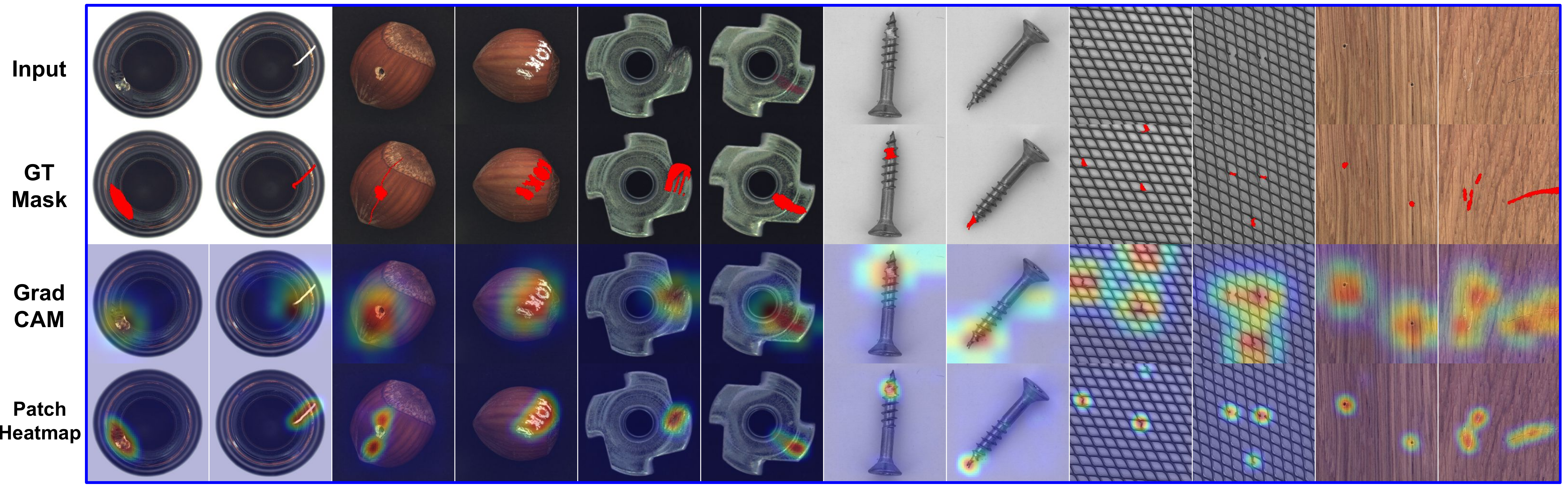}
    \caption{Defect localization on bottle, hazelnut, metal nut, screw, wood and grid classes of MVTec datasets. From top to bottom, input images, those with ground-truth localization mask in red, GradCAM results using image-level detector, and heatmaps using patch-level detector. We provide more examples in Appendix~\ref{sec:app_localization}.}
    \label{fig:localization}
    \vspace{-0.2in}
\end{figure*}

\section{Experiments}
\label{sec:exp}
We conduct most experiments on MVTec Anomaly Detection dataset~\cite{bergmann2019mvtec} that contains 10 object and 5 texture categories for anomaly detection. The dataset is composed of normal images for training and both normal and anomaly images with various types of defect for testing. It also provides pixel-level annotations for defective test images. The dataset is relatively small scale in number of images, where the number of training images varies from $60$ to $391$, posing a unique challenge for learning deep representations.

We follow one-class classification protocol, 
also known as semi-supervised anomaly detection~\cite{chandola2009anomaly},\footnote{While previous works~\cite{bergmann2019mvtec,bergmann2020uninformed} have used \emph{unsupervised} to describe their settings, it could be misleading as training data is curated to include normal data only.}
where we train a one-class classifier for each category on its respective normal training examples. 
Following~\cite{sohn2020learning}, we learn the representations by augmentation prediction from scratch with 
ResNet-18~\cite{he2016deep} plus an MLP projection head on top of average pooling layer followed by the last linear layer. We construct a Gaussian density estimation (GDE) as Equation~\eqref{eq:gde_logdensity} for anomaly detector based on the top pooled features.

We train a model on $256{\times}256$ image. We note that the same training strategy, such as the selection of hyperparameters or data augmentations, is applied to all categories. Detailed settings of training can be found in Appendix~\ref{sec:app_detail_exp}.

\subsection{Main Results}
\label{sec:exp_main}
We report the anomaly detection performance in Table~\ref{tab:mvtec_main}. We run experiments 5 times with different random seeds and report the mean AUC and standard error for each category. We also report the average of mean and standard errors for texture, object, and all categories.

We test representations trained with different proxy tasks of self-supervised learning, 
including baselines such as rotation~\cite{golan2018deep}, Cutout or scar predictions, the proposed CutPaste, CutPaste-Scar predictions, and using both with 3-way classification. 
We also compare with previous works, including deep one-class classifier (DOCC)~\cite{ruff2020unifying}, uninformed student~\cite{bergmann2019mvtec}, and patch SVDD~\cite{yi2020patch}. We note that some of these methods use ImageNet pretrained model for transfer learning, either by fine-tuning (DOCC) or distillation (uninformed student). 
The results are in Table~\ref{tab:mvtec_main}. 

Rotation prediction is demonstrated to be powerful in semantic anomaly detection~\cite{sohn2020learning}. However, it results in unsatisfactory $73.1$ AUC in defect detection compared with the Scar prediction (85.0), a Cutout variant.  Some failure of rotation prediction is due to the unaligned objects, such as \texttt{screw} shown in Figure~\ref{fig:mvtec_defect_visualize}. For aligned objects, although it performs well on \texttt{toothbrush}, it is sub-optimal on \texttt{capsule}. 
Detailed ablation study of Cutout variants can be found in Section~\ref{sec:abl}.

CutPaste and CutPaste-Scar, which improve Cutout and Scar prediction by avoiding potential naive solutions, outperform other augmentation predictions with $90.9$ and $93.5$ AUCs, respectively. With a finer-grained 3-way classification to leverage different scale of CutPaste, we achieve the best $95.2$ AUC, which surpasses existing works on learning from scratch, such as P-SVDD~\cite{yi2020patch} ($92.1$ AUC).  
The proposed data-driven approach via CutPaste is also better than existing works leveraging pretrained networks, including
DOCC~\cite{ruff2020unifying} ($87.9$ AUC) with pretrained VGG16 and Uninformed Student~\cite{bergmann2020uninformed} ($92.5$ AUC) with pretrained ResNet18.  
Last, we further improve the AUC to ${\bf96.1}$ by ensembling anomaly scores from 5 CutPaste (3-way) models.

\subsection{Defect Localization}
\label{sec:exp_localization}
We conduct anomaly localization experiments using our representations trained with 3-way classification task. 
One challenge to accurate localization of defect is that it is difficult to use a heatmap-style approach for localization as our model learns a holistic representation of an image. Instead, we use visual explanation techniques, GradCAM~\cite{selvaraju2017grad}, to highlight the area affecting the decision of anomaly detector. We show qualitative results in the second row of Figure~\ref{fig:localization}, which are visually pleasing. We further evaluate the pixel-wise localization AUC, achieving $88.3$.

Instead, we learn a representation of an image patch using CutPaste prediction, as in Section~\ref{sec:method_patch}. 
We train models of $64{\times}64$ patches from $256{\times}256$ image.
At test time, we densely extract anomaly scores with a stride of $4$ and propagate  the anomaly scores via receptive field upsampling with Gaussian smoothing~\cite{liznerski2020explainable}. 
We report a localization AUC in Table~\ref{tab:mvtec_localization}.
Our patch-based model achieves ${\bf96.0}$ AUC. Specifically, our model shows strong performance on texture categories over previous state-of-the-art ($96.3$ AUC compared to $93.7$). 
We also outperforms the DistAug contrastive learning~\cite{sohn2020learning}, which only results in 90.4 localization AUC.
Finally, we visualize representative samples for localization in Figure~\ref{fig:localization}, showing accurate localization even when defects are tiny. More comprehensive results on defect localization are given in Appendix~\ref{sec:app_localization}.

\begin{table}[t]
    \centering
    \caption{Pixel-wise localization AUC on MVTec dataset. The best and models within standard error are bold-faced.}
    \label{tab:mvtec_localization}
    \vspace{-0.1in}
    \resizebox{0.45\textwidth}{!}{
    \begin{tabular}{l|l|c|c|c}
        \toprule		
        \multicolumn{2}{c|}{Category}  &   FCDD~\cite{liznerski2020explainable}    &   P-SVDD~\cite{yi2020patch}  &   CutPaste (3-way)               \\ \midrule
\multirow{6}{*}{texture}    &   carpet  &   96  &   92.6    &   \textbf{98.3}\scriptsize{$\pm0.0$}  \\
    &   grid    &   91  &   96.2    &   \textbf{97.5}\scriptsize{$\pm0.1$}  \\
    &   leather &   98  &   97.4    &   \textbf{99.5}\scriptsize{$\pm0.0$}  \\
    &   tile    &   91  &   \textbf{91.4}    &   90.5\scriptsize{$\pm0.2$}  \\
    &   wood    &   88  &   90.8    &   \textbf{95.5}\scriptsize{$\pm0.1$}  \\ \cmidrule{2-5}
    &   average &   93  &   93.7    &   \textbf{96.3}\scriptsize{$\pm0.1$}  \\ \midrule
\multirow{11}{*}{object}    &   bottle  &   97  &   \textbf{98.1}    &   97.6\scriptsize{$\pm0.1$}  \\
    &   cable   &   90  &   \textbf{96.8}    &   90.0\scriptsize{$\pm0.2$}  \\
    &   capsule &   93  &   95.8    &   \textbf{97.4}\scriptsize{$\pm0.1$}  \\
    &   hazelnut    &   95  &   \textbf{97.5}    &   97.3\scriptsize{$\pm0.1$}  \\
    &   metal nut   &   94  &   \textbf{98.0}    &   93.1\scriptsize{$\pm0.4$}  \\
    &   pill    &   81  &   95.1    &   \textbf{95.7}\scriptsize{$\pm0.1$}  \\
    &   screw   &   86  &   95.7    &   \textbf{96.7}\scriptsize{$\pm0.1$}  \\
    &   toothbrush  &   94  &   \textbf{98.1}    &   \textbf{98.1}\scriptsize{$\pm0.0$}  \\
    &   transistor  &   88  &   \textbf{97.0}    &   93.0\scriptsize{$\pm0.2$}  \\
    &   zipper  &   92  &   95.1    &   \textbf{99.3}\scriptsize{$\pm0.0$}  \\ \cmidrule{2-5}
    &   average &   91  &   \textbf{96.7}    &   95.8\scriptsize{$\pm0.1$}  \\ \midrule
        \multicolumn{2}{c|}{average}    &   92  &   95.7    &   \textbf{96.0}\scriptsize{$\pm0.1$}  \\ \bottomrule
    \end{tabular}
    }
    \vspace{-0.1in}
\end{table}

\subsection{Transfer Learning with Pretrained Models}
\label{sec:exp_fine-tune}
In Section~\ref{sec:exp_main}, we have shown the proposed data-driven approach is better than leveraging pretrained networks, such as DOCC~\cite{ruff2020unifying} and Uninformed Student~\cite{bergmann2020uninformed}. It is consistent with the prior study on semantic anomaly detection~\cite{sohn2020learning}.
On the other hand, pretrained EfficientNet~\cite{tan2019efficientnet} is found useful for defect detection~\cite{rippel2020modeling}. 
As shown in Table~\ref{tab:mvtec_pretrain}, without fine-tuning, the representation from the pretrained EfficientNet (B4) results in $94.5$ AUC, which is competitive with the proposed CutPaste prediction ($95.2$ from Table~\ref{tab:mvtec_main}).

\begin{table}[t]
    \centering
    \caption{Detection performance on MVTec dataset using representations of EfficientNet (B4)~\cite{tan2019efficientnet} pretrained on ImageNet~\cite{deng2009imagenet} and fine-tuned by the CutPaste (3-way). The number is bold when it is better than its pretrained or finetuned counterpart under the same feature (pool v.s. level-7). }
    \label{tab:mvtec_pretrain}
    \vspace{-0.1in}
    \resizebox{0.45\textwidth}{!}{
    \begin{tabular}{l|l|c|c||c|c}
        \toprule
		\multicolumn{2}{c|}{\multirow{2}{*}{Category}}&\multicolumn{2}{c||}{Pool}&\multicolumn{2}{c}{Level-7}\\
		\cmidrule{3-6}
		\multicolumn{2}{c|}{}&Pretrain & Finetune&Pretrain & Finetune\\

		\midrule
\multirow{6}{*}{texture}    &   carpet  &   98.3    &   {\bf100.0}\scriptsize{$\pm    0.0$}  &   97.6    &   {\bf100.0}\scriptsize{$\pm    0.0$}\\    
    &   grid    &   96.4    &   {\bf98.8}\scriptsize{$\pm0.1$}  &   98.2    &   {\bf99.1}\scriptsize{$\pm0.0$}\\    
    &   leather &   {\bf100.0}  &   {\bf100.0}\scriptsize{$\pm    0.0$}  &   {\bf100.0}  &   {\bf100.0}\scriptsize{$\pm    0.0$}\\    
    &   tile    &   {\bf99.9}   &   98.9\scriptsize{$\pm0.2$}  &   {\bf99.9}   &   {\bf99.8}\scriptsize{$\pm0.2$}\\    
    &   wood    &   99.7    &   {\bf99.8}\scriptsize{$\pm0.0$}  &   99.6    &   {\bf99.8}\scriptsize{$\pm0.0$}\\    \cmidrule{2-6}
    &   average &   98.9    &   {\bf99.5}\scriptsize{$\pm0.0$}  &   99  &   {\bf99.7}\scriptsize{$\pm0.0$}\\    \midrule
\multirow{11}{*}{object}    &   bottle  &   99.8    &   {\bf100.0}\scriptsize{$\pm    0.0$}  &   {\bf100.0}  &   {\bf100.0}\scriptsize{$\pm    0.0$}\\    
    &   cable   &   91.2    &   {\bf93.9}\scriptsize{$\pm0.1$}  &   {\bf96.5}   &   {\bf96.2}\scriptsize{$\pm0.3$}\\    
    &   capsule &   93  &   {\bf94.3}\scriptsize{$\pm0.3$}  &   94.7    &   {\bf95.4}\scriptsize{$\pm0.1$}\\    
    &   hazelnut    &   96.6    &   {\bf99.7}\scriptsize{$\pm0.0$}  &   {\bf100.0}  &   99.9\scriptsize{$\pm0.0$}\\    
    &   metalnut    &   94.3    &   {\bf98.7}\scriptsize{$\pm0.1$}  &   97.7    &   {\bf98.6}\scriptsize{$\pm0.0$}\\    
    &   pill    &   81.9    &   {\bf91.3}\scriptsize{$\pm0.2$}  &   91  &   {\bf93.3}\scriptsize{$\pm0.2$}\\    
    &   screw   &   {\bf86.3}   &   86\scriptsize{$\pm    0.1$}  &   {\bf92.0}   &   86.6\scriptsize{$\pm0.2$}\\    
    &   toothbrush  &   89.3    &   {\bf92.8}\scriptsize{$\pm0.2$}  &   90.3    &   {\bf90.7}\scriptsize{$\pm0.1$}\\    
    &   transistor  &   94.6    &   {\bf95.6}\scriptsize{$\pm0.2$}  &   {\bf97.5}   &   {\bf97.5}\scriptsize{$\pm0.2$}\\    
    &   zipper  &   95.6    &   {\bf99.9}\scriptsize{$\pm0.0$}  &   97  &   {\bf99.9}\scriptsize{$\pm0.1$}\\    \cmidrule{2-6}
    &   average &   92.3    &   {\bf95.2}\scriptsize{$\pm0.1$}  &   95.7    &   {\bf95.8}\scriptsize{$\pm0.1$}\\    \midrule
        \multicolumn{2}{c|}{average}    &   94.5    &   {\bf96.6}\scriptsize{$\pm0.1$}  &   96.8    &   {\bf97.1}\scriptsize{$\pm0.0$}\\    \bottomrule
    \end{tabular}
    }
    \vspace{-0.1in}
\end{table}
Here we demonstrate that the proposed self-supervised learning via CutPaste is versatile, which can also be used to improve the pretrained networks to better adapt to the data. 
We use pretrained EfficientNet (B4) as a backbone, and follow the standard fine-tuning steps to train with the same CutPaste prediction (3-way) task. Detailed settings can be found in Appendix~\ref{sec:app_detail_exp}. We show the results in Table~\ref{tab:mvtec_pretrain}. 
After fine-tuning via CutPaste, we achieve the {\bf new state-of-the-art $\mathbf{96.6}$ AUC}. Furthermore, CutPaste prediction is a general and useful strategy to adapt to the data for most of the situations. For example, CutPaste improves by a large margin on class \texttt{pill} ($81.9\,{\rightarrow}\,91.3$). For many nearly perfect situations, such as \texttt{bottle}, CutPaste is still able to improve by a small margin.
Last, as suggested by~\cite{lee2018simple,rippel2020modeling}, we investigate the performance of various deep features. We find that level-7 feature shows the best performance, and we further improve the level-7 feature of EfficientNet from $96.8$ (pretrained) to $\mathbf{97.1}${\scriptsize{$\pm0.0$}} with CutPaste.

\section{Ablation Study}
\label{sec:abl}
We conduct various additional studies to provide deeper insights of the proposed CutPaste. 
We first compare CutPaste with different Cutout variants in addition to the standard ones reported in Section~\ref{sec:exp_main}. Second, we showcase the representation learned via predicting CutPaste generalizes well to more crafted unseen defects. Last, we compare with the semantic anomaly detection.

\begin{table}[t]
    \centering
    \caption{Detection AUCs of representations trained to predict Cutout, with mean pixel value, with random color, Confetti noise~\cite{liznerski2020explainable}, or the proposed CutPaste.}
    \label{tab:cutout_to_cutpaste}
    \vspace{-0.1in}
    \resizebox{0.48\textwidth}{!}{
    \begin{tabular}{l|C{1.5cm}|C{1.4cm}|C{1.4cm}|C{1.4cm}|C{1.4cm}}
        \toprule
        \multirow{2}{*}{Category} & \multirow{2}{*}{Cutout} & Cutout & Cutout &  \multirow{2}{*}{Confetti}  & \multirow{2}{*}{CutPaste} \\ 
        & & (Mean) & (Color) & \\ \midrule
        texture & 64.9\scriptsize{$\pm2.3$} & 65.5\scriptsize{$\pm1.8$} & 70.5\scriptsize{$\pm2.2$} & 80.1\scriptsize{$\pm2.3$} & 91.7\scriptsize{$\pm0.7$} \\
        object & 74.5\scriptsize{$\pm1.3$} & 78.1\scriptsize{$\pm1.1$} & 78.9\scriptsize{$\pm1.0$} & 76.7\scriptsize{$\pm0.3$} & 90.5\scriptsize{$\pm0.6$} \\ \midrule
        all & 71.3\scriptsize{$\pm1.7$} & 73.9\scriptsize{$\pm1.3$} & 76.1\scriptsize{$\pm1.4$} & 77.8\scriptsize{$\pm1.5$} & 90.9\scriptsize{$\pm0.7$} \\
        \bottomrule
    \end{tabular}
    }
    \vspace{-0.1in}
\end{table}

\begin{figure}[t]
    \centering
    \includegraphics[width=0.98\linewidth]{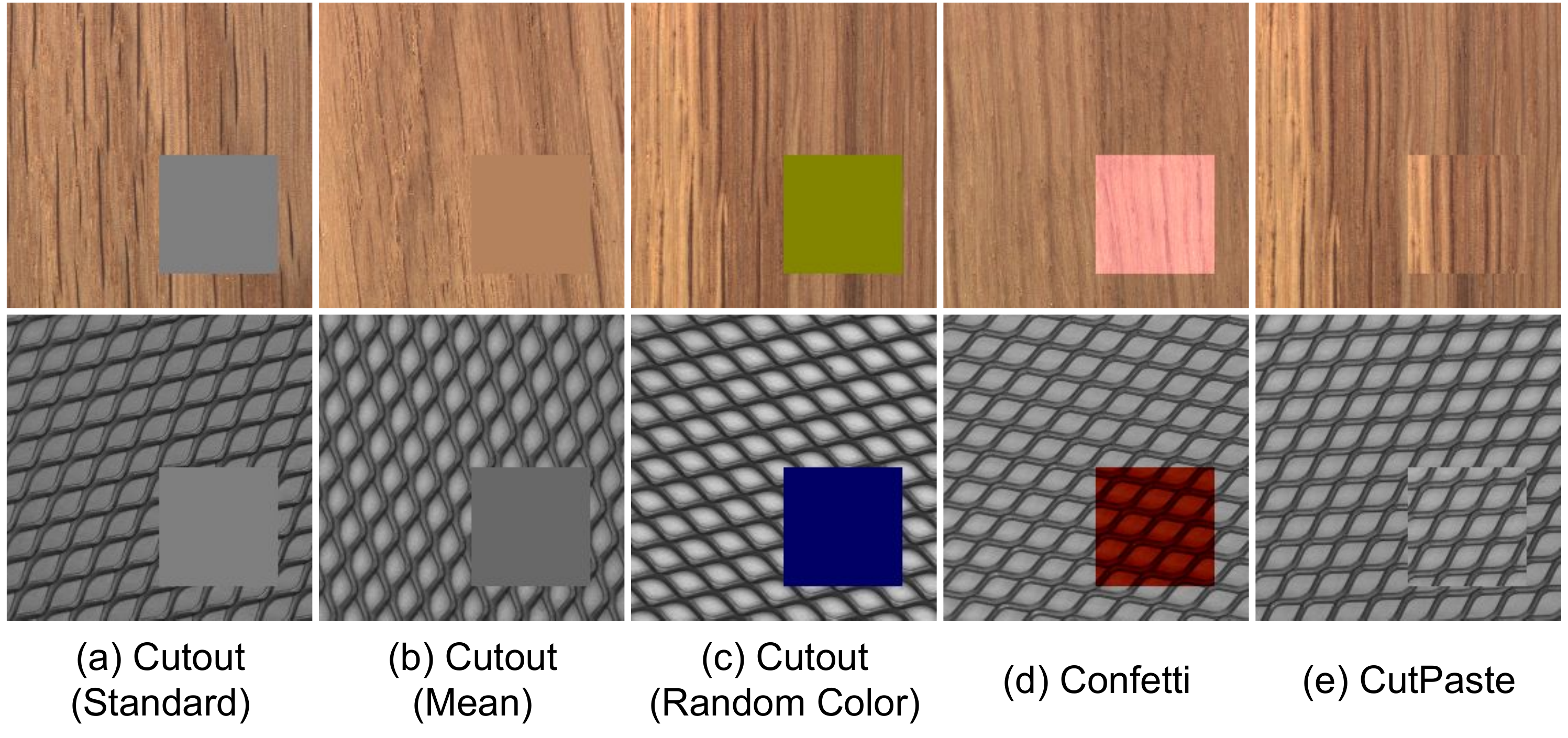}
    \vspace{-0.1in}
    \caption{Visual comparison between the proposed CutPaste and Cutout variants, including filling with grey color, mean pixel values, random colors and Confetti noise~\cite{liznerski2020explainable}.  }
    \label{fig:cutout_comparison}
    \vspace{-0.15in}
\end{figure}

\subsection{From Cutout to CutPaste}
\label{sec:abl_Cutout_to_cutpaste}
We evaluate the performance of representations trained to predict variants of Cutout augmentations whose areas are filled by grey color (standard), mean pixel values, random color, or image patch from different location, i.e., CutPaste. We also test Confetti noise~\cite{liznerski2020explainable} that jitters a color of a local patch. 
We show samples from considered augmentations in Figure~\ref{fig:cutout_comparison}
and report the detection AUCs in Table~\ref{tab:cutout_to_cutpaste}.
While achieving $71.3$ AUC that already is significantly better than random guessing, predicting a standard Cutout augmentation is still a simple task and the network may have learned a naive solution from easy proxy task, as discussed in Section~\ref{sec:method}. By gradually increasing the difficulty of proxy task to avoid known trivial solutions with random color to the patch, or with structures similar to local patterns of the normal data (Confetti noise, CutPaste), the network learns to find irregularity and generalizes better to detect real defects.

\begin{table}[t]
    \centering
    \caption{Detection AUCs of representations trained with binary classification between normal and the union of CutPaste and CutPaste-Scar examples and 3-way classification among normal, CutPaste and CutPaste-scar examples.}
    \label{tab:binary_vs_3way_vs_mix}
    \vspace{-0.1in}
    \resizebox{0.43\textwidth}{!}{
    \begin{tabular}{l|c|c|c|c}
        \toprule
        Category & CutPaste & CutPaste (scar) & Binary (Union) & 3-Way \\ \midrule
        texture &  91.7\scriptsize{$\pm0.7$} & 95.7\scriptsize{$\pm0.8$} & 97.3\scriptsize{$\pm0.3$} & 97.0\scriptsize{$\pm0.5$} \\
        object &  90.5\scriptsize{$\pm0.6$} & 92.4\scriptsize{$\pm0.8$} & 92.8\scriptsize{$\pm0.5$} & 94.3\scriptsize{$\pm0.6$} \\ \midrule
        all & 90.9\scriptsize{$\pm0.7$} & 93.5\scriptsize{$\pm0.8$} & 94.3\scriptsize{$\pm0.5$} & 95.2\scriptsize{$\pm0.6$} \\
        \bottomrule
    \end{tabular}
    }
    \vspace{-0.1in}
\end{table}

\vspace{-0.05in}
\subsection{Binary v.s. Finer-Grained Classification}
\label{sec:abl_2way_vs_3way}
In Table~\ref{tab:mvtec_main}, although CutPaste-scar shows better performance on average than CutPaste, there is no clear winner that works the best for all. As there are diverse types of defect in practice, we leverage the strength of both augmentations for representation learning.
In Section~\ref{sec:method_cutpaste_multiclass}, we train a model by solving a 3-way classification task between normal, CutPaste and CutPaste-scar. Alternatively, we train to solve a binary classification task by discriminating normal examples and the union of
two augmentations.

The results, along with those of representations trained with CutPaste and CutPaste-scar, are in Table~\ref{tab:binary_vs_3way_vs_mix}. It is clear that using both augmentations improve the performance. Between binary with union of augmentations and 3-way, we observe better detection performance with representations trained by 3-way classification task.
A plausible hypothesis on the superiority of 3-way formulation in our case is that it is more natural to model CutPaste and CutPaste-scar augmentations separately than together as there exists a systematic difference between them in the size, shape, and rotation angle of patches.

\subsection{CutPaste on Synthetic Anomaly Detection}
\label{sec:abl_synthetic}

\begin{figure}[t]
    \centering
    \includegraphics[width=0.48\textwidth]{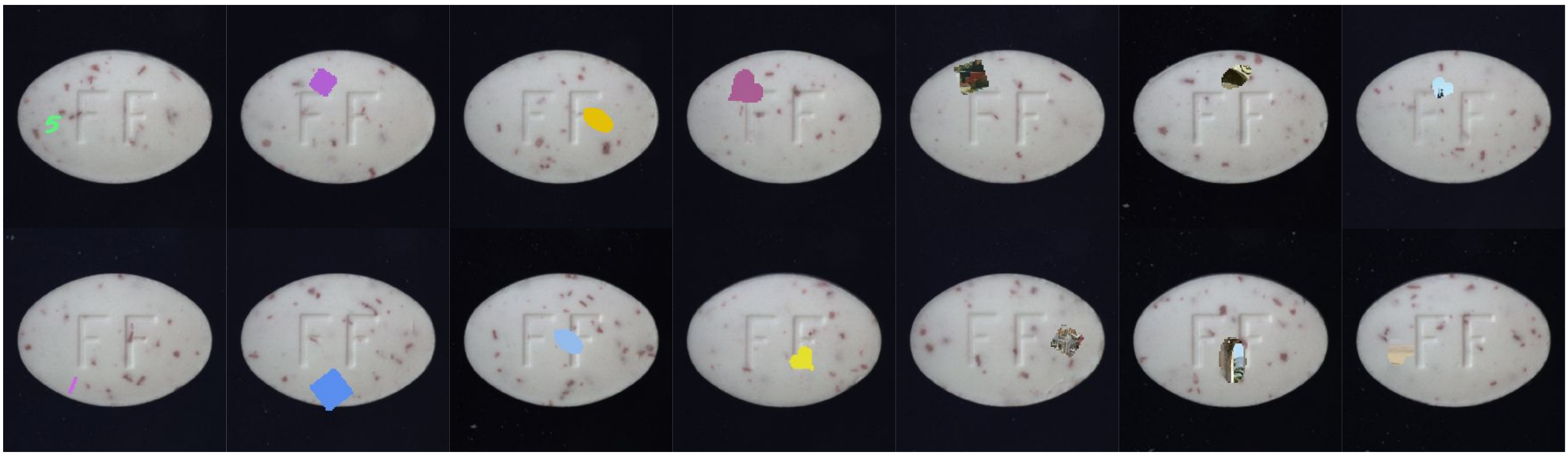}
    \vspace{-0.2in}
    \caption{Synthetic defects on pill class. From left to right, we use MNIST~\cite{lecun1998gradient}, square, ellipse, heart~\cite{matthey2017dsprites} with random color, and those filled with natural image patches.}
    \label{fig:synthetic_data}
    \vspace{-0.1in}
\end{figure}

We further study the generalization of our models to unseen anomalies. Specifically, we test on synthetic anomaly datasets created by patching diverse shape masks to normal data, such as digits~\cite{lecun1998gradient}, square, ellipse, or heart~\cite{matthey2017dsprites}, filled with random color or natural images. Samples of synthetic anomalies are shown in Figure~\ref{fig:synthetic_data} and detection results are in Table~\ref{tab:exp_synthetic}.
We first note that these datasets are not trivial -- a model trained by predicting Cutout augmentations achieves only $81.5$. Our proposed CutPaste (3-way) model performs well on synthetic dataset, achieving $98.3$ AUC on average. 
We highlight that some shapes (e.g., ellipse, heart) or color statistics inside the patch (e.g., constant color, natural images) are not seen at training, but we can still generalize to these unseen cases.

\begin{table}[t]
    \centering
    \caption{Detection AUCs on synthetic data. Various shapes, such as digit, square, ellipse, or heart, are patched to normal images with random color or natural images ($^{\dagger}$). }
    \label{tab:exp_synthetic}
    \vspace{-0.1in}
    \resizebox{0.48\textwidth}{!}{
    \begin{tabular}{c|c|c|c|c|c|c|c||c}
        \toprule
        Dataset & MNIST & Square & Ellipse & Heart & Square$^{\dagger}$ & Ellipse$^{\dagger}$ & Heart$^{\dagger}$ & Avg \\ \midrule
        Cutout & 52.3 & 90.6 & 89.3 & 87.5 & 86.4 & 84.0 & 80.7 & 81.5 \\
        CutPaste & 96.1 & 98.4 & 98.2 & 97.9 & 99.3 & 99.2 & 99.0 & 98.3 \\
        \bottomrule
    \end{tabular}
    }
    \vspace{-0.1in}
\end{table}

\subsection{Application to Semantic Outlier Detection}
\label{sec:abl_semantic}
We also conduct the semantic anomaly detection experiment on CIFAR-10~\cite{krizhevsky2009learning} following the protocol in~\cite{golan2018deep,sohn2020learning}, where a single class is treated as normal and remaining 9 classes are anomalies. We make a comparison of Cutout, CutPaste and rotation prediction~\cite{sohn2020learning}. Cutout results in $60.2$ AUC, and CutPaste achieves $69.4$ AUC, which significantly improves upon Cutout ($60.2$). However, these are still far behind that of rotation prediction ($91.3$ AUC) on CIFAR-10 semantic anomaly detection.
On the other hand, in Section~\ref{sec:exp_main}, we have discussed the reversed situation that rotation prediction is much worse than 3-way CutPaste prediction. 
The results suggest the difference between semantic anomaly detection and defect detection, which needs different algorithm and augmentation designs.

\section{Conclusion}
\label{sec:concl}
We propose a data-driven approach for defect detection and localization. The key to our success is self-supervised learning of representations with CutPaste, a simple yet effective augmentation that encourages the model to find local irregularities. We show superior image-level anomaly detection performance on the real-world dataset. Furthermore, by learning and extracting patch-level representations, we demonstrate state-of-the-art pixel-wise anomaly localization performance. We envision the CutPaste augmentation could be a cornerstone for building a powerful model for semi-supervised and unsupervised defect detection.

\paragraph{Acknowledgment.}
We thank Yang Feng for sharing the implementation of uninformed student and Sercan Arik for the proofread of our manuscript.

\newpage
{\small
\bibliographystyle{ieee_fullname}
\bibliography{egbib}
}

\newpage
\appendix

\onecolumn

\section{Details on Experiments}
\label{sec:app_detail_exp}

\subsection{Experiment with ResNet-18}
\label{sec:exp_setting_resnet}
We train a model on $256{\times}256$ image. Model parameters are updated for $65$k steps using momentum SGD with the learning rate of $0.03$, momentum of $0.9$, and the batch size of 64 (or 96 for 3-way). A single cycle of cosine learning rate decay schedule~\cite{loshchilov2016sgdr} and L2 weight regularization with a coefficient of $0.00003$ are used. We apply random translation and color jitters for data augmentation to enhance invariance of representations. We note that the same training strategy, such as the selection of hyperparameters or data augmentations, is applied to different categories. Full range of hyperparameters we studied are provided in below. We use Tensorflow~\cite{abadi2016tensorflow}, and scikit-learn~\cite{scikit-learn} for GDE implementation.

\subsection{Implementation details on CutPaste}
\label{sec:exp_cutpaste}
Our implementation of the CutPaste augmentation closely follows that of RandomErasing\footnote{\url{https://pytorch.org/docs/stable/_modules/torchvision/transforms/transforms.html##RandomErasing}}~\cite{zhong2020random}.
First, we determine the size of the patch by sampling the area ratio between the patch and the full image from $(0.02, 0.15)$. We then determine the aspect ratio by sampling from $(0.3,1)\,{\cup}\, (1, 3.3)$. The locations of the patch where we cut from and paste to are randomly selected in a way that the entire patch appears in the full image.
To construct CutPaste-Scar, we directly sample the width between $[2, 16]$ and the length between $[10, 25]$ in terms of number of pixels. 
CutPaste-Scar is randomly rotated between $(-45, 45)$ degrees. Before pasting, we apply color jitter,\footnote{\url{https://pytorch.org/docs/stable/_modules/torchvision/transforms/transforms.html##ColorJitter}} which applies brightness, contrast, saturation, and hue transformations in sequence with random order, with the maximum intensity of $0.1$.

\subsection{Ablation Study on Hyperparameters}
\label{sec:exp_hparams}

We study the impact of different values for optimization hyperparameters. Furthermore, we study the impact of CutPaste hyperparameters, such as jitter intensity or the size of the patch. All experiments are done on CutPaste 3-way setting. Below, we enumerate the ranges for hyperparameters we studied. The bold-faced items are default values used for experiments in the main text as well as in this section. 

\begin{enumerate}
    \setlength\itemsep{0em}
    \item Learning rates ${\in}\{0.1, \mathbf{0.03}, 0.01, 0.003\}$.
    \item Number of training epochs ${\in}\{128,192,\mathbf{256},320,384\}$.\footnote{Note that, unlike conventional definition for an epoch, we define $256$ parameter update steps as one epoch.}
    \item Maximum jitter intensity on patch ${\in}\{0, \mathbf{0.1}, 0.2, 0.3\}$.
    \item Maximum size of patch ${\in}\{0.1, \mathbf{0.15}, 0.2, 0.3\}$.
\end{enumerate}

We report the mean and standard error of detection AUCs in Table~\ref{tab:sensitivity_anal_lr_epoch} and \ref{tab:sensitivity_anal_cutpaste}. We observe that our method is fairly robust across different learning rates and number of epochs. When the learning rate becomes too small ($\leq0.003$) we observe slow convergence, so it suggests to train longer. The number of training epochs is also an important hyperparameter for semi-supervised anomaly detection as early stopping is particularly difficult. We observe that our method provides a reliable solution when trained for different number of epochs.

For jitter intensity on patch of the CutPaste augmentation, we find it more important for texture categories. Part of the reasons is that the CutPaste augmentation, when applied without jitter augmentation, makes it too difficult to distinguish from the original images, as texture categories contain repetitive patterns. By adding jitter on the patch, the contrast between pasted patch and the surrounding area becomes more apparent and this makes the CutPaste prediction task a bit easier. Similarly, the size of the patch mostly affects the performance on texture categories and we observe the method prefers generally smaller patch sizes.

\begin{table}[t]
    \centering
    \caption{Detection AUCs using (1) different learning rates and (2) the number of epochs. We report the mean and standard error of AUCs for 5 runs with different random seeds. }
    \label{tab:sensitivity_anal_lr_epoch}
    \resizebox{0.8\textwidth}{!}{
    \begin{tabular}{l|c|c|c|c||c|c|c|c|c}
        \toprule
        \multirow{2}{*}{Category}& \multicolumn{4}{c||}{Learning rates} & \multicolumn{5}{c}{Number of epochs} \\
        \cmidrule{2-10}
         & 0.1 & 0.03 & 0.01 & 0.003 & 128 & 192 & 256 & 320 & 384 \\ \midrule
        texture & 97.1\scriptsize{$\pm$0.3} & 97.0\scriptsize{$\pm$0.5} & 97.2\scriptsize{$\pm$0.3} &96.1\scriptsize{$\pm$0.7} & 96.6\scriptsize{$\pm$0.4} & 96.1\scriptsize{$\pm$0.7} & 97.0\scriptsize{$\pm$0.5} & 97.0\scriptsize{$\pm$0.4} & 96.3\scriptsize{$\pm$0.4} \\
        object & 94.4\scriptsize{$\pm$0.6} & 94.3\scriptsize{$\pm$0.6} & 94.2\scriptsize{$\pm$0.6} &93.9\scriptsize{$\pm$0.5} &94.9\scriptsize{$\pm$0.6} & 94.5\scriptsize{$\pm$0.4} & 94.3\scriptsize{$\pm$0.6} & 94.7\scriptsize{$\pm$0.5} & 94.0\scriptsize{$\pm$0.6} \\ \midrule
        all & 95.3\scriptsize{$\pm$0.5} & 95.2\scriptsize{$\pm$0.6} & 95.2\scriptsize{$\pm$0.5} &94.6\scriptsize{$\pm$0.6} & 95.4\scriptsize{$\pm$0.5} & 95.0\scriptsize{$\pm$0.5} & 95.2\scriptsize{$\pm$0.6} & 95.5\scriptsize{$\pm$0.4} & 94.8\scriptsize{$\pm$0.5} \\
        \bottomrule
    \end{tabular}
    }
    \vspace{0.1in}
    \centering
    \caption{Detection AUCs using (1) different jitter intensity and (2) the size of patch of the CutPaste augmentation. We report the mean and standard error of AUCs for 5 runs with different random seeds.}
    \label{tab:sensitivity_anal_cutpaste}
    \resizebox{0.72\textwidth}{!}{
    \begin{tabular}{l|c|c|c|c||c|c|c|c}
        \toprule
        \multirow{2}{*}{Category}& \multicolumn{4}{c||}{Jitter intensity} & \multicolumn{4}{c}{Size of patch} \\
        \cmidrule{2-9}
         & 0.0 & 0.1 & 0.2 & 0.3 & 0.1 & 0.15 & 0.2 & 0.3 \\ \midrule
        texture & 96.2\scriptsize{$\pm$0.6} & 97.0\scriptsize{$\pm$0.5} & 97.4\scriptsize{$\pm$0.3} &97.5\scriptsize{$\pm$0.2} & 97.1\scriptsize{$\pm$0.4} &97.0\scriptsize{$\pm$0.5} &95.8\scriptsize{$\pm$0.9} &96.6\scriptsize{$\pm$0.4} \\
        object & 94.3\scriptsize{$\pm$0.6} & 94.3\scriptsize{$\pm$0.6} & 94.5\scriptsize{$\pm$0.5} &94.5\scriptsize{$\pm$0.5} & 94.5\scriptsize{$\pm$0.5} &94.3\scriptsize{$\pm$0.6} &94.4\scriptsize{$\pm$0.5} &94.5\scriptsize{$\pm$0.5} \\ \midrule
        all & 94.9\scriptsize{$\pm$0.6} & 95.2\scriptsize{$\pm$0.6} &95.5\scriptsize{$\pm$0.4} &95.5\scriptsize{$\pm$0.4} &95.3\scriptsize{$\pm$0.5} &95.2\scriptsize{$\pm$0.6} &94.8\scriptsize{$\pm$0.6} &95.2\scriptsize{$\pm$0.5} \\
        \bottomrule
    \end{tabular}
    }
\end{table}

\subsection{Experiment with EfficientNet}
We follow the Keras guide\footnote{\url{https://keras.io/examples/vision/image_classification_efficientnet_fine_tuning/}} of fine-tuning EfficientNet. 
We use EfficientNet B4 with batch size 24, which is the limit we tried to fit into single GPU training. We first train only the MLP head for 10 epochs with learning rate of $0.03$ while freezing the pretrained backbone. We then fine-tune all layers for 64 epochs with learning rate of $0.0001$. We note that the batchnorm layers are kept frozen as suggested by the Keras guide. The other unstated hyperparameters are the same as in Section~\ref{sec:exp_setting_resnet}.

\subsection{Details on Localization with Patch-based model}

The training procedure of patch-based model with CutPaste should be straightforward from Section~\ref{sec:method_patch}. 
Once patch-based models are trained, we densely extract representations of $32{\times}32$ patches with the stride of $4$, which results in $(\frac{256-32}{4}+1){\times}(\frac{256-32}{4}+1){\times}512\,{=}\,57{\times}57{\times}512$ dimensional tensor of embedding vectors.
Then we compute the anomaly score of $512$ dimensional embedding vector at each location to obtain $57{\times}57$ anomaly score map. Finally, to obtain an anomaly score map of full resolution ($256{\times}256$), we apply receptive field upsampling via Gaussian smoothing following~\cite{liznerski2020explainable}, which essentially applies the transposed convolution with the stride of $4$, the same stride that we used for dense feature extraction, using a single convolution kernel of size $32{\times}32$ whose weights are determined by a Gaussian distribution.

Depending on the category, we find different strategies for computing score improves the performance. For those categories with ``aligned objects'' (bottle, cable, capsule, metal nut, pill, toothbrush, transistor, zipper), we find it useful to construct one-class classifiers at each location separately. This is particularly useful for detecting missing or dislocated components (e.g., examples in the second row of Figure~\ref{fig:heatmap_transistor}) as global context is captured by classifiers at each location. For some object categories, such as hazelnut or screw whose objects are randomly rotated, and texture categories, we use a single one-class classifier applied to all locations.

%
%


\section{More Localization Visualizations}
\label{sec:app_localization}

\begin{figure}
    \centering
    \includegraphics[width=0.85\textwidth]{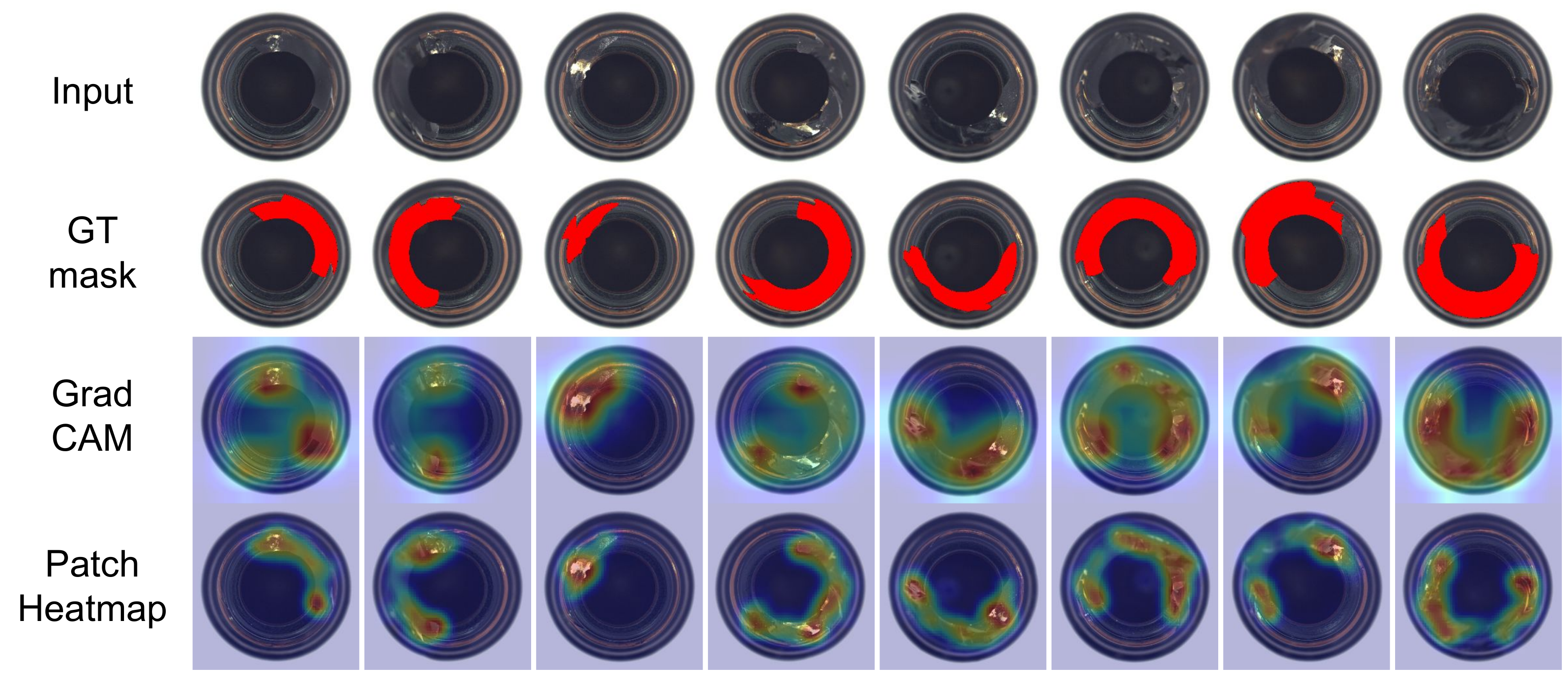}\\
    \vspace{0.05in}
    \includegraphics[width=0.85\textwidth]{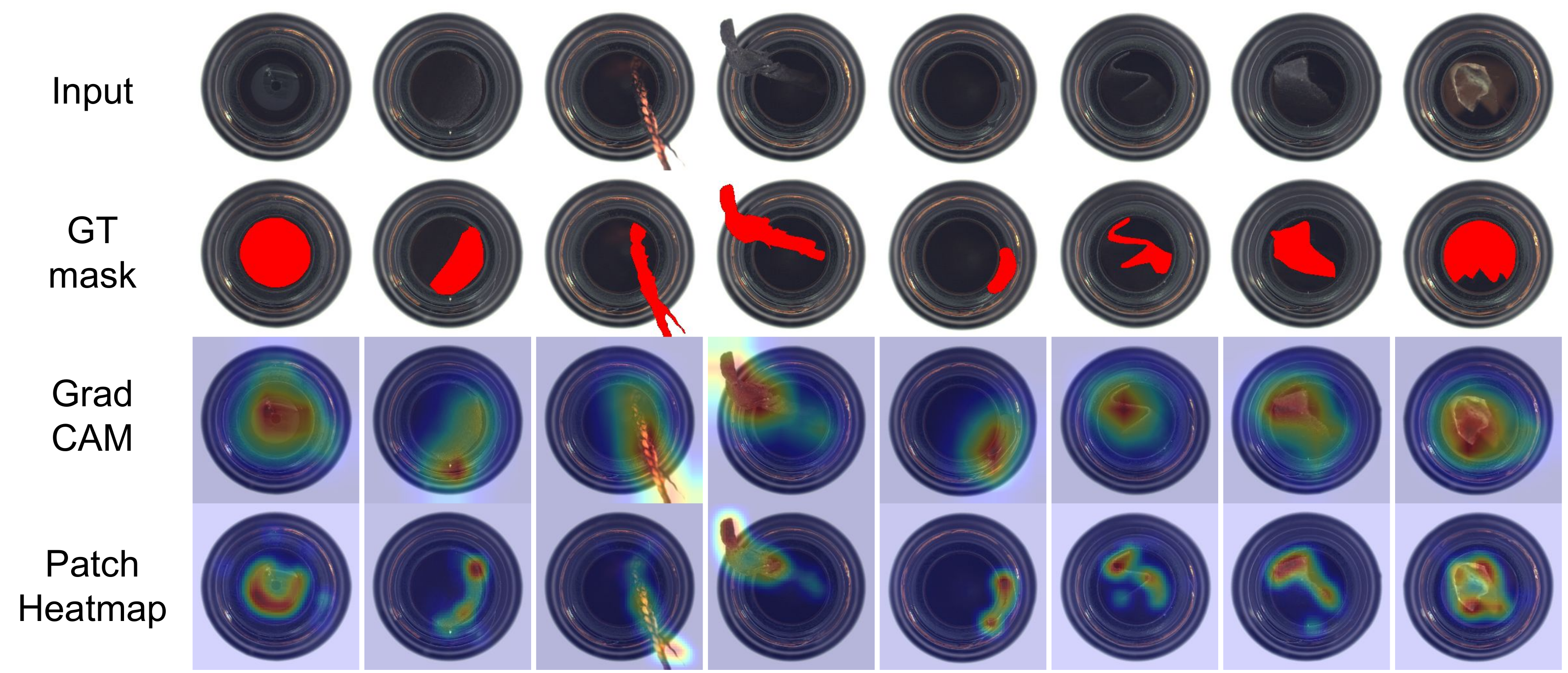}\\
    \vspace{0.05in}
    \includegraphics[width=0.85\textwidth]{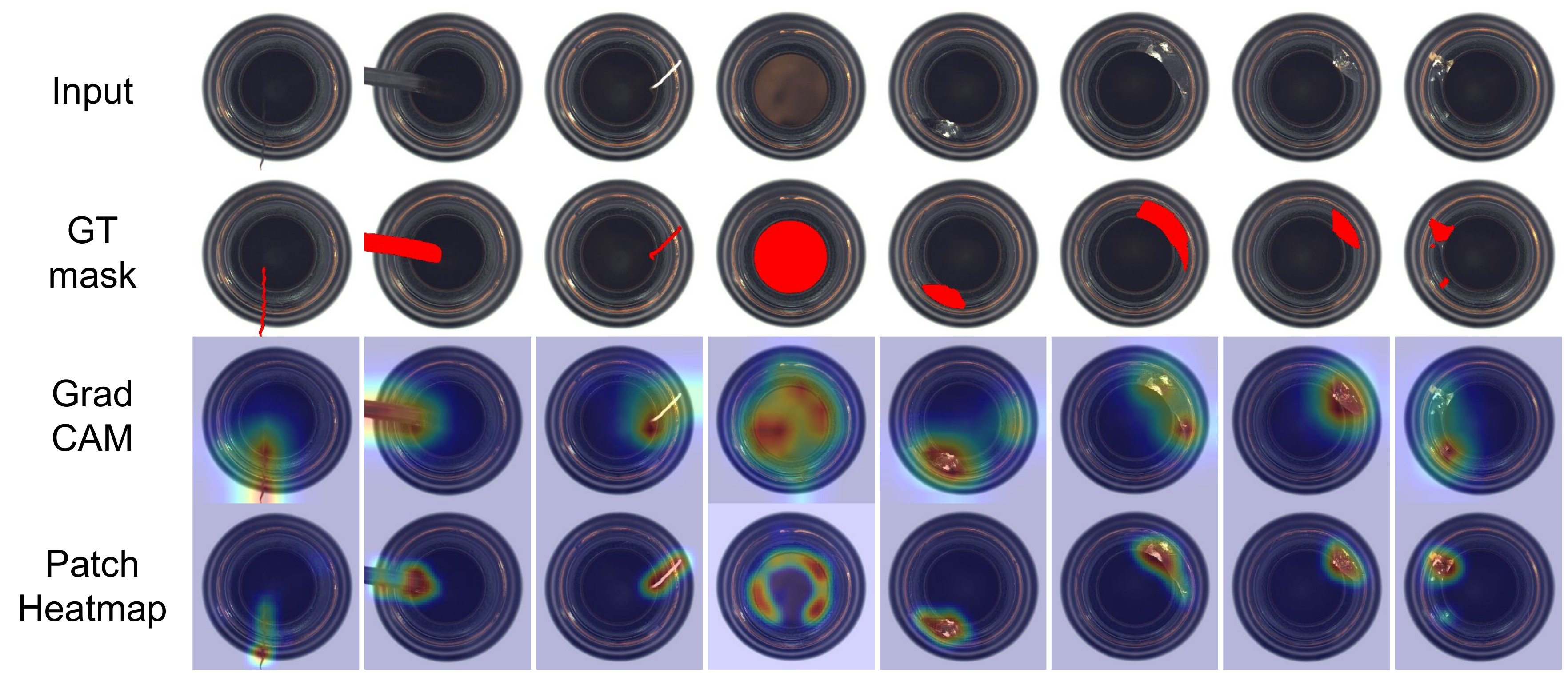}
    \caption{Defect localization on bottle class of MVTec dataset. From top to bottom, input images, those with ground-truth localization mask in red, GradCAM results using image-level detector, and heatmaps using patch-level detector.}
    \label{fig:heatmap_bottle}
\end{figure}

\begin{figure}
    \centering
    \includegraphics[width=0.85\textwidth]{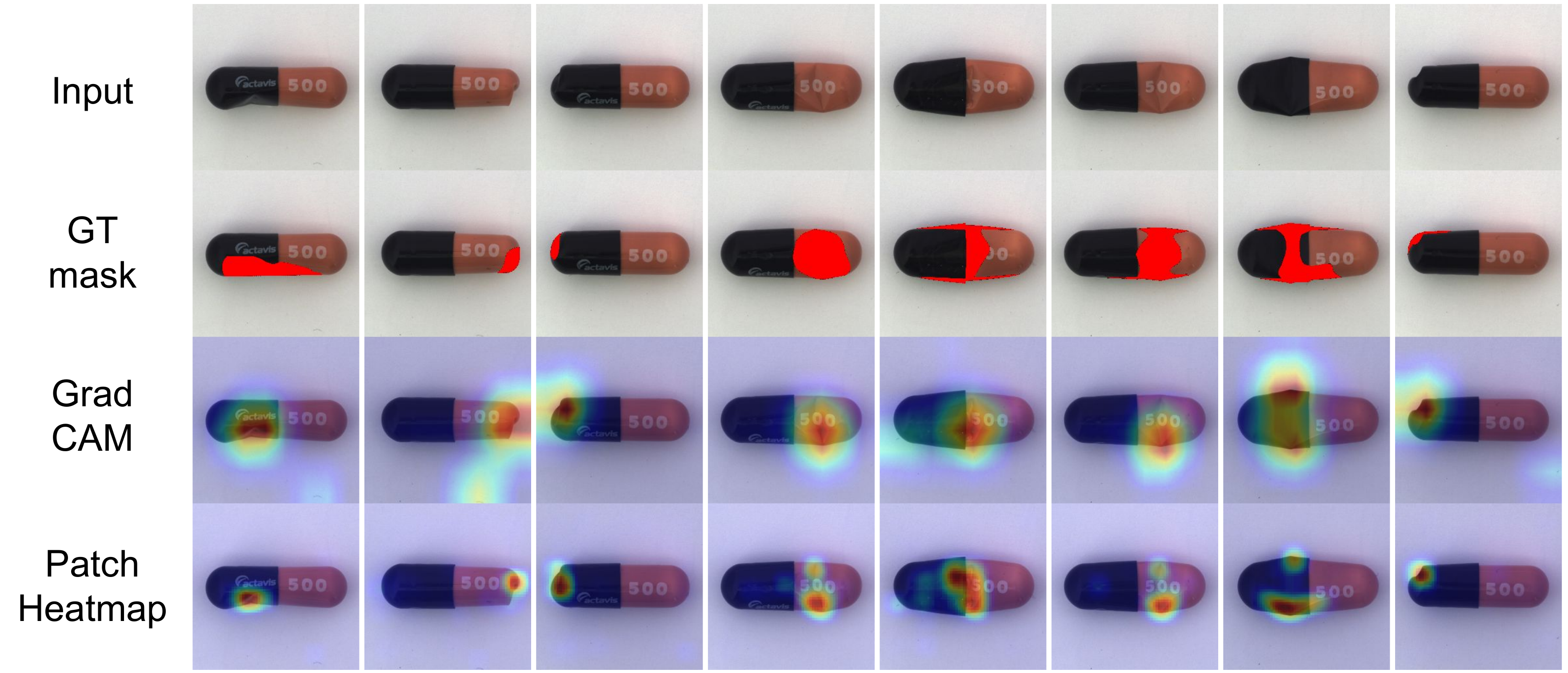}\\
    \vspace{0.05in}
    \includegraphics[width=0.85\textwidth]{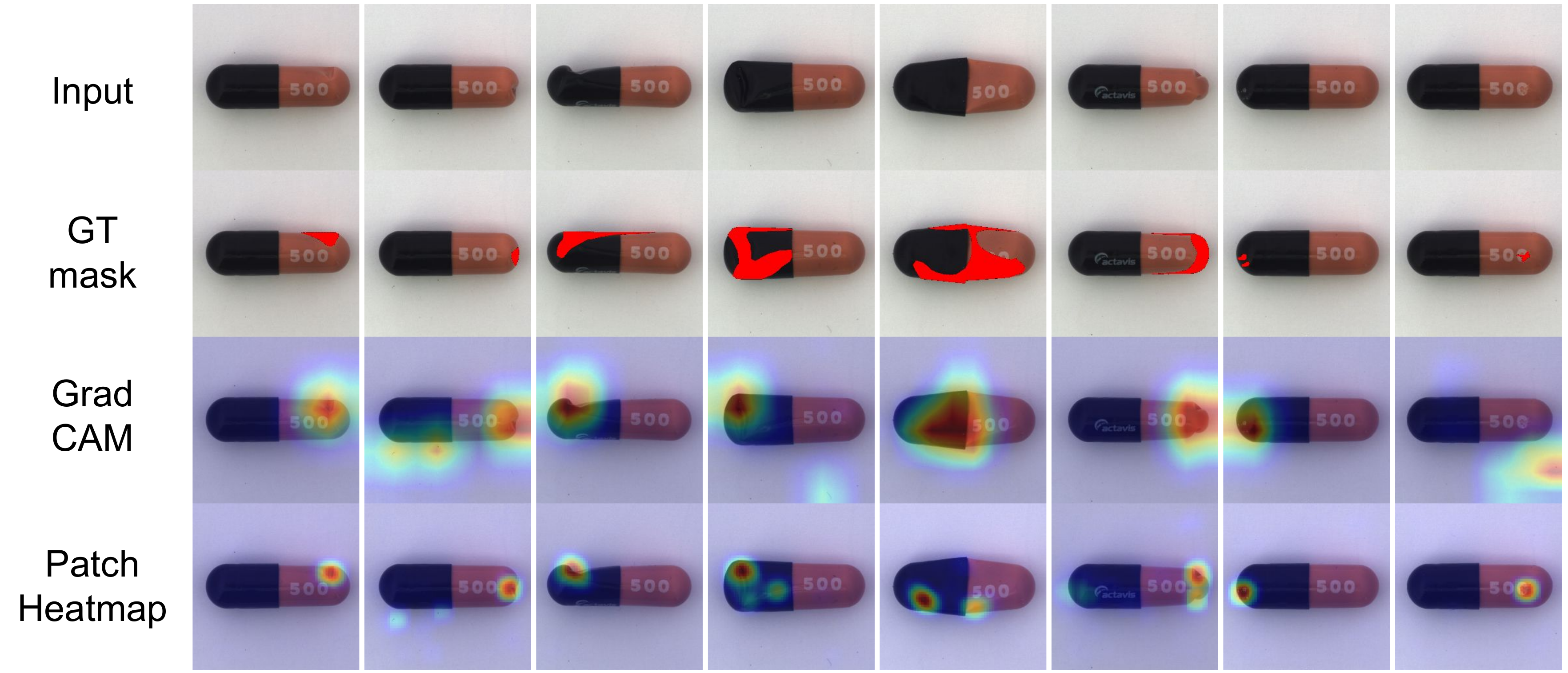}\\
    \vspace{0.05in}
    \includegraphics[width=0.85\textwidth]{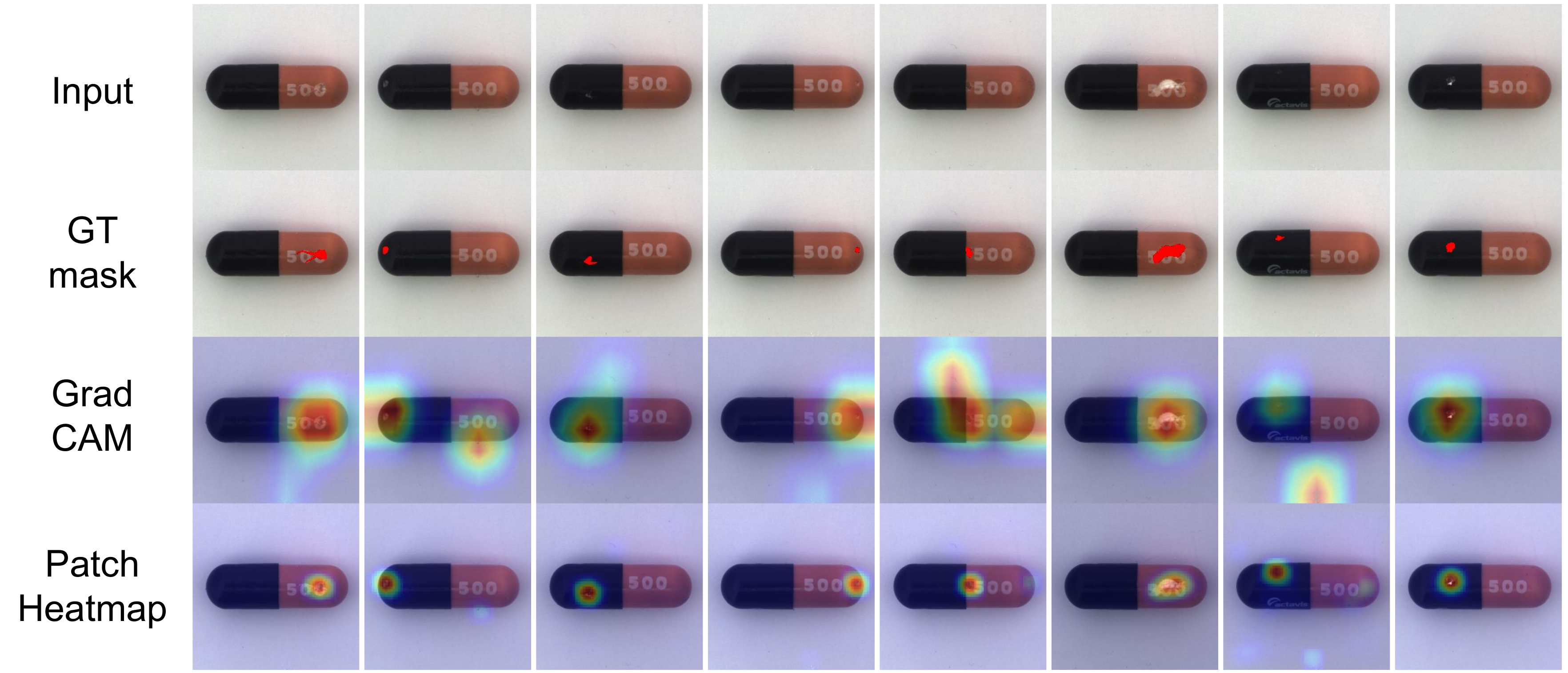}
    \caption{Defect localization on capsule class of MVTec dataset. From top to bottom, input images, those with ground-truth localization mask in red, GradCAM results using image-level detector, and heatmaps using patch-level detector.}
    \label{fig:heatmap_capsule}
\end{figure}

\begin{figure}
    \centering
    \includegraphics[width=0.85\textwidth]{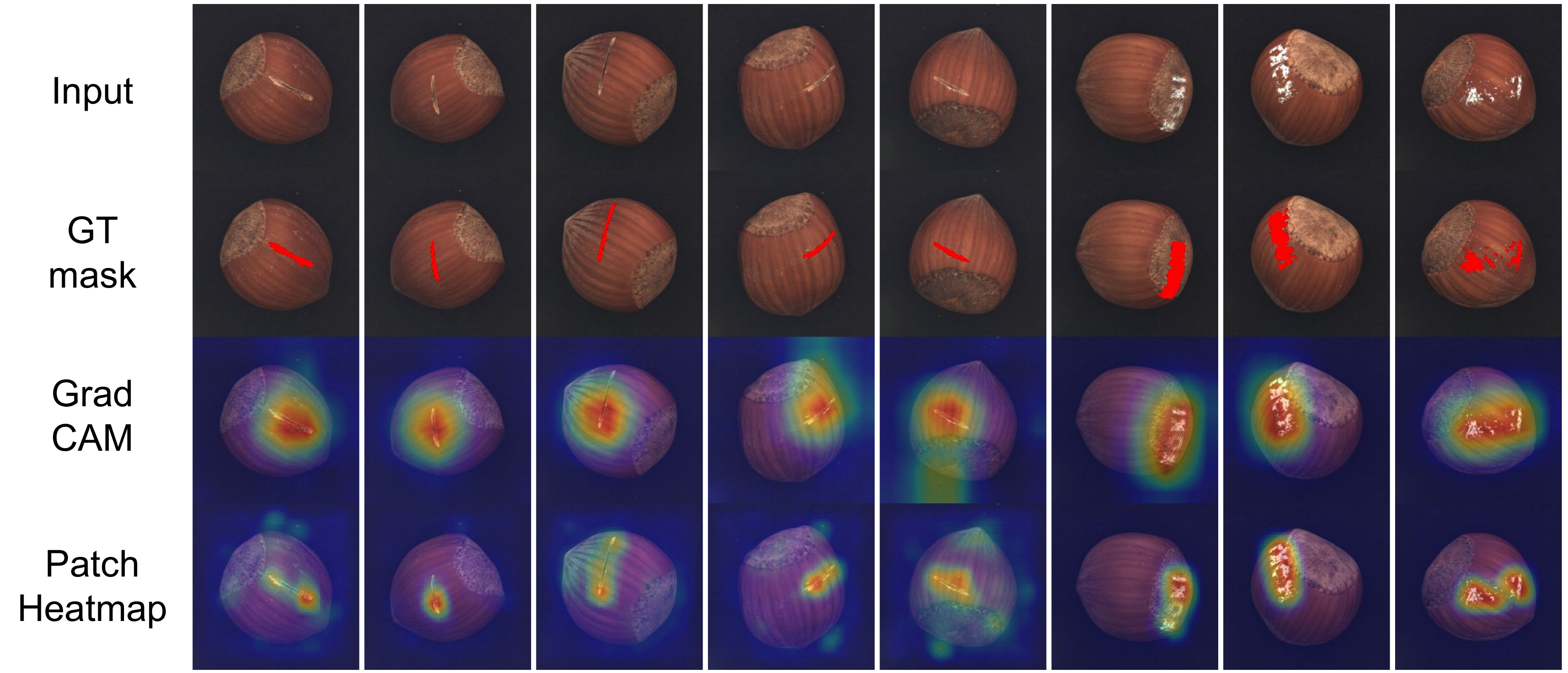}\\
    \vspace{0.05in}
    \includegraphics[width=0.85\textwidth]{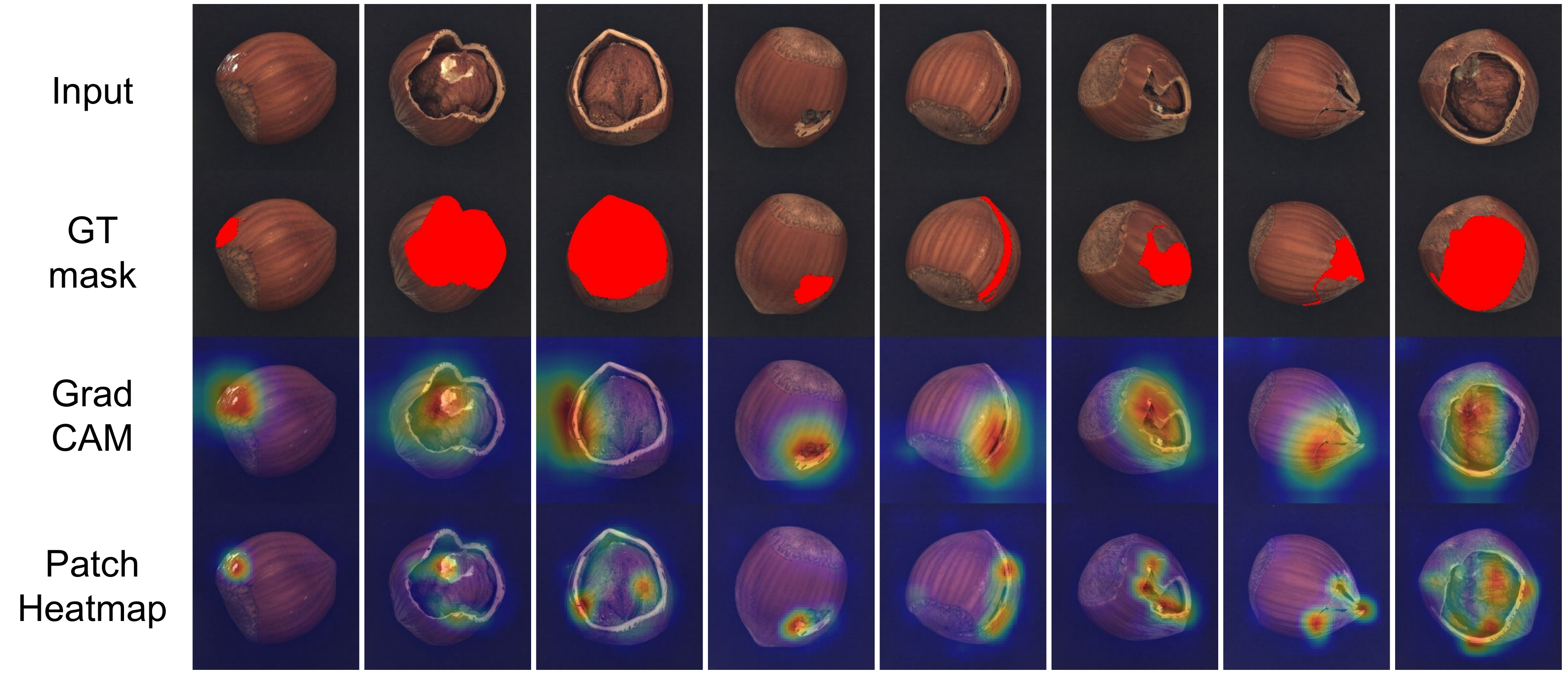}\\
    \vspace{0.05in}
    \includegraphics[width=0.85\textwidth]{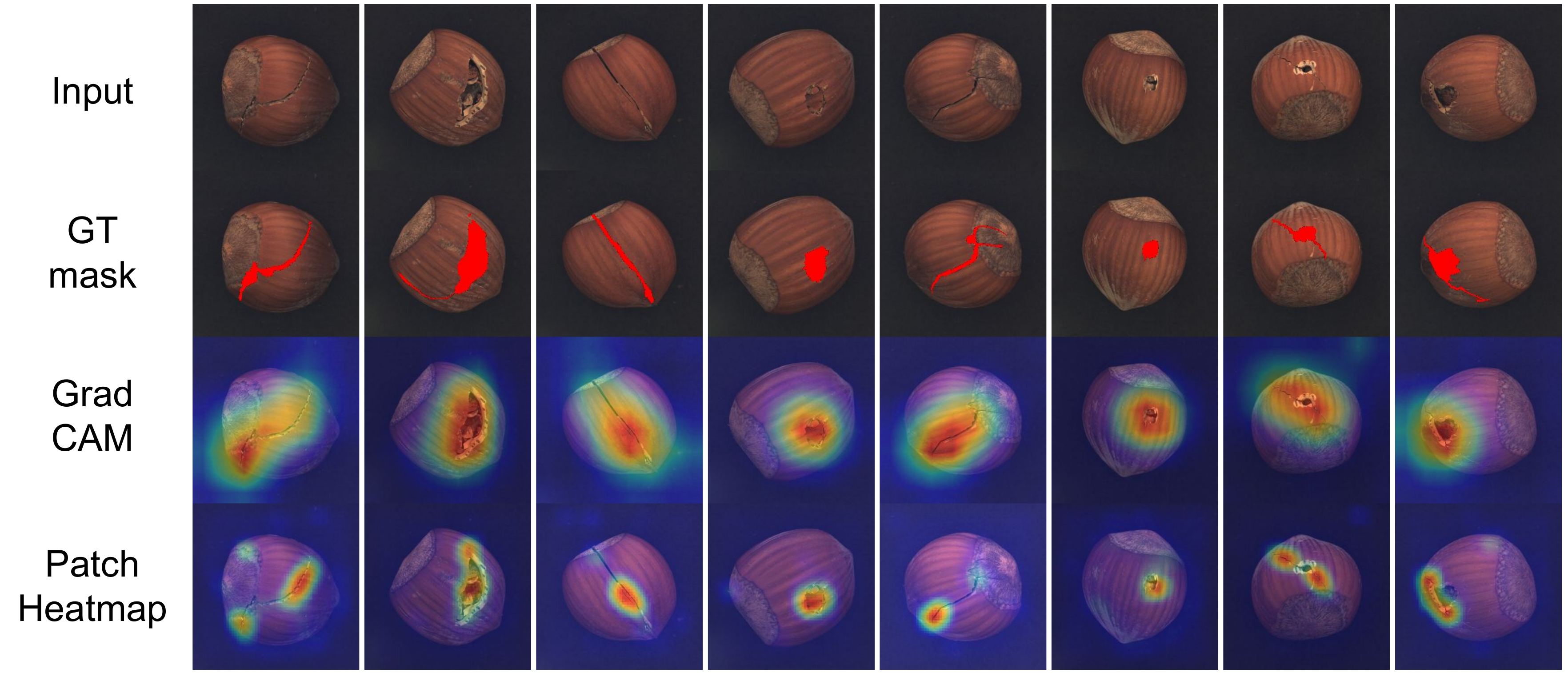}
    \caption{Defect localization on hazelnut class of MVTec dataset. From top to bottom, input images, those with ground-truth localization mask in red, GradCAM results using image-level detector, and heatmaps using patch-level detector.}
    \label{fig:heatmap_hazelnut}
\end{figure}

\begin{figure}
    \centering
    \includegraphics[width=0.85\textwidth]{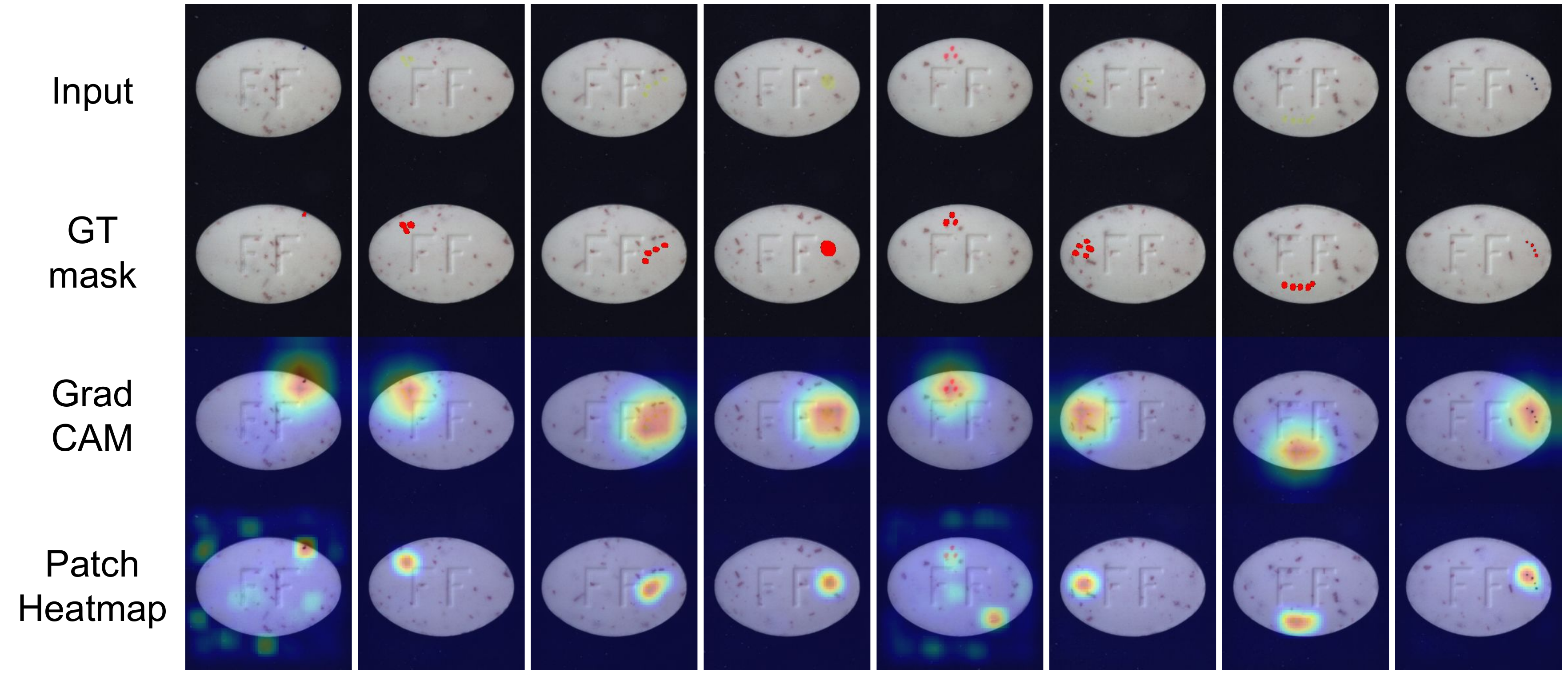}\\
    \vspace{0.05in}
    \includegraphics[width=0.85\textwidth]{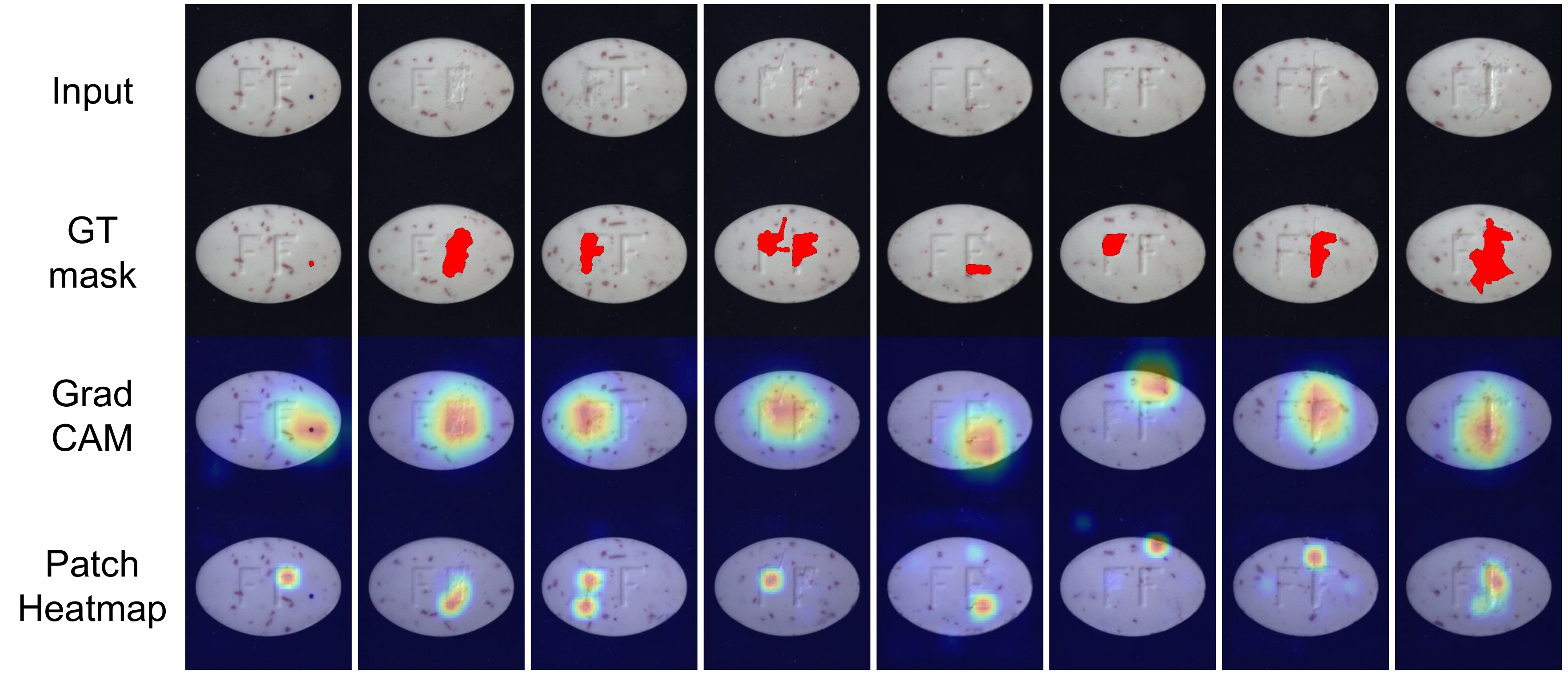}\\
    \vspace{0.05in}
    \includegraphics[width=0.85\textwidth]{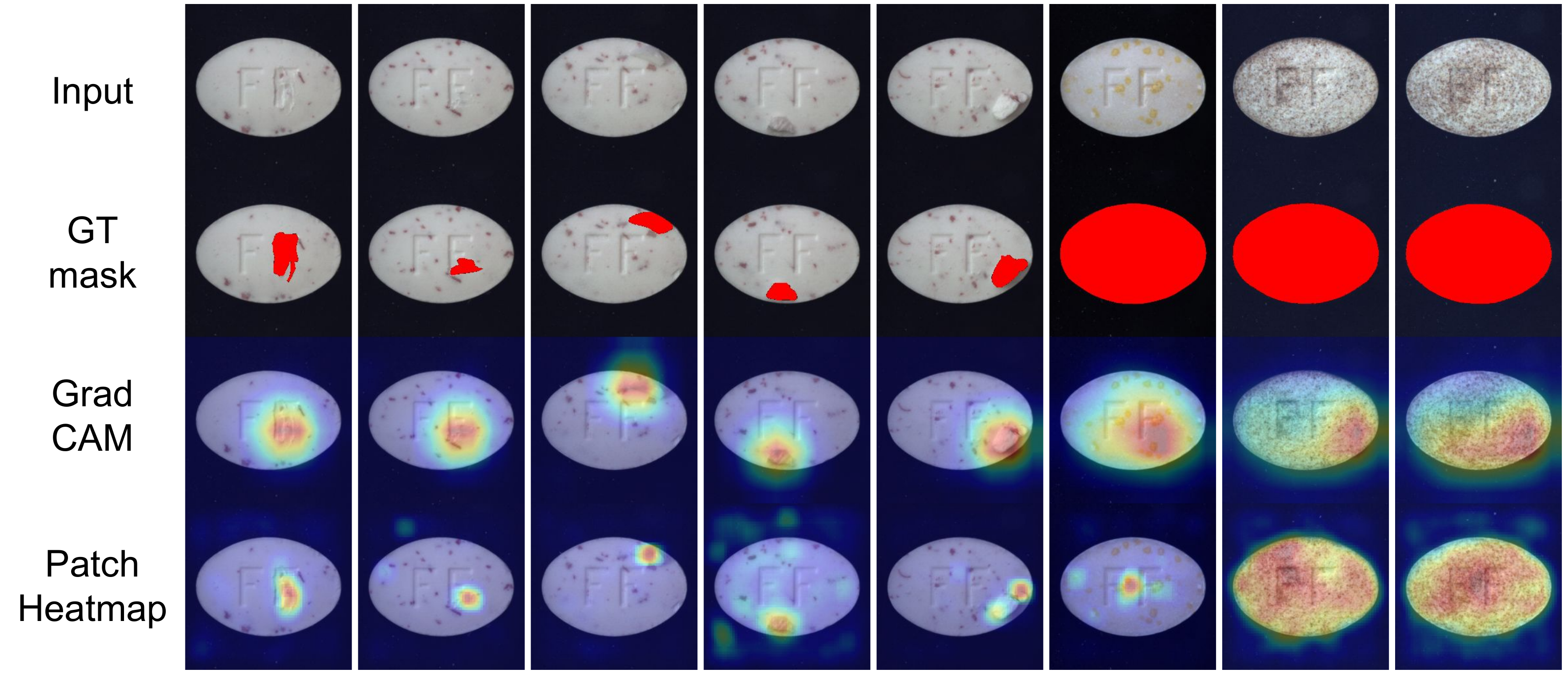}
    \caption{Defect localization on pill class of MVTec dataset. From top to bottom, input images, those with ground-truth localization mask in red, GradCAM results using image-level detector, and heatmaps using patch-level detector.}
    \label{fig:heatmap_pill}
\end{figure}

\begin{figure}
    \centering
    \includegraphics[width=0.85\textwidth]{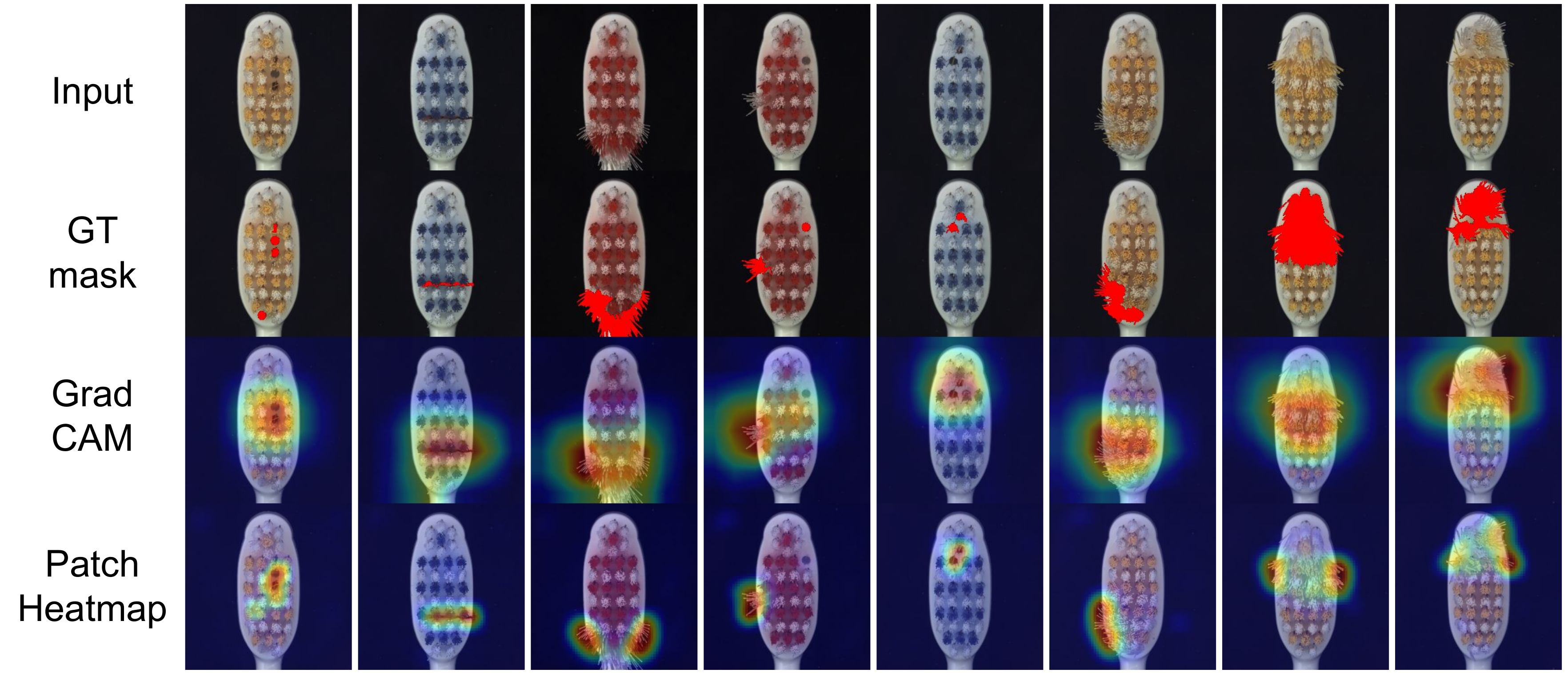}\\
    \vspace{0.05in}
    \includegraphics[width=0.85\textwidth]{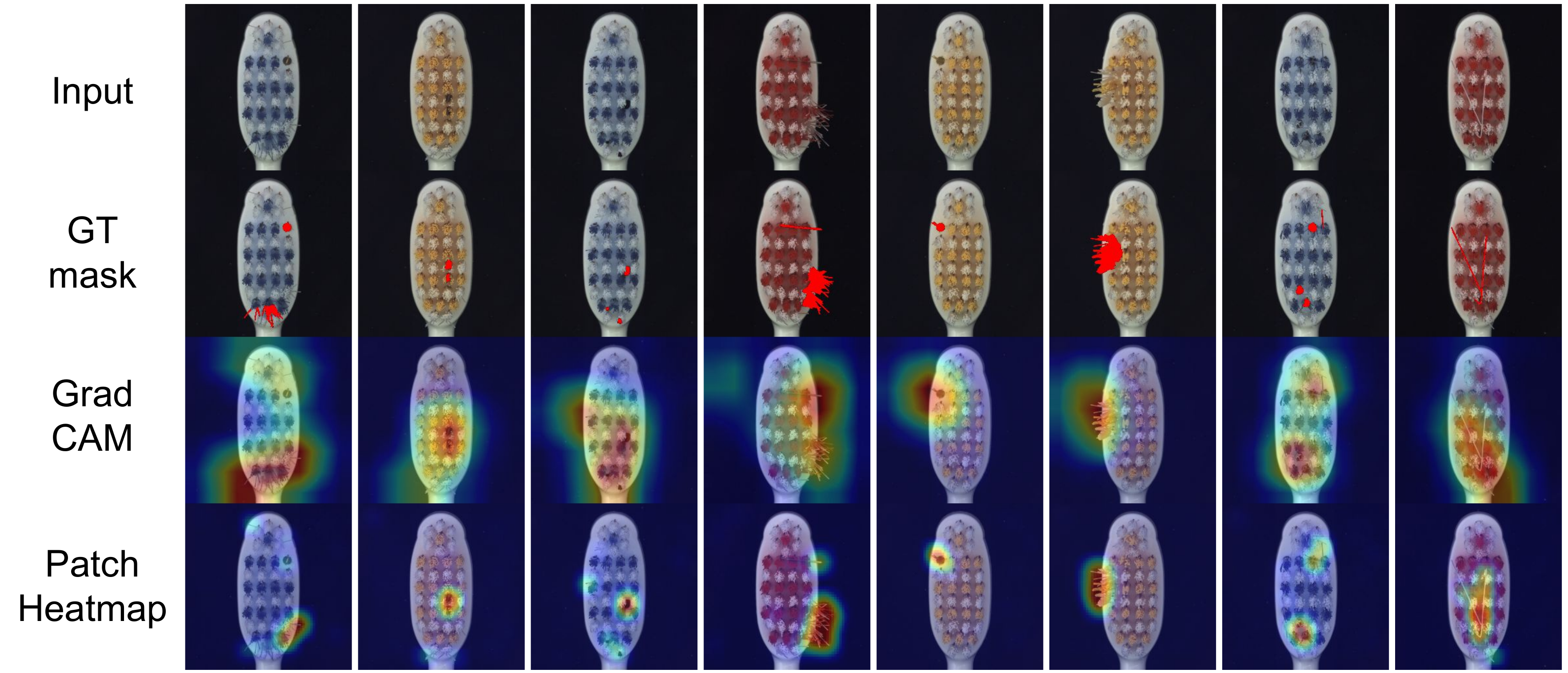}\\
    \vspace{0.05in}
    \includegraphics[width=0.85\textwidth]{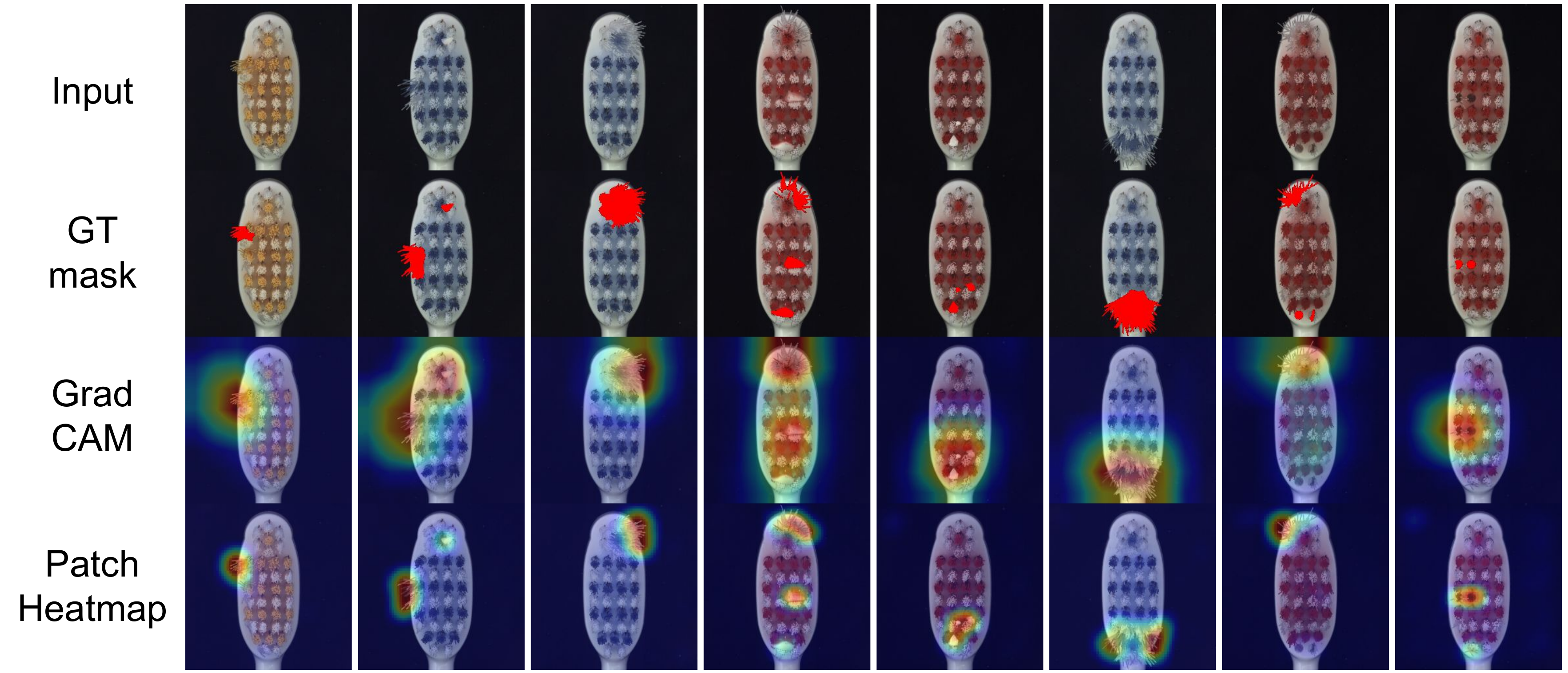}
    \caption{Defect localization on toothbrush class of MVTec dataset. From top to bottom, input images, those with ground-truth localization mask in red, GradCAM results using image-level detector, and heatmaps using patch-level detector.}
    \label{fig:heatmap_toothbrush}
\end{figure}

\begin{figure}
    \centering
    \includegraphics[width=0.85\textwidth]{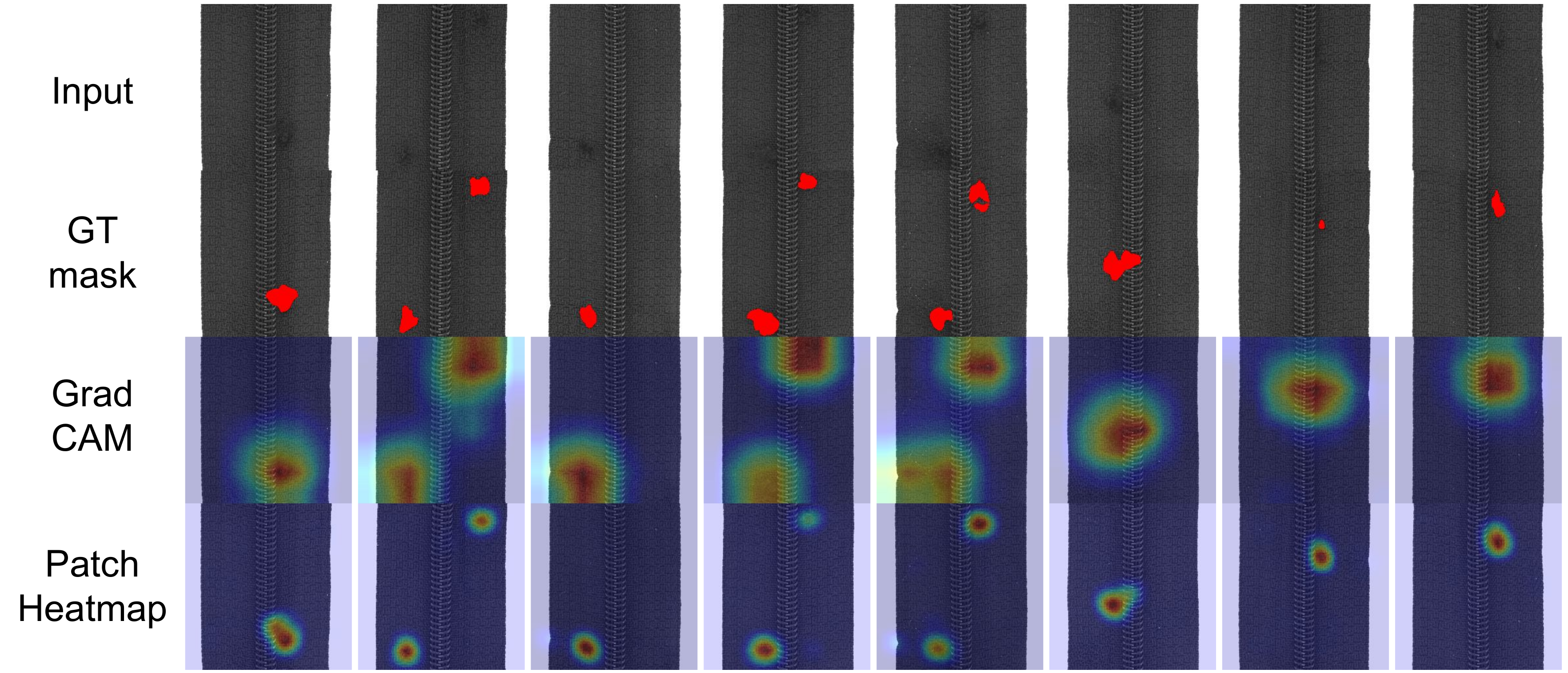}\\
    \vspace{0.05in}
    \includegraphics[width=0.85\textwidth]{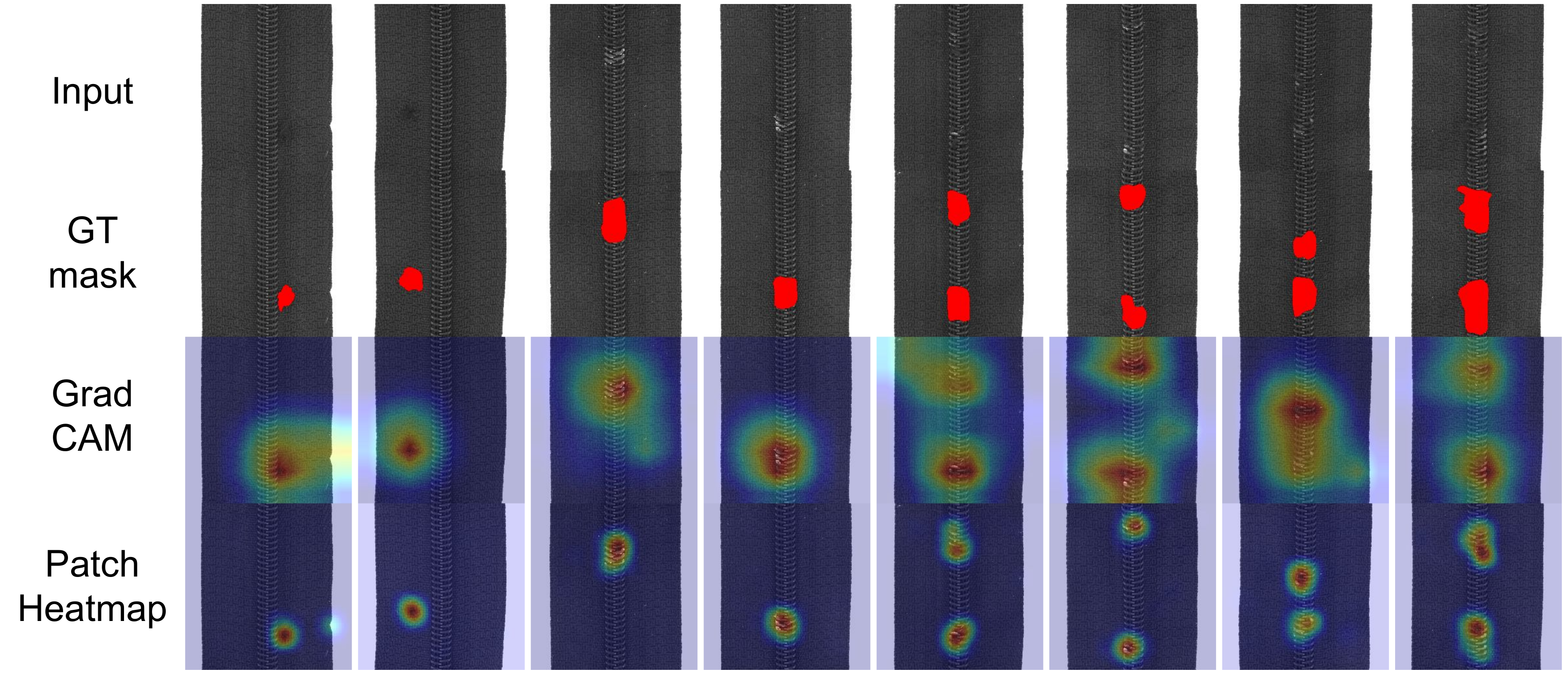}\\
    \vspace{0.05in}
    \includegraphics[width=0.85\textwidth]{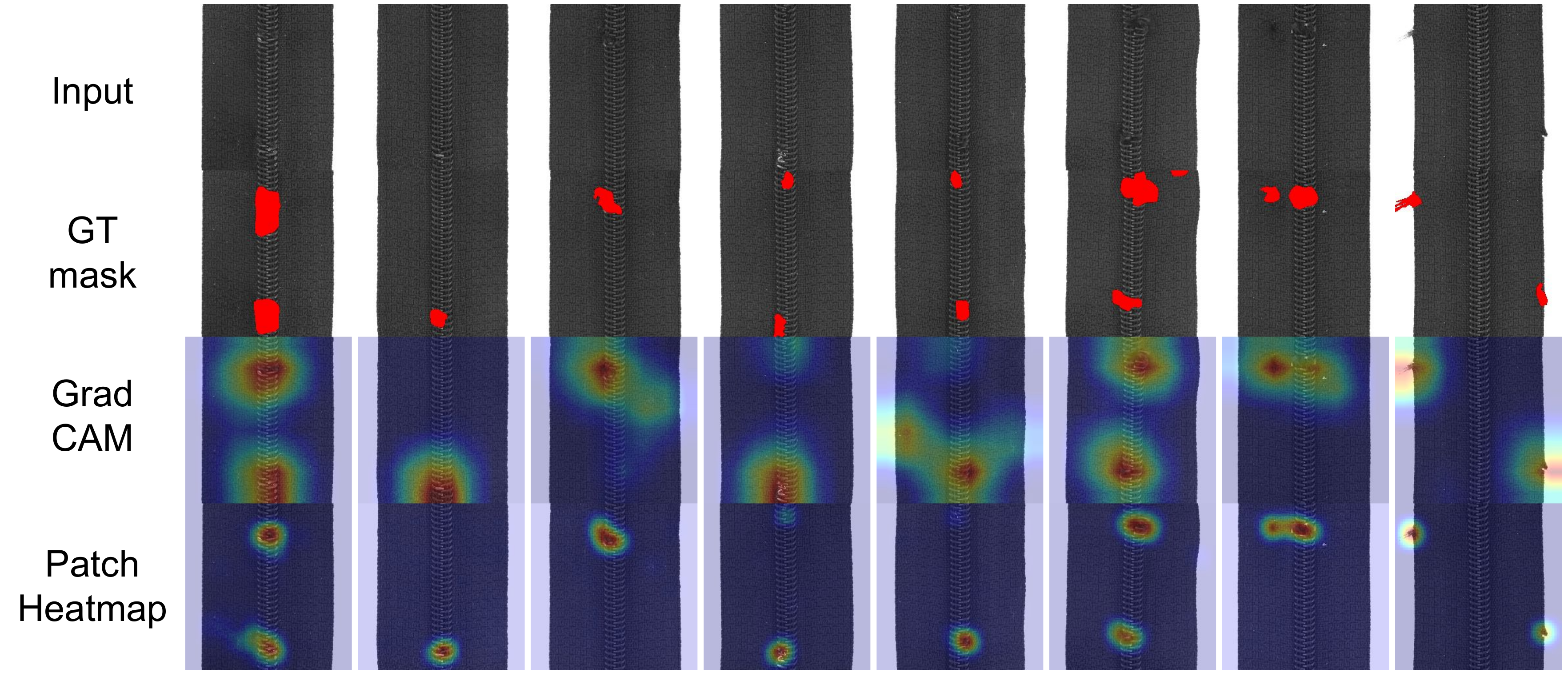}
    \caption{Defect localization on zipper class of MVTec dataset. From top to bottom, input images, those with ground-truth localization mask in red, GradCAM results using image-level detector, and heatmaps using patch-level detector.}
    \label{fig:heatmap_zipper}
\end{figure}

\begin{figure}
    \centering
    \includegraphics[width=0.85\textwidth]{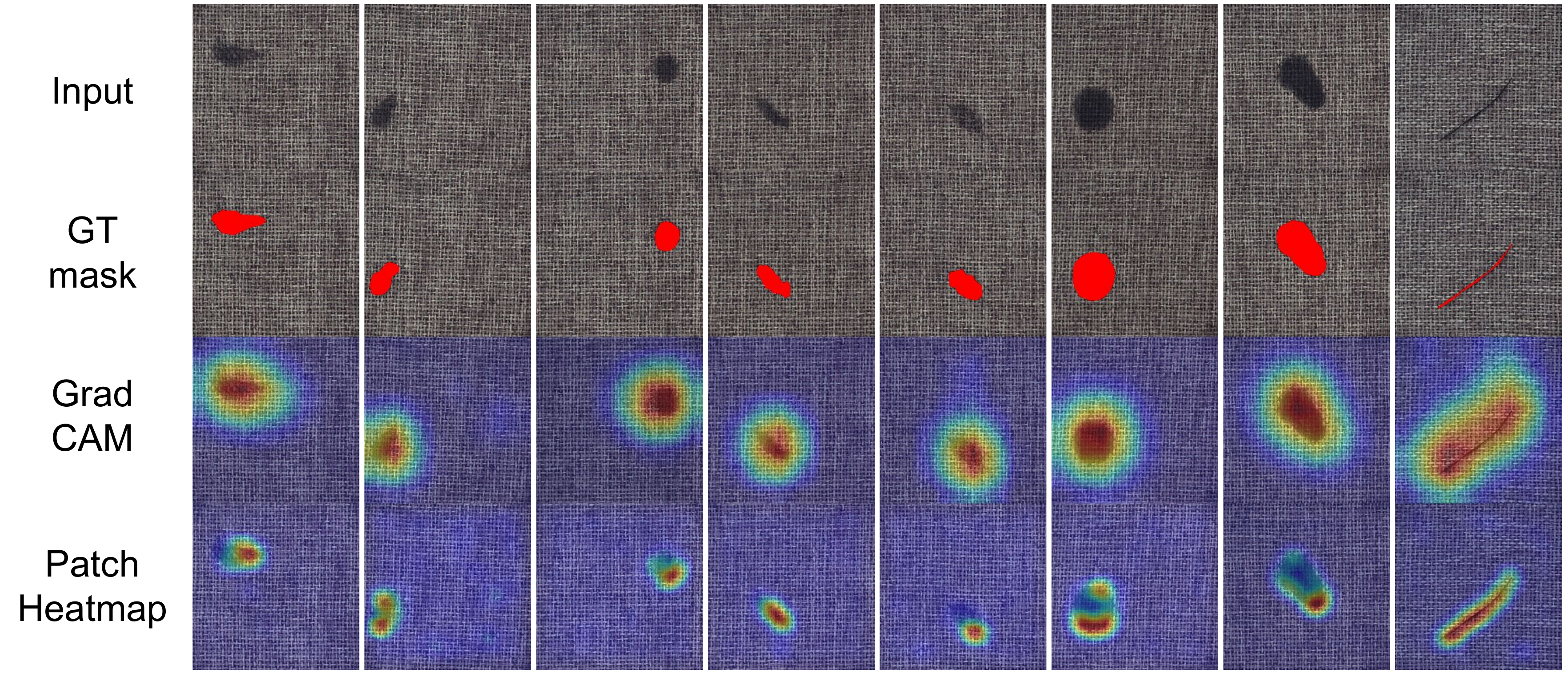}\\
    \vspace{0.05in}
    \includegraphics[width=0.85\textwidth]{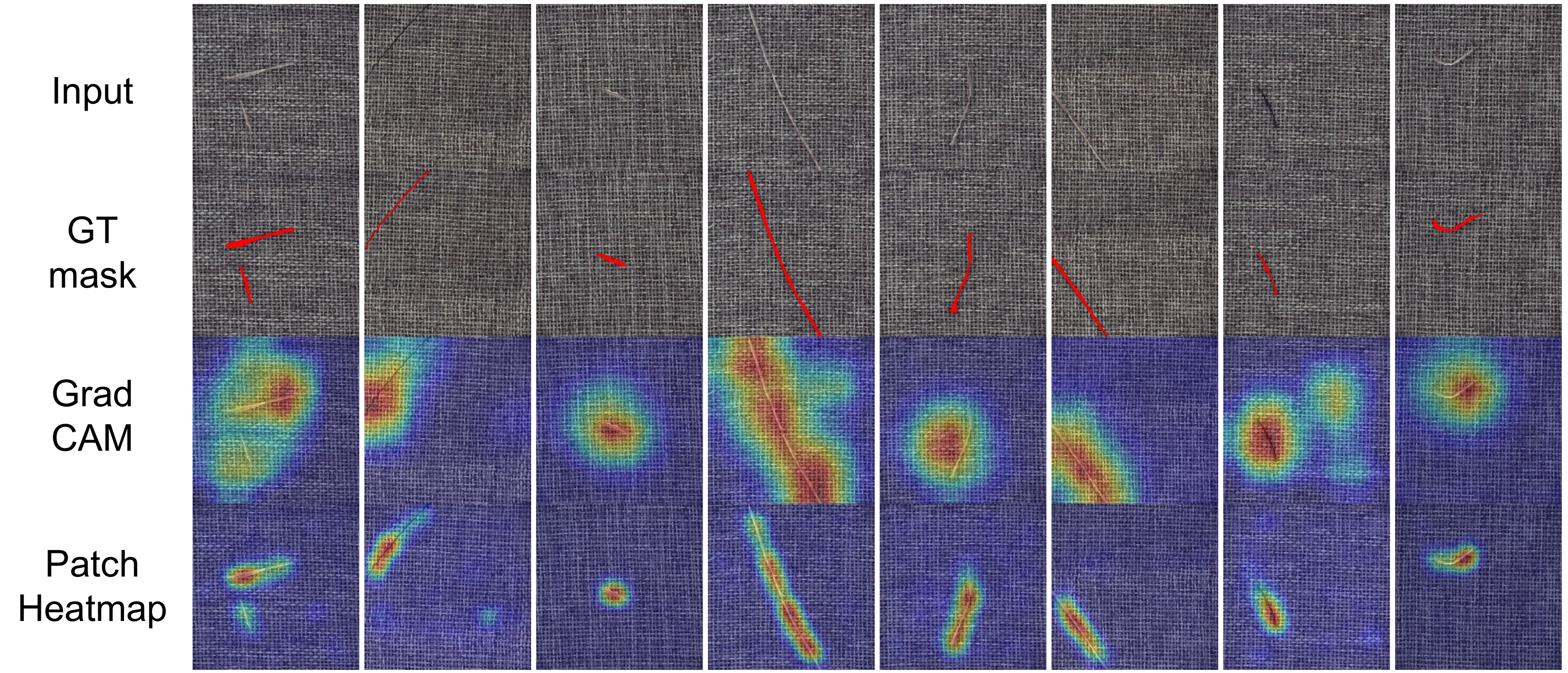}\\
    \vspace{0.05in}
    \includegraphics[width=0.85\textwidth]{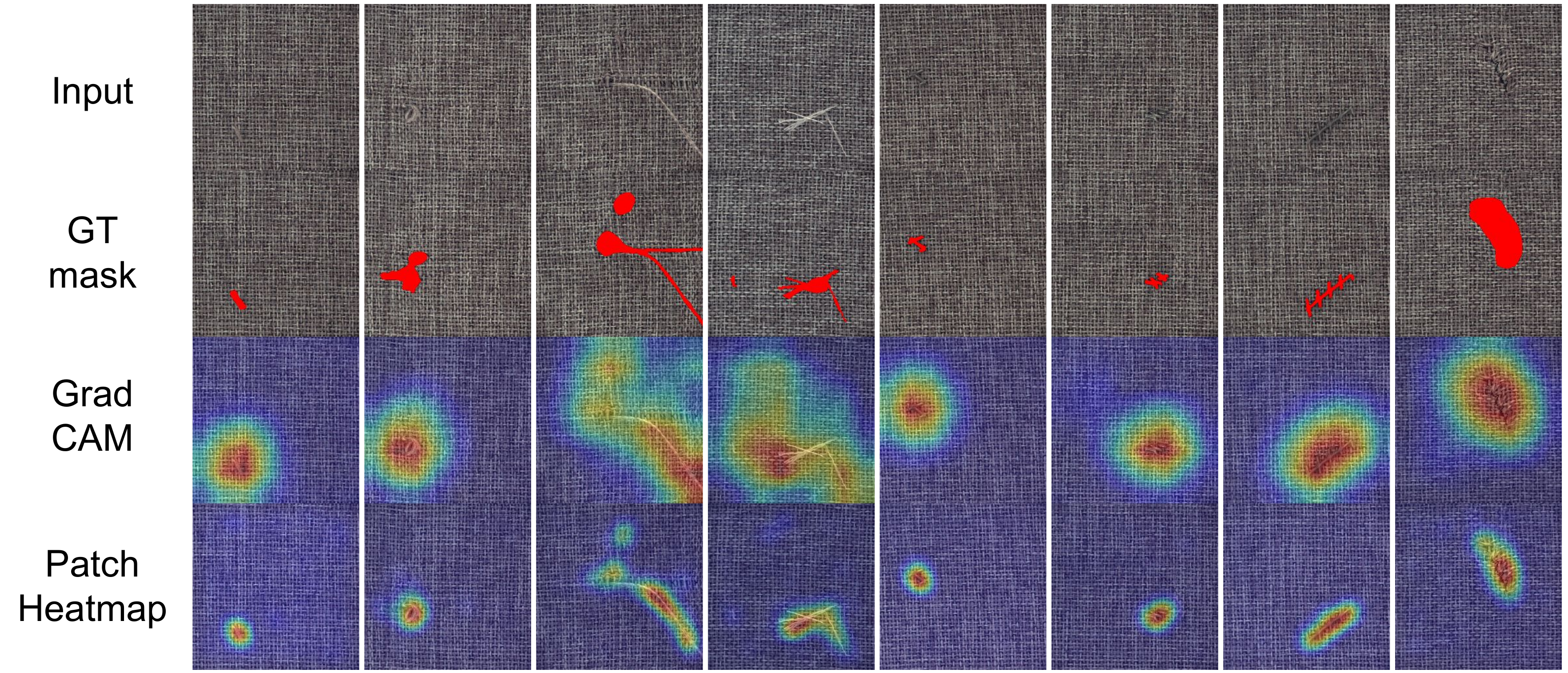}
    \caption{Defect localization on carpet class of MVTec dataset. From top to bottom, input images, those with ground-truth localization mask in red, GradCAM results using image-level detector, and heatmaps using patch-level detector.}
    \label{fig:heatmap_carpet}
\end{figure}

\begin{figure}
    \centering
    \includegraphics[width=0.85\textwidth]{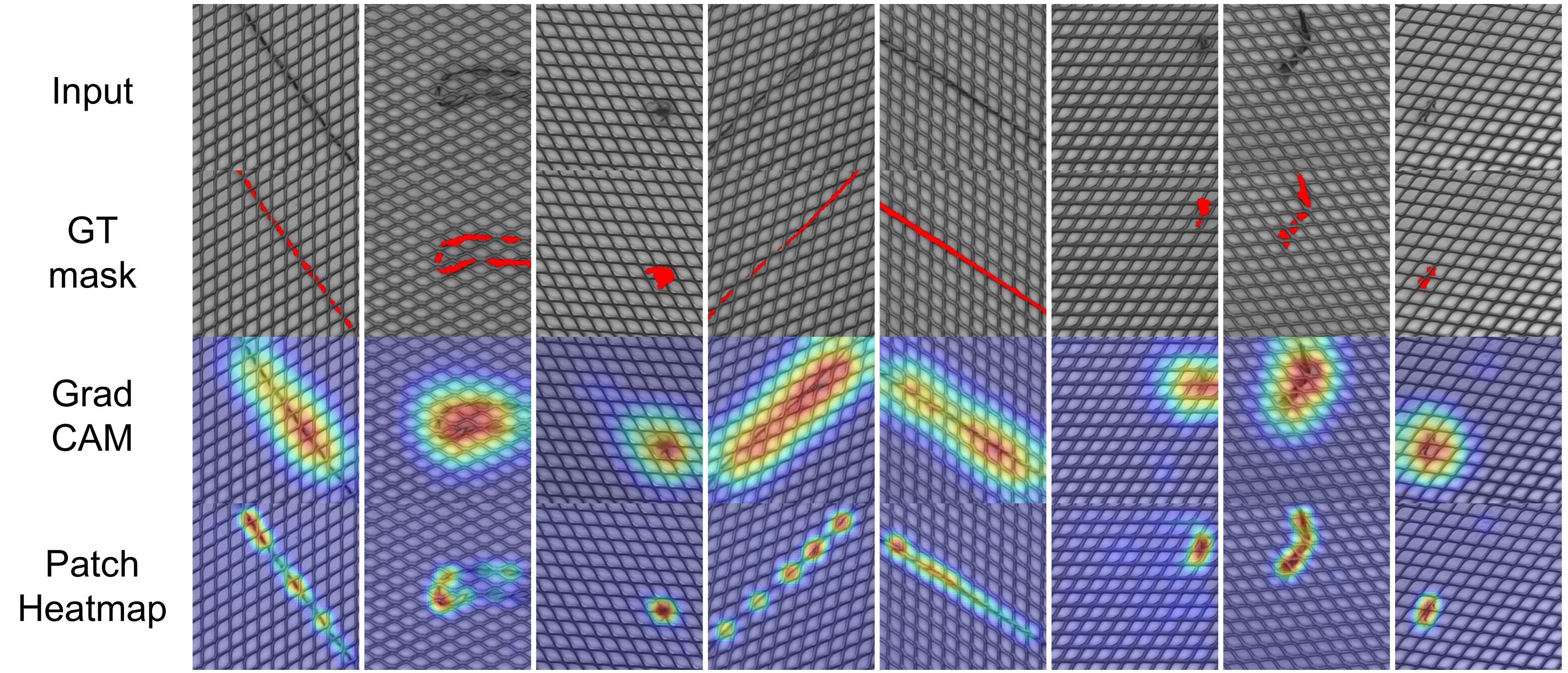}\\
    \vspace{0.05in}
    \includegraphics[width=0.85\textwidth]{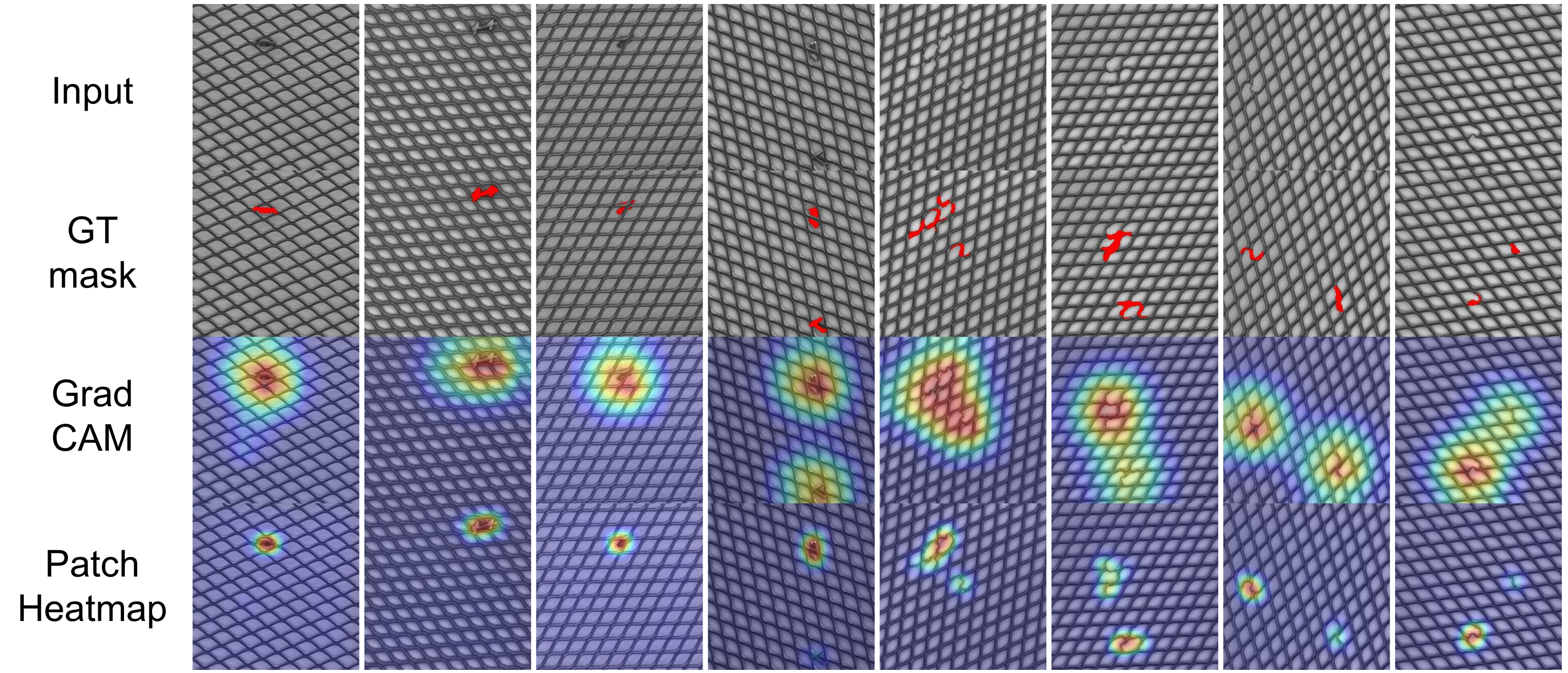}\\
    \vspace{0.05in}
    \includegraphics[width=0.85\textwidth]{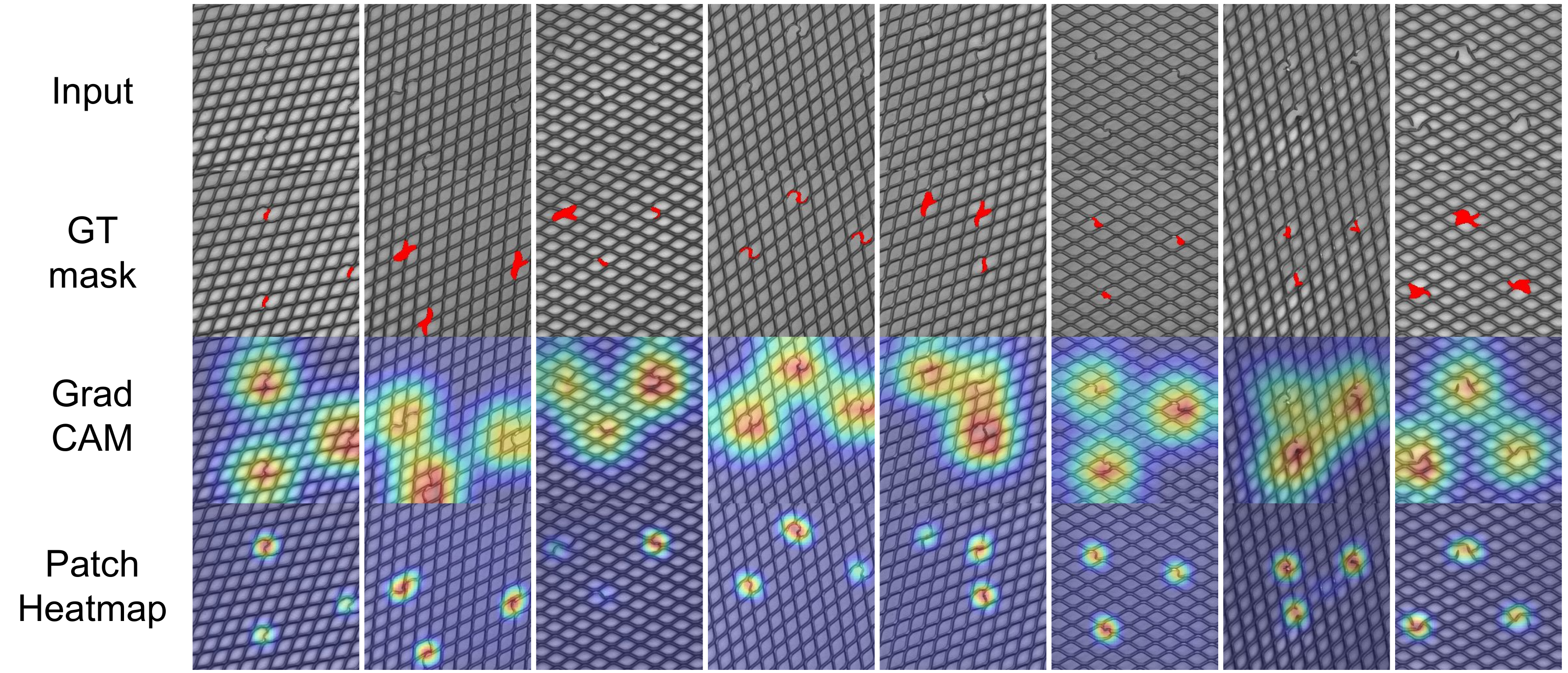}
    \caption{Defect localization on grid class of MVTec dataset. From top to bottom, input images, those with ground-truth localization mask in red, GradCAM results using image-level detector, and heatmaps using patch-level detector.}
    \label{fig:heatmap_grid}
\end{figure}

\begin{figure}
    \centering
    \includegraphics[width=0.85\textwidth]{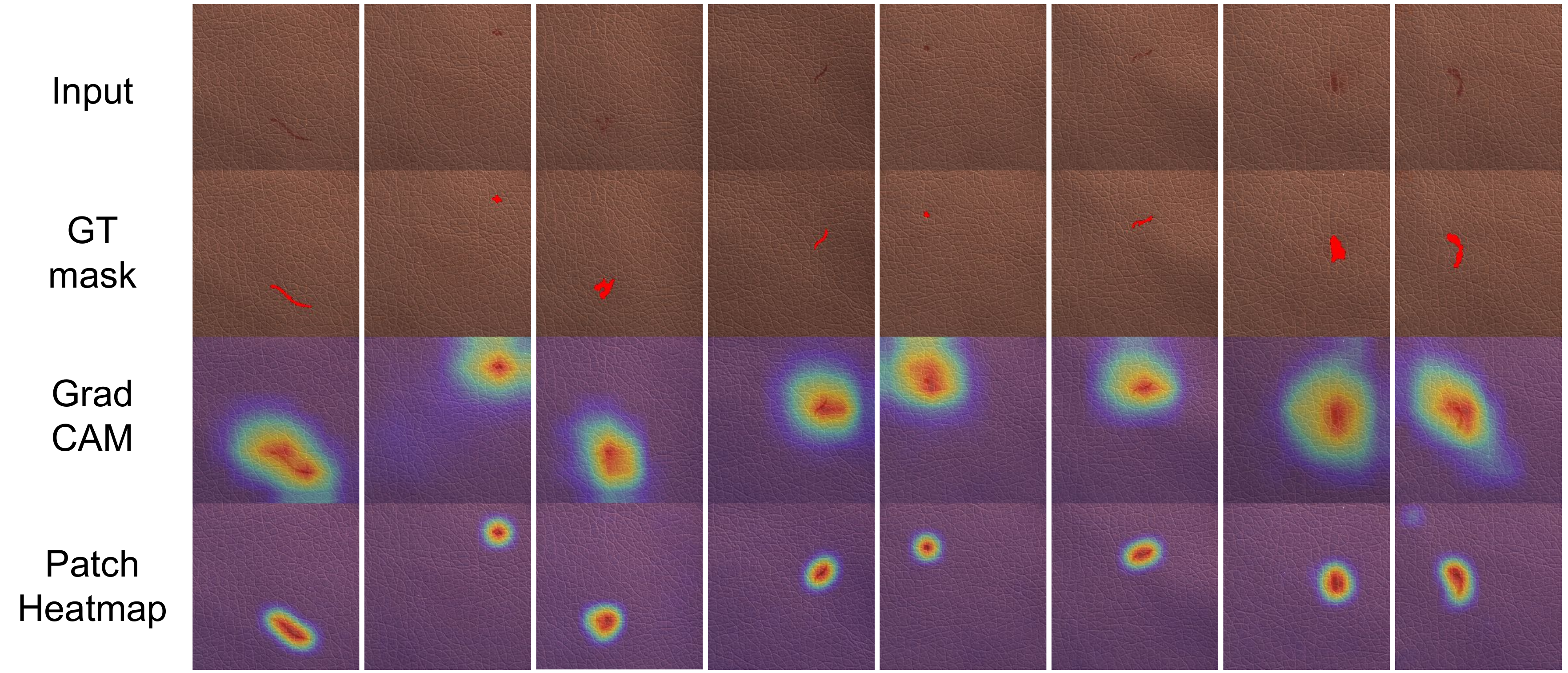}\\
    \vspace{0.05in}
    \includegraphics[width=0.85\textwidth]{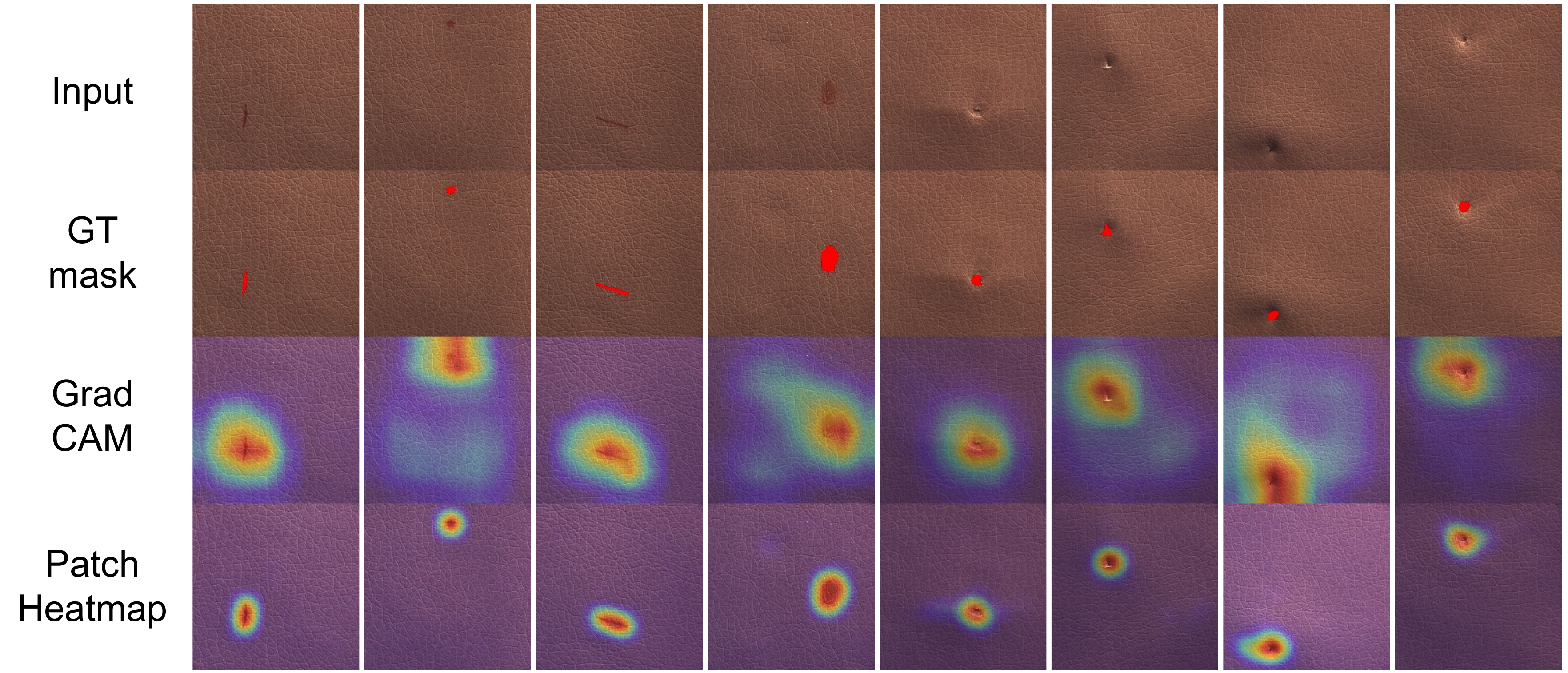}\\
    \vspace{0.05in}
    \includegraphics[width=0.85\textwidth]{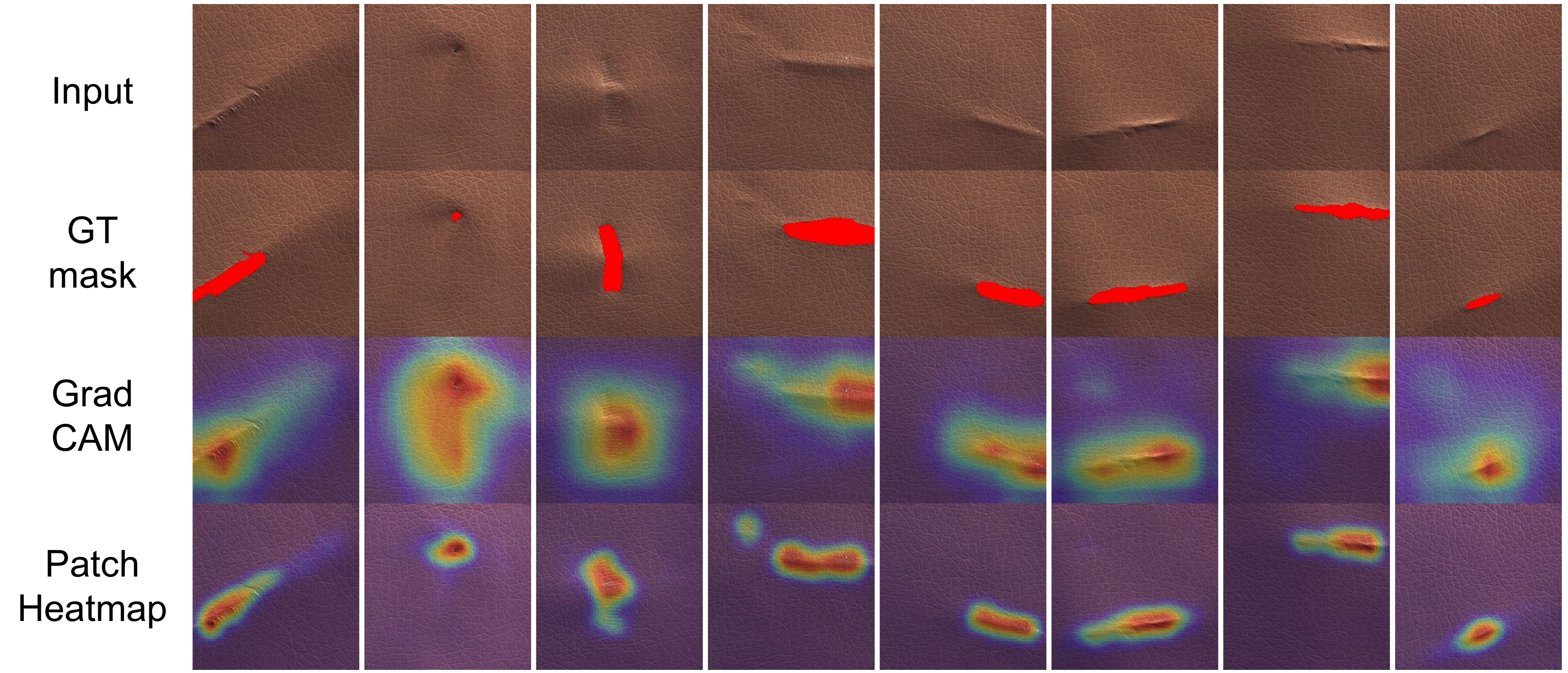}
    \caption{Defect localization on leather class of MVTec dataset. From top to bottom, input images, those with ground-truth localization mask in red, GradCAM results using image-level detector, and heatmaps using patch-level detector.}
    \label{fig:heatmap_leather}
\end{figure}

\begin{figure}
    \centering
    \includegraphics[width=0.85\textwidth]{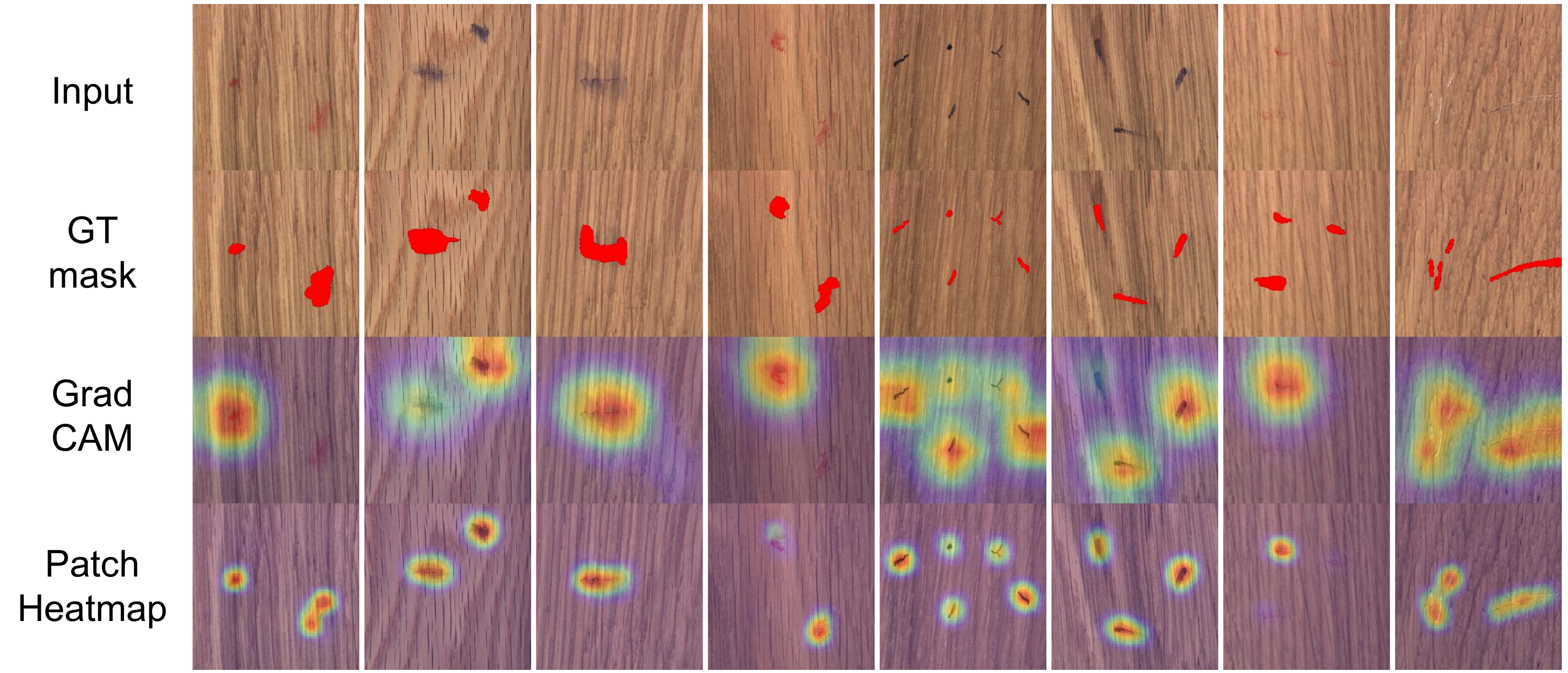}\\
    \vspace{0.05in}
    \includegraphics[width=0.85\textwidth]{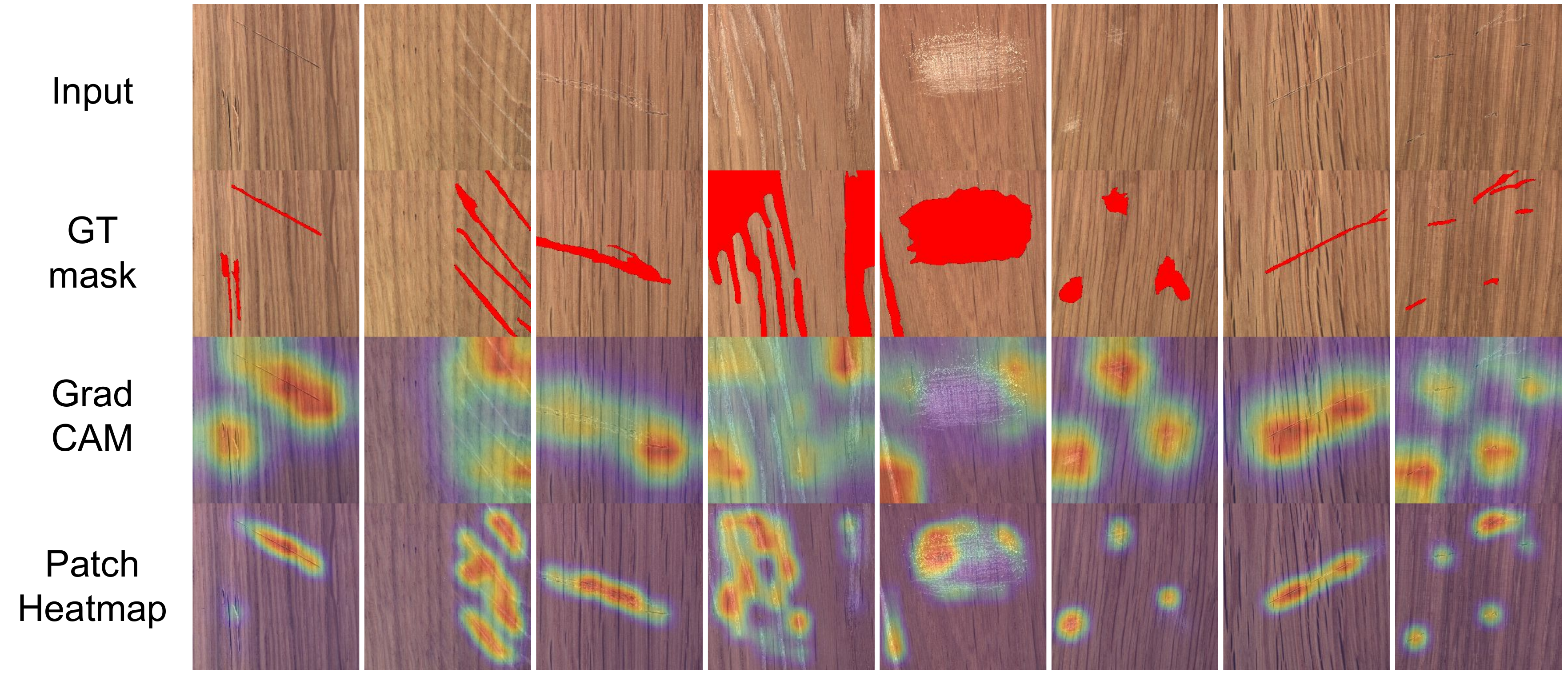}\\
    \vspace{0.05in}
    \includegraphics[width=0.85\textwidth]{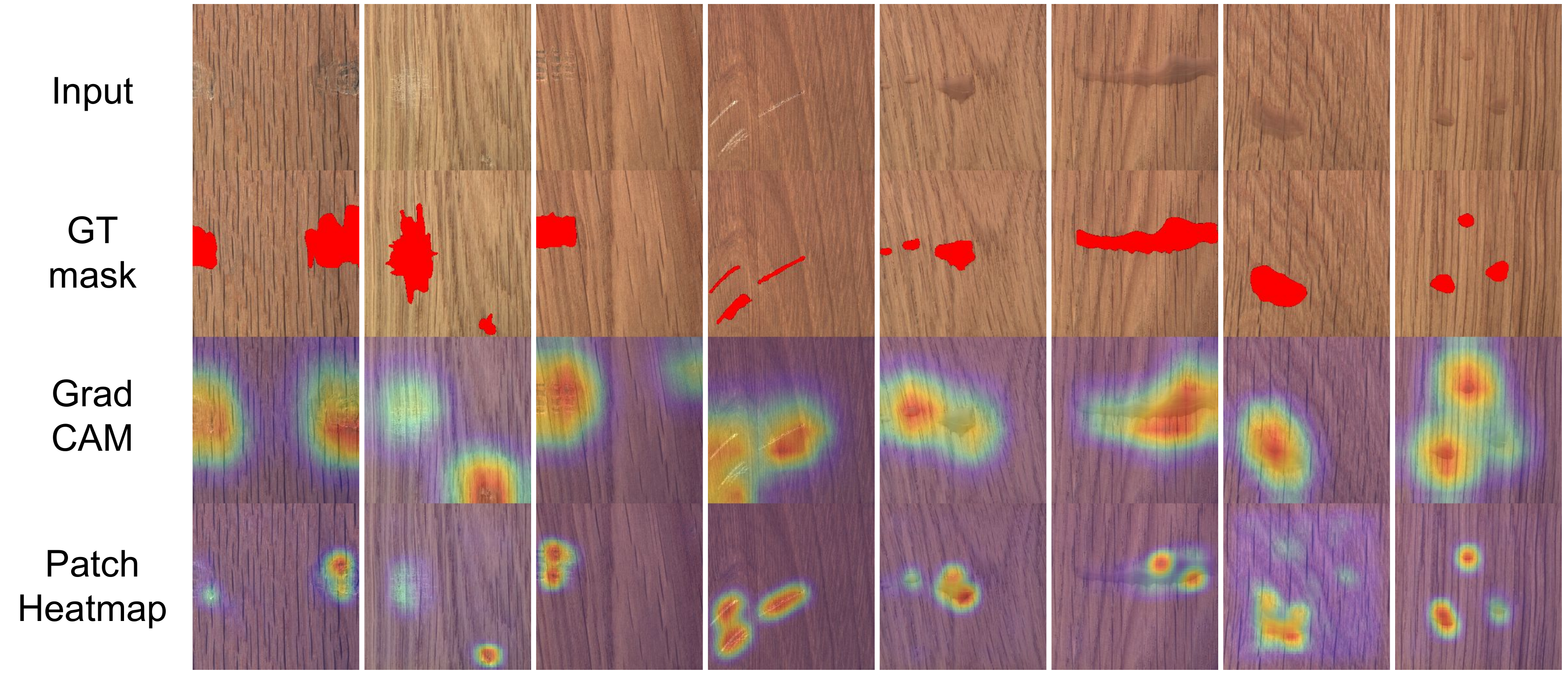}
    \caption{Defect localization on wood class of MVTec dataset. From top to bottom, input images, those with ground-truth localization mask in red, GradCAM results using image-level detector, and heatmaps using patch-level detector.}
    \label{fig:heatmap_wood}
\end{figure}

From Figure~\ref{fig:heatmap_bottle} to Figure~\ref{fig:heatmap_wood}, we show localization visualizations of $24$ examples for each of 10 object and 5 texture categories via GradCAM for image-level CutPaste models and patch heatmap for patch-based models. We not only show successful cases, but also some failure cases.

\subsection{Failure Case Analysis}
\label{sec:app_localization_failure}

Here we enumerate some failure cases. Note that this is not the comprehensive list of all failure cases, but a few representative failure cases we find via visually inspecting localization visualizations.

\begin{figure}
    \centering
    \includegraphics[width=0.85\textwidth]{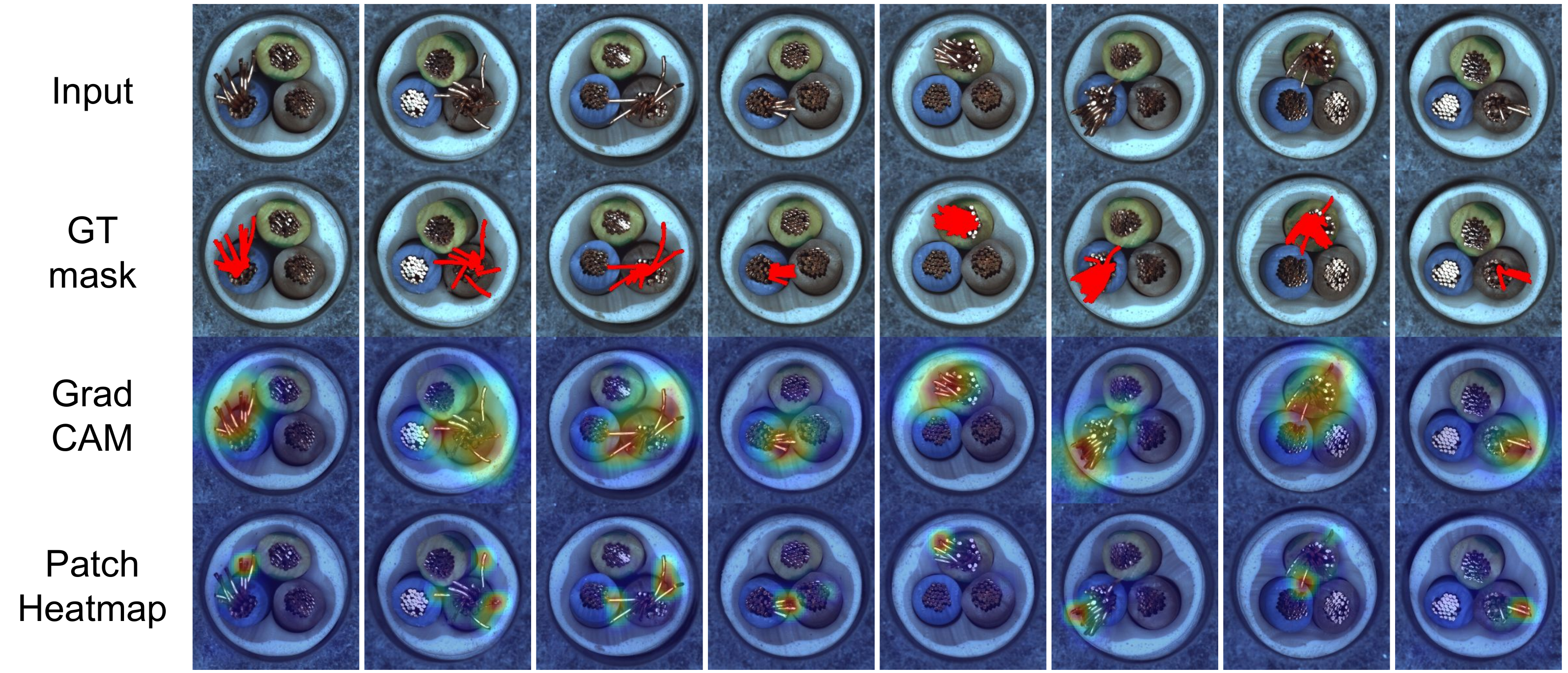}\\
    \vspace{0.05in}
    \includegraphics[width=0.85\textwidth]{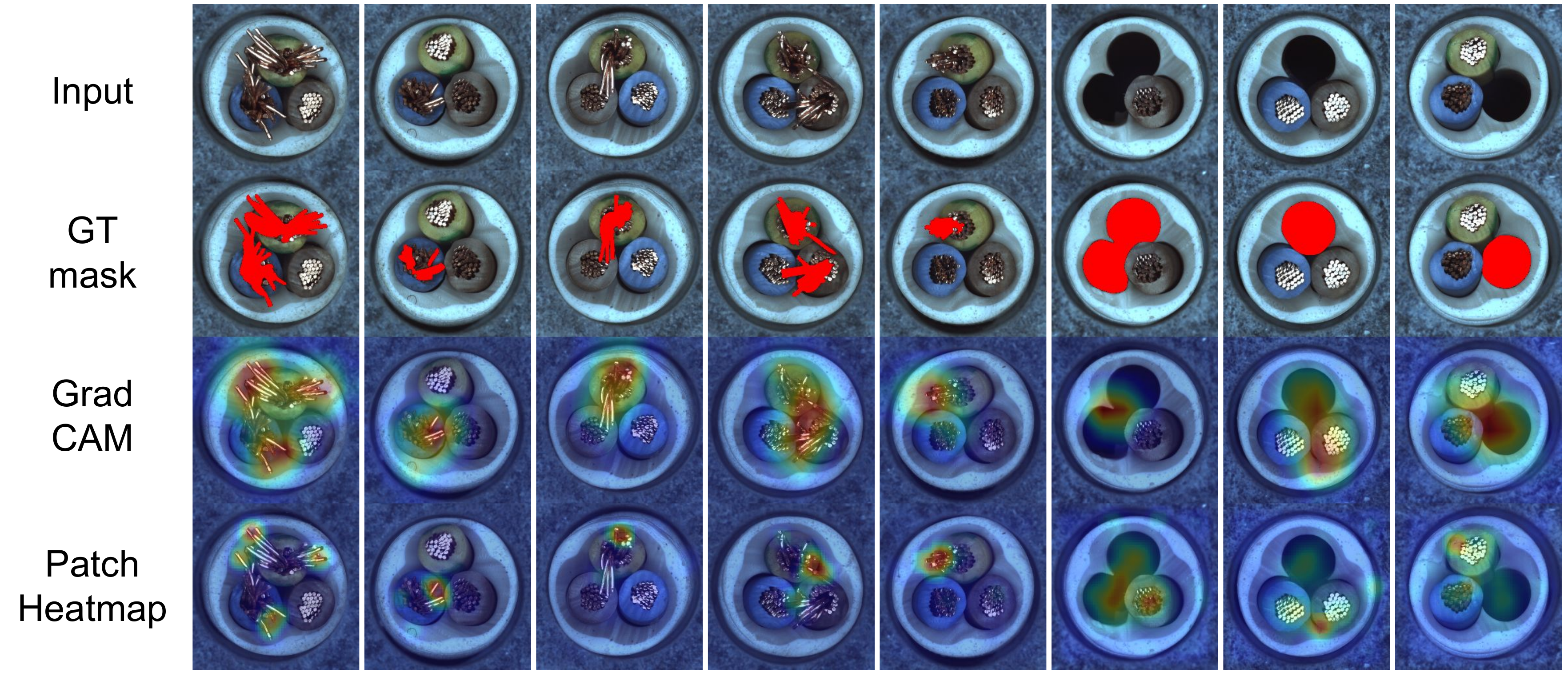}\\
    \vspace{0.05in}
    \includegraphics[width=0.85\textwidth]{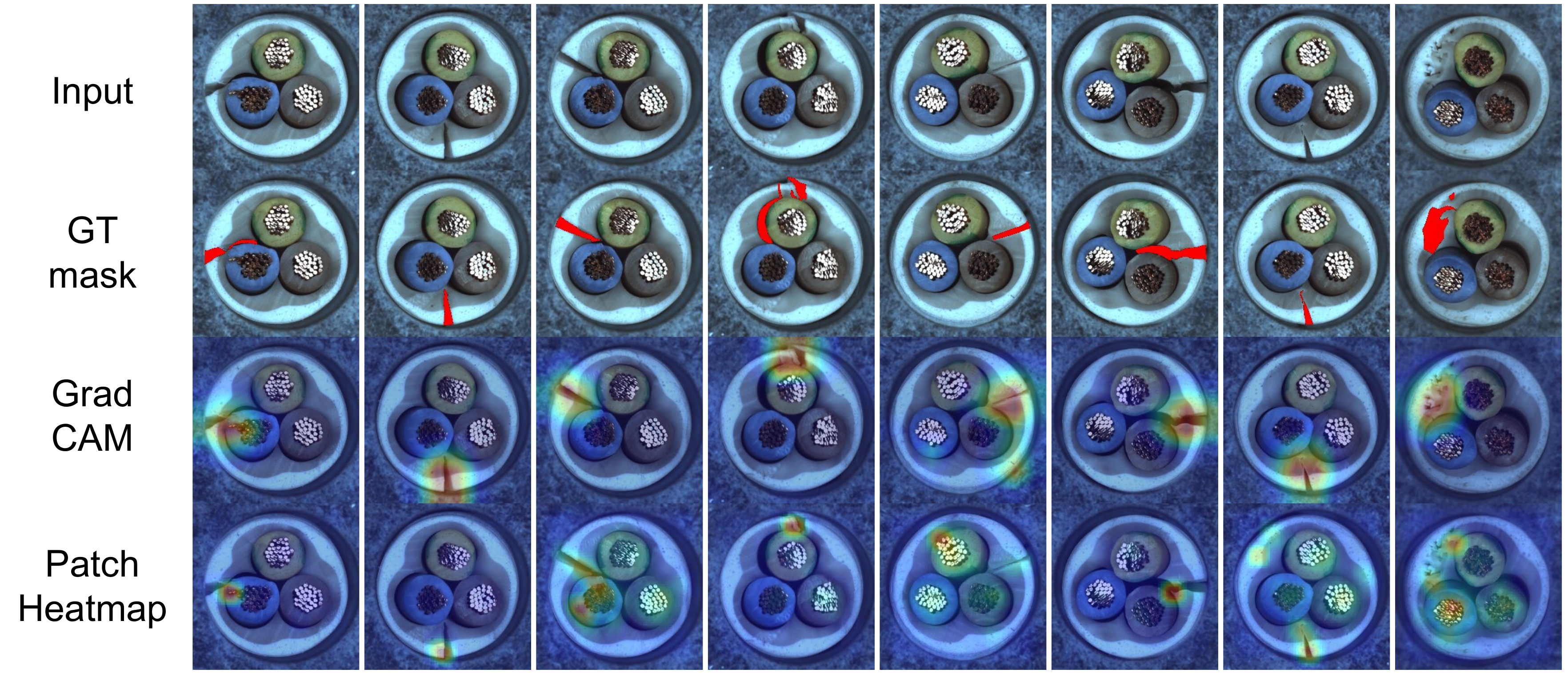}
    \caption{Defect localization on cable class of MVTec dataset. From top to bottom, input images, those with ground-truth localization mask in red, GradCAM results using image-level detector, and heatmaps using patch-level detector.}
    \label{fig:heatmap_cable}
\end{figure}

\begin{figure}
    \centering
    \includegraphics[width=0.85\textwidth]{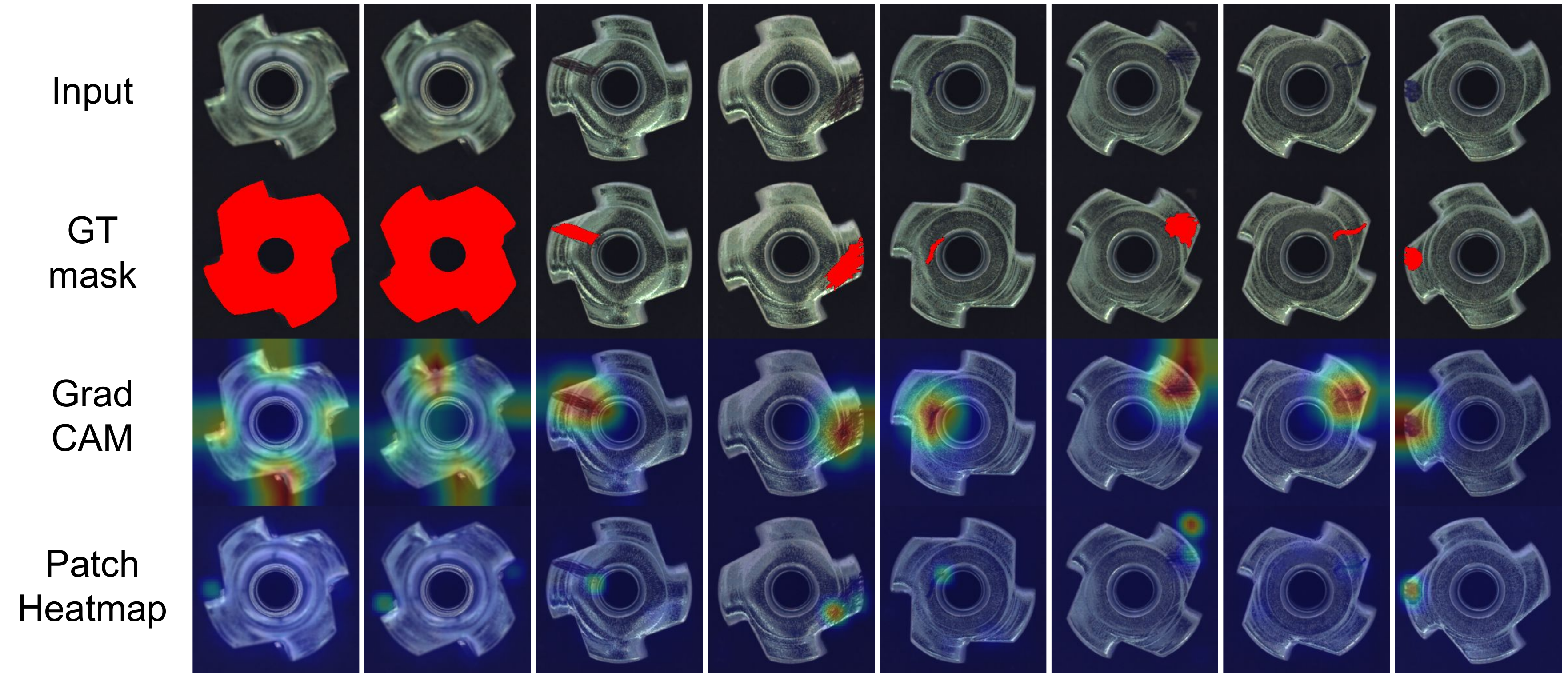}\\
    \vspace{0.05in}
    \includegraphics[width=0.85\textwidth]{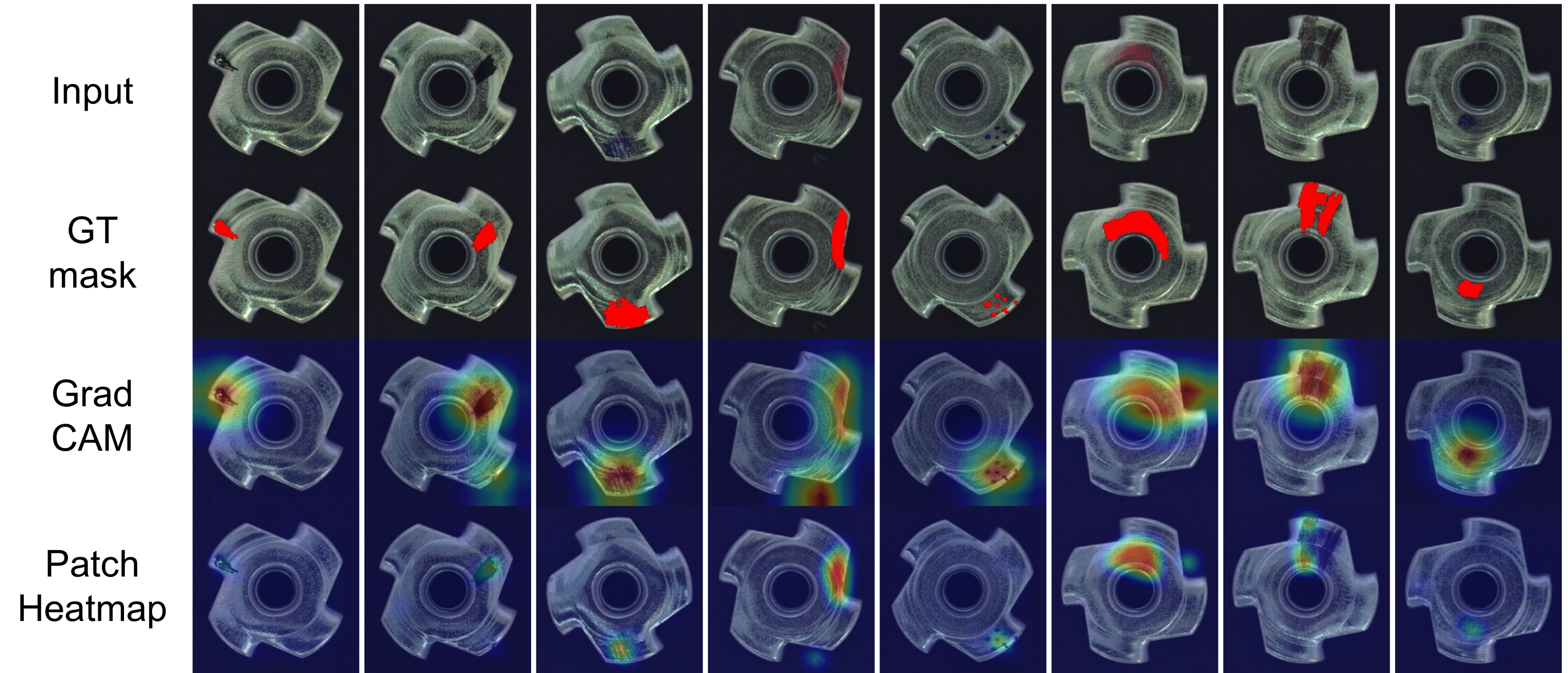}\\
    \vspace{0.05in}
    \includegraphics[width=0.85\textwidth]{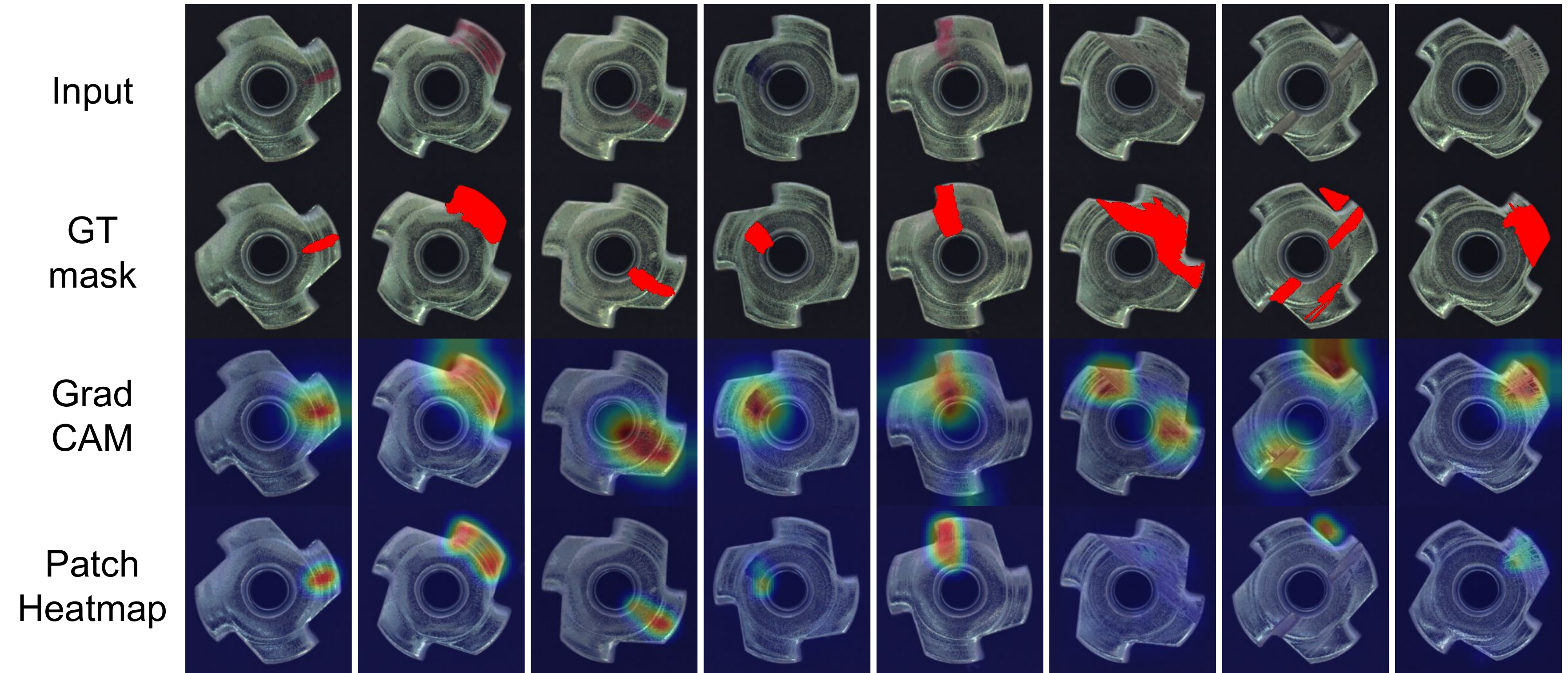}
    \caption{Defect localization on metal nut class of MVTec dataset. From top to bottom, input images, those with ground-truth localization mask in red, GradCAM results using image-level detector, and heatmaps using patch-level detector.}
    \label{fig:heatmap_metalnut}
\end{figure}

\begin{figure}
    \centering
    \includegraphics[width=0.85\textwidth]{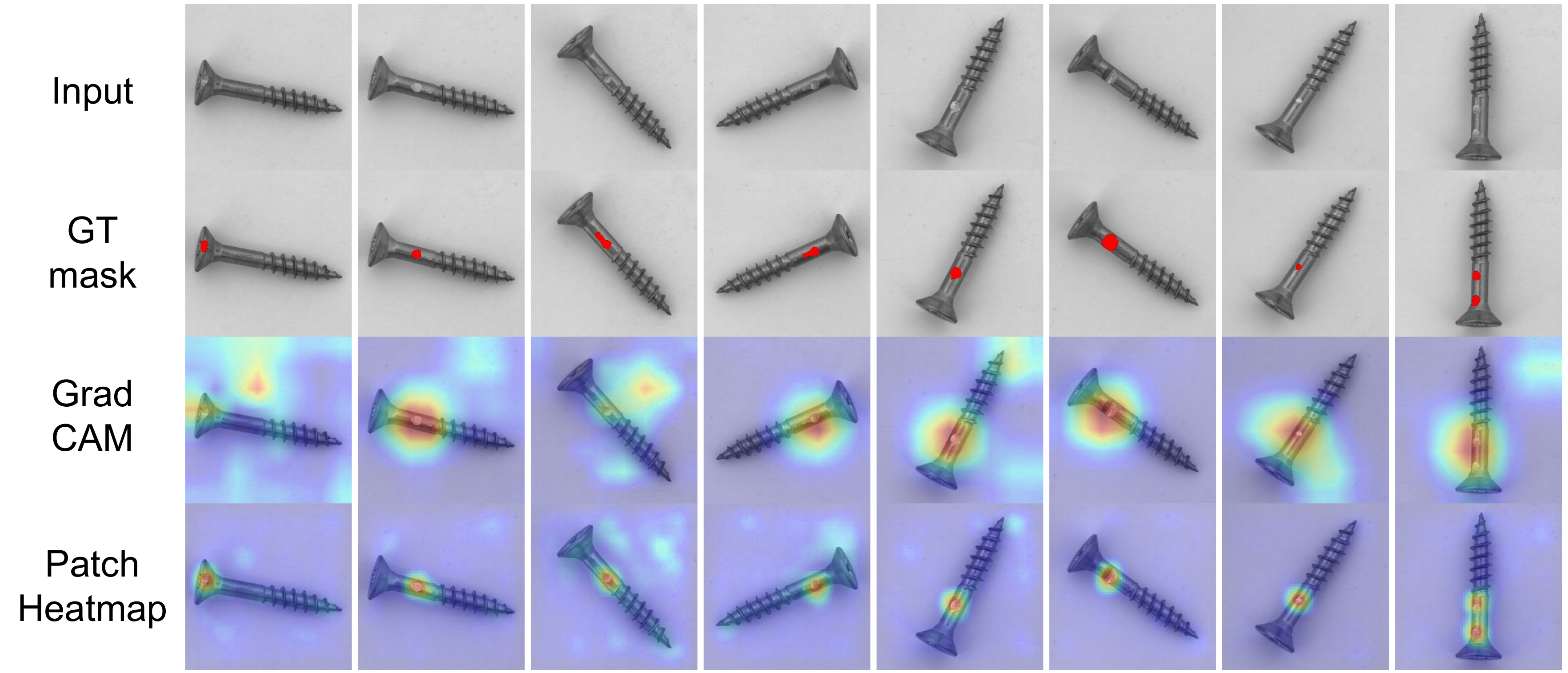}\\
    \vspace{0.05in}
    \includegraphics[width=0.85\textwidth]{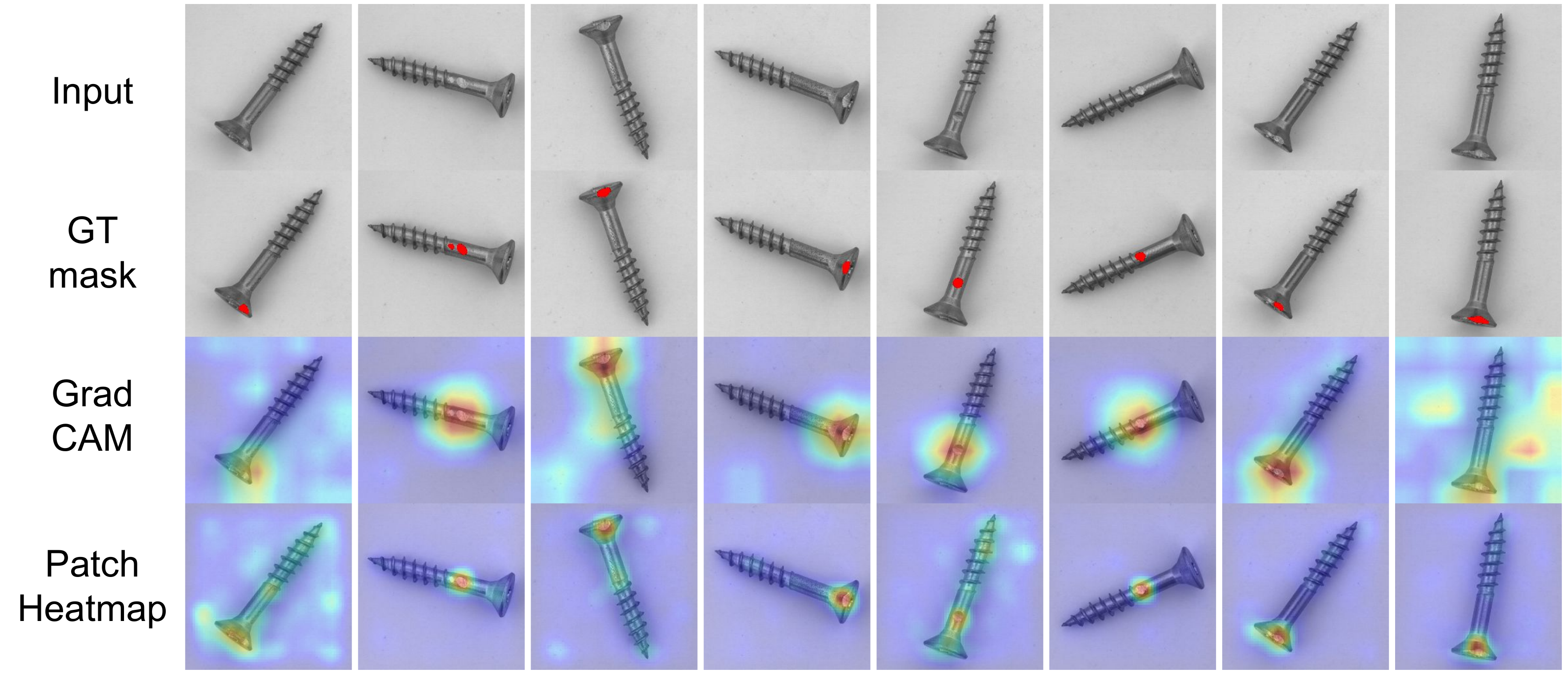}\\
    \vspace{0.05in}
    \includegraphics[width=0.85\textwidth]{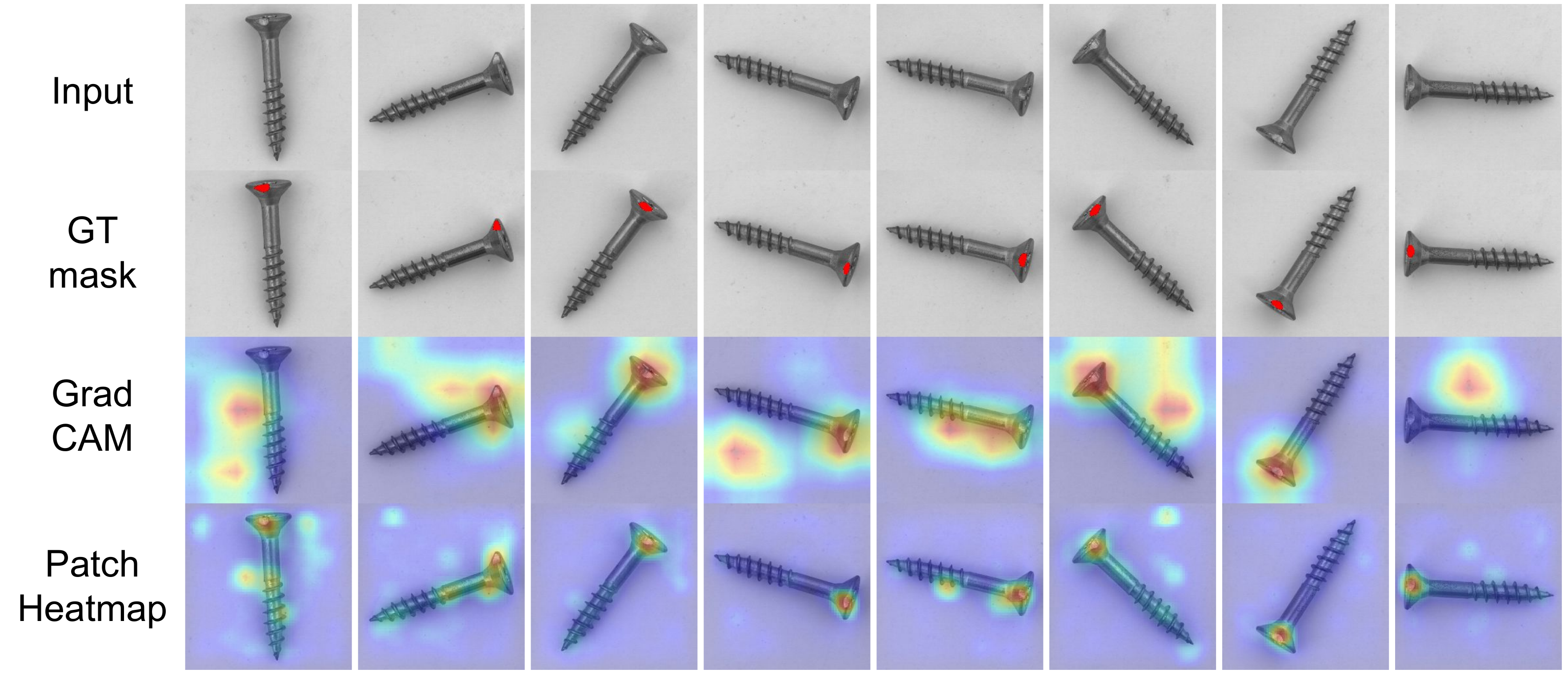}
    \caption{Defect localization on screw class of MVTec dataset. From top to bottom, input images, those with ground-truth localization mask in red, GradCAM results using image-level detector, and heatmaps using patch-level detector.}
    \label{fig:heatmap_screw}
\end{figure}

\begin{figure}
    \centering
    \includegraphics[width=0.85\textwidth]{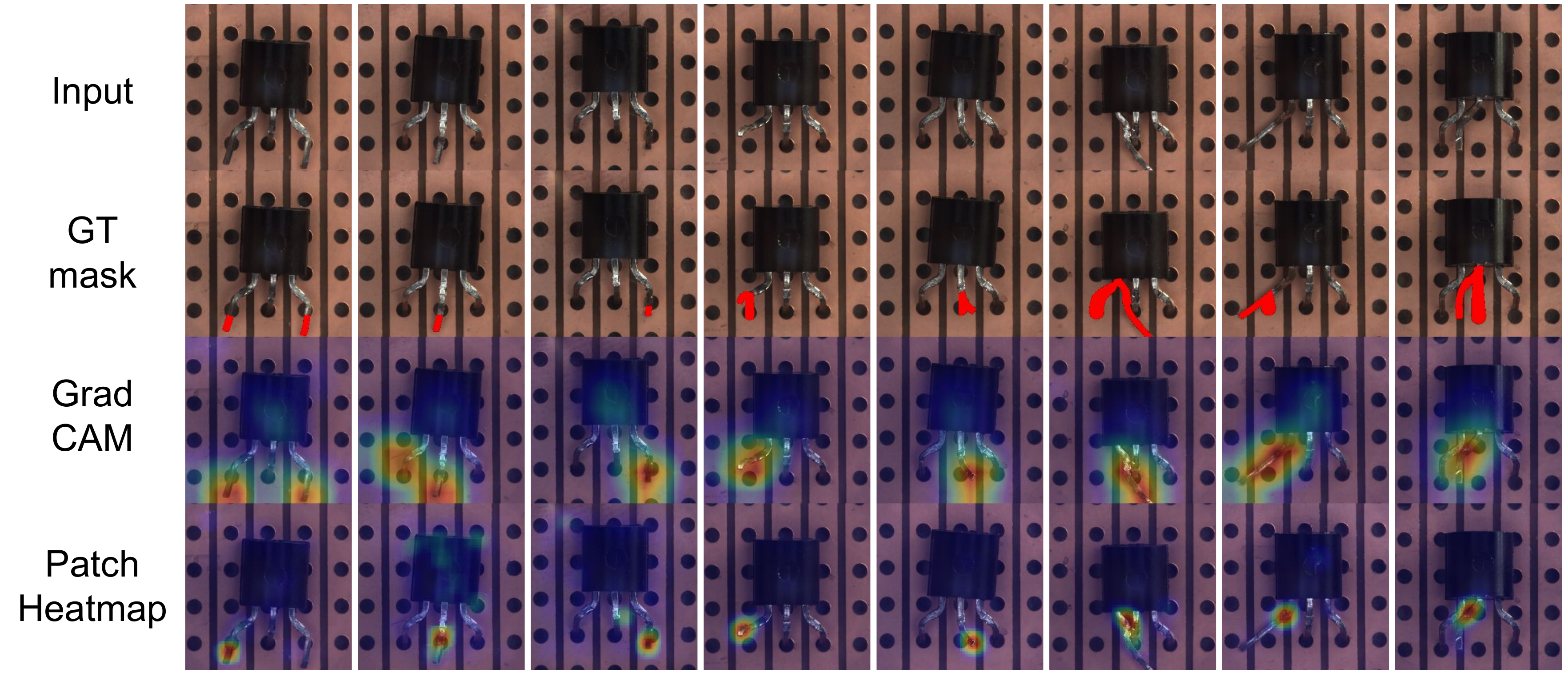}\\
    \vspace{0.05in}
    \includegraphics[width=0.85\textwidth]{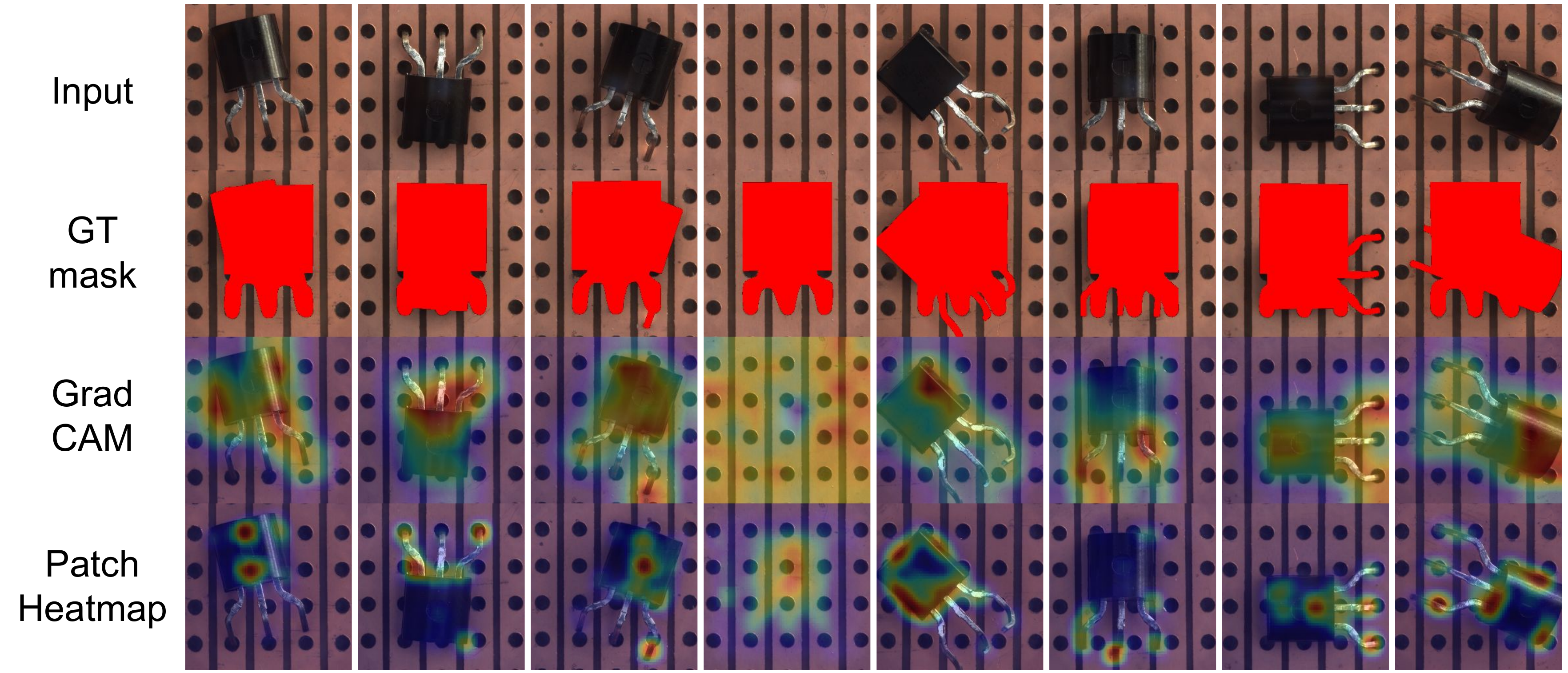}\\
    \vspace{0.05in}
    \includegraphics[width=0.85\textwidth]{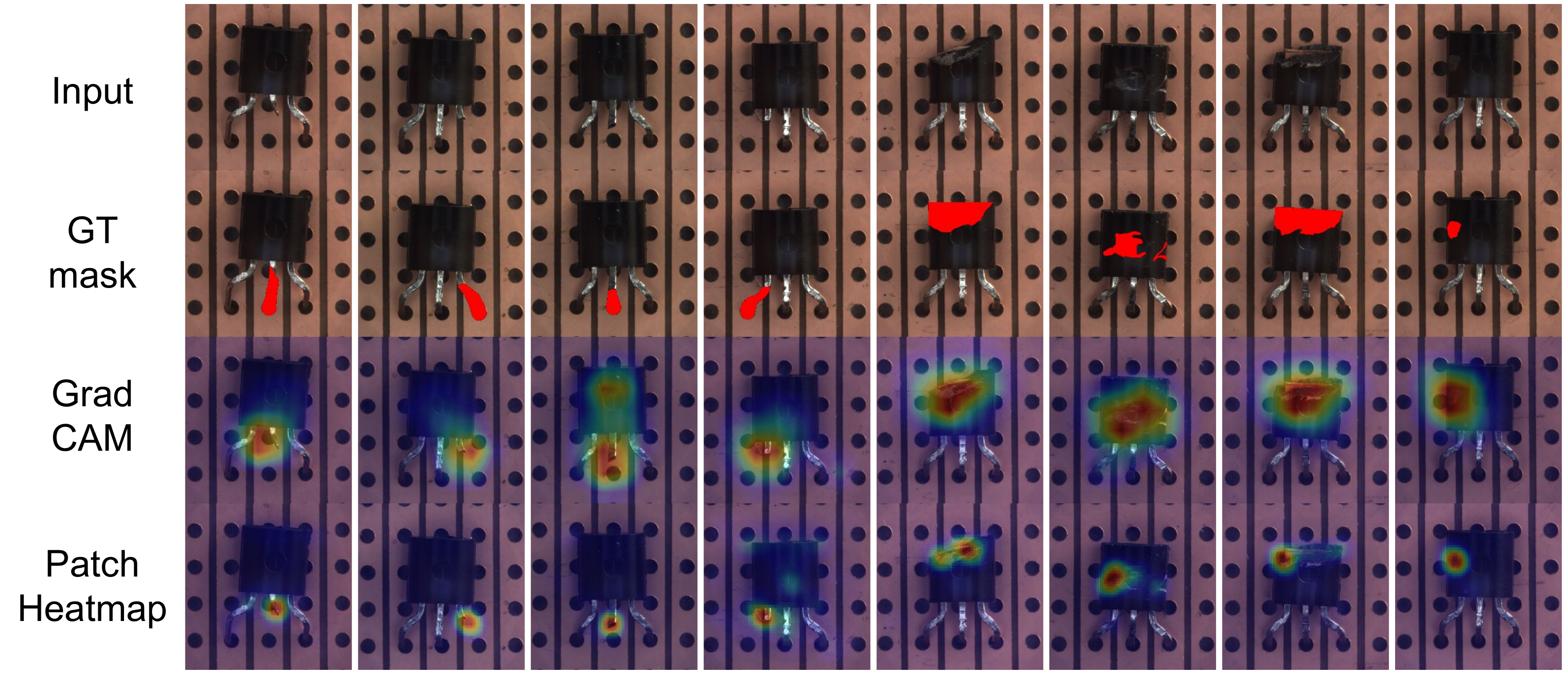}
    \caption{Defect localization on transistor class of MVTec dataset. From top to bottom, input images, those with ground-truth localization mask in red, GradCAM results using image-level detector, and heatmaps using patch-level detector.}
    \label{fig:heatmap_transistor}
\end{figure}

\begin{figure}
    \centering
    \includegraphics[width=0.85\textwidth]{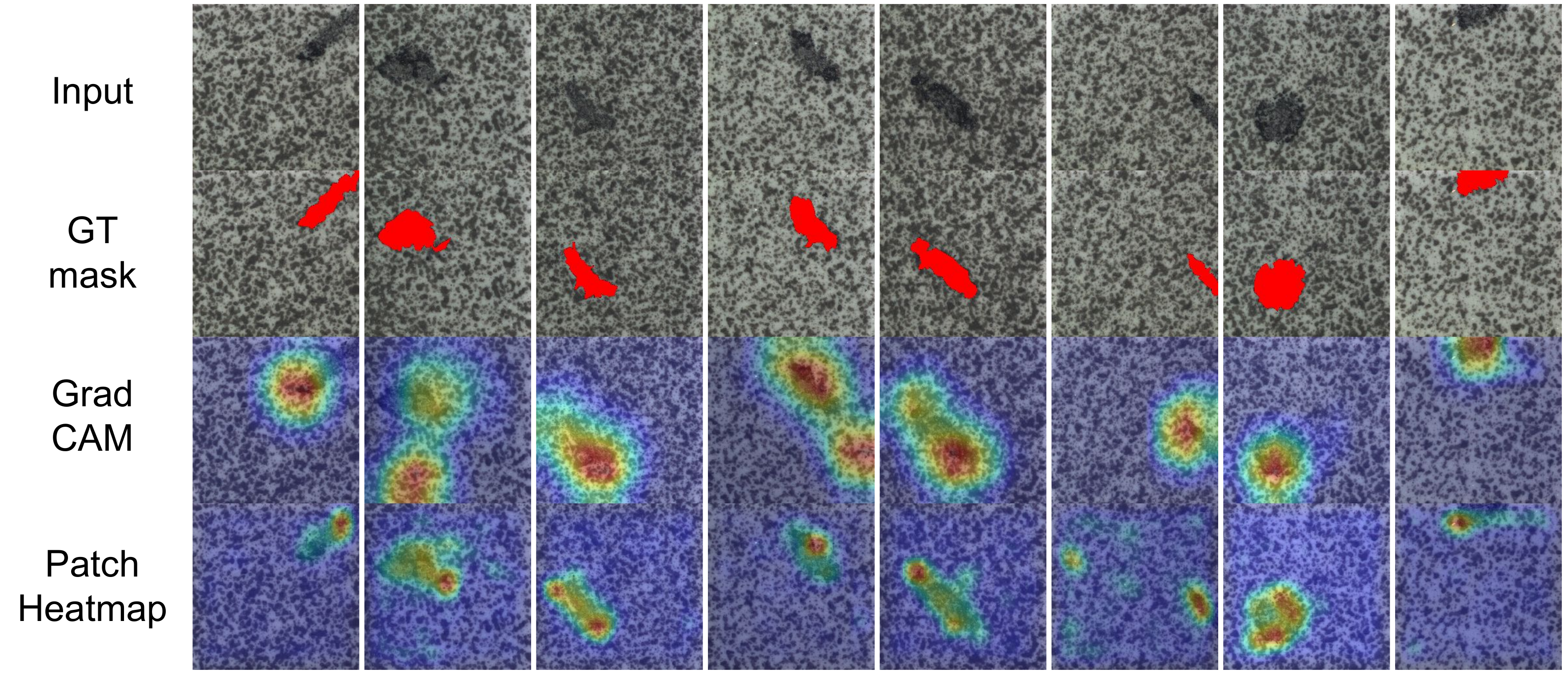}\\
    \vspace{0.05in}
    \includegraphics[width=0.85\textwidth]{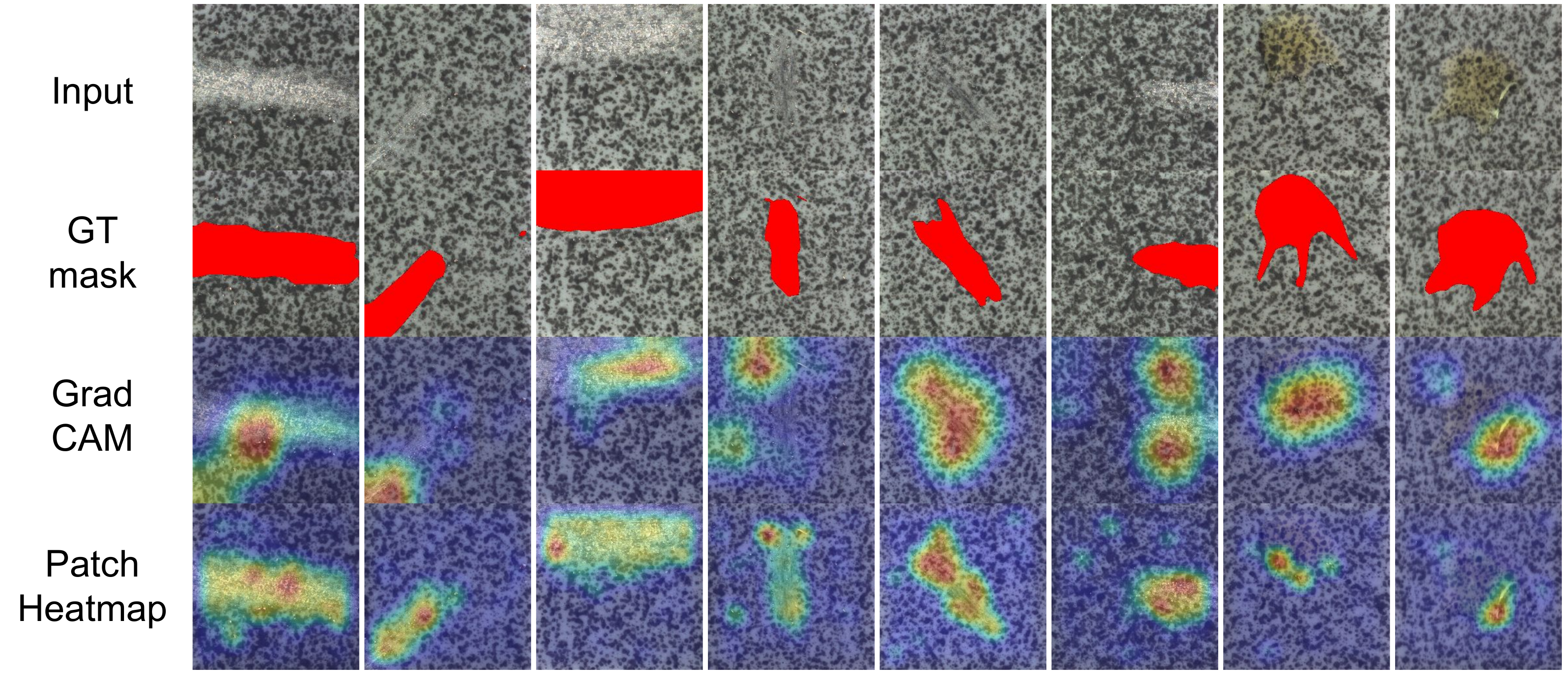}\\
    \vspace{0.05in}
    \includegraphics[width=0.85\textwidth]{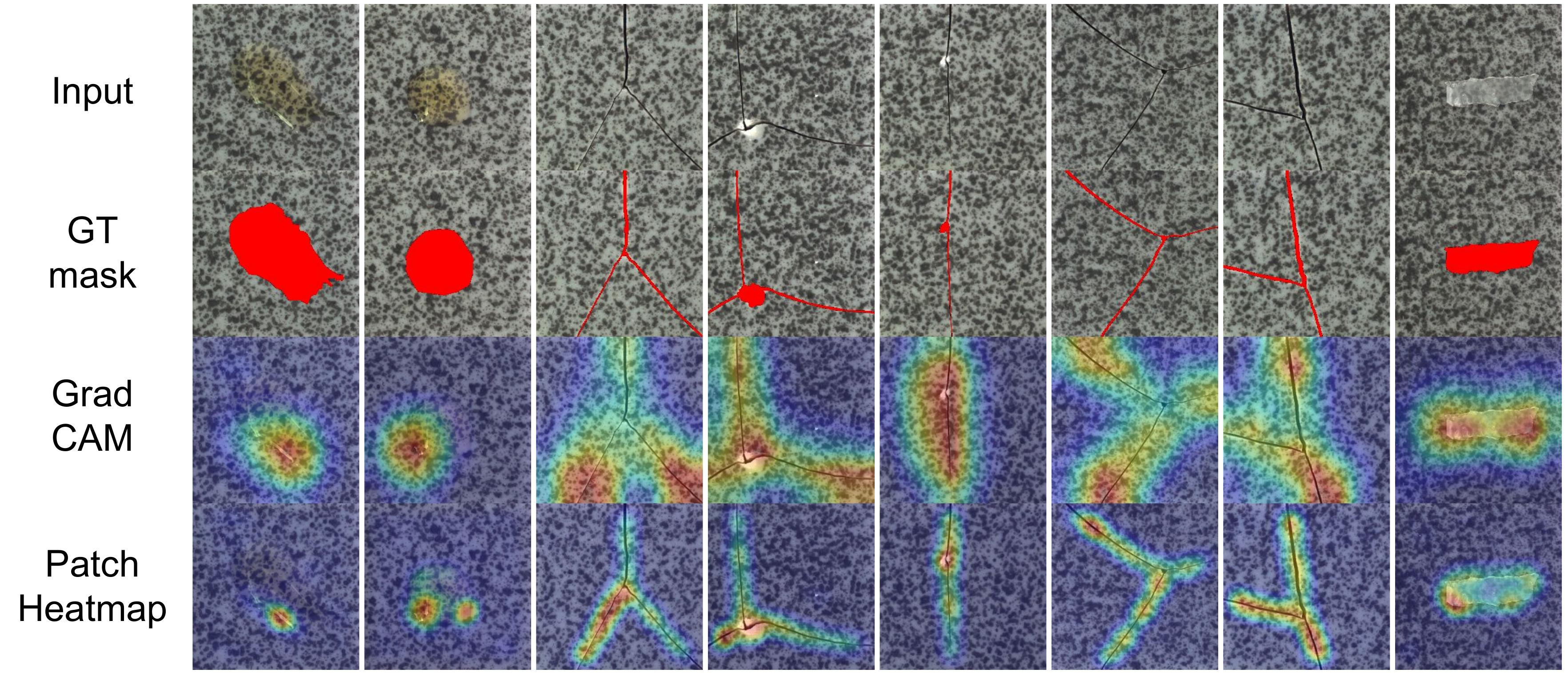}
    \caption{Defect localization on tile class of MVTec dataset. From top to bottom, input images, those with ground-truth localization mask in red, GradCAM results using image-level detector, and heatmaps using patch-level detector.}
    \label{fig:heatmap_tile}
\end{figure}

\begin{itemize}\setlength\itemsep{0em}
    \item {Cable (Figure~\ref{fig:heatmap_cable}): Missing components (row 2 column 6--8).}
    \item {Metal Nut (Figure~\ref{fig:heatmap_metalnut}): Flipped components (row 1 column 1--2).}
    \item {Screw (Figure~\ref{fig:heatmap_screw}): Speckle noise in the background (row 3 column 1, 6).}
    \item {Transistor (Figure~\ref{fig:heatmap_transistor}): Dislocated or missing components (row 2).}
    \item {Tile (Figure~\ref{fig:heatmap_tile}): dyed tiles (row 2 column 7--8, row 3 column 1--2).}
\end{itemize}

\end{document}